\documentclass[sigconf]{acmart}

\AtBeginDocument{%
  }

\acmDOI{XXXXXXX.XXXXXXX}

\acmConference[AsiaCCS 2023]{}
\acmPrice{15.00}
\acmISBN{978-1-4503-XXXX-X/18/06}

\usepackage{algorithm}
\usepackage{algpseudocode}%
\usepackage{amsmath}
\usepackage{caption}

\usepackage{amssymb}

\usepackage{comment}
\usepackage{float}
\usepackage{lipsum}
\usepackage{varwidth}
\usepackage{array}
\usepackage{booktabs}
\usepackage{adjustbox}
\usepackage{subcaption}
\usepackage{xspace}

\usepackage{bbm}
\usepackage{multirow}
\usepackage{soul,color}
\usepackage{listings}
\usepackage{amsthm}

\newcolumntype{P}[1]{>{\centering\arraybackslash}p{#1}}
\newcolumntype{M}[1]{>{\centering\arraybackslash}m{#1}}
\newcommand{\eg}{{e.g.,}\xspace}

\newcommand{\ie}{{i.e.,}\xspace}

\usepackage{array}
\usepackage{url}

\begin{document}

\title{LDL: A Defense for Label-Based Membership Inference Attacks}


\author{Arezoo Rajabi}
\affiliation{%
  \institution{University of Washington}
  \streetaddress{University of Washington}
  \city{Seattle}
  \country{USA}}
\email{rajabia@uw.edu}

\author{Dinuka Sahabandu}
\affiliation{%
  \institution{University of Washington}
  \streetaddress{University of Washington}
  \city{Seattle}
  \country{USA}}
\email{sdinuka@uw.edu}

\author{Luyao Niu}
\affiliation{%
  \institution{University of Washington}
  \streetaddress{University of Washington}
  \city{Seattle}
  \country{USA}}
\email{luyaoniu@uw.edu}

\author{Bhaskar Ramasubramanian}
\affiliation{%
  \institution{Western Washington University}
  \streetaddress{Western Washington University}
  \city{Bellingham}
  \country{USA}}
\email{ramasub@wwu.edu}

\author{Radha Poovendran}
\affiliation{%
  \institution{University of Washington}
  \streetaddress{University of Washington}
  \city{Seattle}
  \country{USA}}
\email{rp3@uw.edu}








  \begin{abstract}
The data used to train deep neural network (DNN) models in applications such as healthcare and finance typically contain sensitive information. 
A DNN model may suffer from overfitting-- it will perform very well on samples seen during training, and poorly on samples not seen during training. 
Overfitted models have been shown to be susceptible to query-based 
attacks such as membership inference attacks (MIAs). 
MIAs aim to determine whether a sample belongs to the dataset used to train a classifier (members) or not (nonmembers). 
Recently, a new class of \emph{label-based MIAs} (LAB MIAs) was proposed, where an adversary was only required to have knowledge of predicted labels of samples. 
LAB MIAs used the insight that member samples were typically located farther away from a classification decision boundary than nonmembers, and were shown to be highly effective across multiple datasets. 
Developing a defense against an adversary carrying out a LAB MIA on DNN models that cannot be retrained remains an open problem. 

We present \textbf{LDL}, 
a light weight defense against LAB MIAs. 
LDL works by constructing a high-dimensional sphere around queried samples 
such that the model decision is unchanged for (noisy) variants of the sample 
within the sphere. 
This sphere of \emph{label-invariance} creates ambiguity and prevents a querying adversary from correctly determining whether a sample is a member or a nonmember. 
We analytically characterize the 
success rate of an adversary carrying out a LAB MIA when LDL is deployed, and show that the formulation is 
consistent with experimental observations. 
We evaluate LDL on seven datasets-- CIFAR-10, CIFAR-100, GTSRB, Face, Purchase, Location, and Texas-- with varying sizes of training data. 
All of these datasets have been used by SOTA  LAB MIAs. 
Our experiments demonstrate that LDL reduces the success rate of an adversary carrying out a LAB MIA in each case. 
We empirically compare LDL with defenses against LAB MIAs that require retraining of DNN models, and show that LDL performs favorably \emph{despite not needing to retrain} the DNNs. 
\end{abstract}



\maketitle

\section{Introduction}
\label{sec:Intro}
%

%
Advances in machine learning (ML) combined with acceleration in the development of parallelizable computing resources have enabled designing complex models for a wide range of privacy-critical applications~\cite{liu2018survey, liu2020privacy}. 
An example is the use of deep neural networks (DNNs) for medical image analysis~\cite{ccs2021}. 
DNNs have been shown to be able to match the performance of human experts in some domains~\cite{DNNsCapabilty}. However, 
recent research has demonstrated that DNN models are vulnerable to privacy attacks~\cite{liu2021machine, rigaki2020survey}. 
One such privacy attack that was shown to reveal sensitive information about data used to train DNN models is a membership inference attack (MIA)~\cite{veale2018awhyMIA}. 
A MIA aims to determine whether a specific sample belongs to the training set of a DNN model or not~\cite{PETS2021,jia2019memguard,shokri2017membership,ye2022one,ye2021shokri2,yeom2018privacy}. 
Information about membership of a sample available to an adversary can result in the revelation of sensitive information, which might pose a serious privacy threat. As an example, in ~\cite{carlini2019secret}, an adversary was shown to have the ability to infer social security numbers that belonged to the training set of a generative language model. 
In this paper, we will use the terms \emph{member} (\emph{nonmember)} to refer to samples that belong (do not belong) to the training set.

A DNN classifier might perform extremely well on member samples by `memorizing' patterns in the data and any fluctuations in them due to noise. 
However, the model may not demonstrate the same performance on nonmembers. 
This behavior arises due to a phenomenon called \emph{overfitting}~\cite{hastie2009elements}, and is a major reason for the feasibility of mounting MIAs~\cite{jia2019memguard}. 
By leveraging the insight that overfitted DNNs behave differently for members and nonmembers, seminal MIAs~\cite{jia2019memguard, murphy2012machine,  shokri2017membership, yeom2018privacy} observed that DNN models produced higher confidence scores for members and lower scores for nonmembers. 
These confidence scores could then be used by an adversary to distinguish members from nonmembers. 

%

In order to thwart confidence score-based MIAs while preserving classification accuracy of the DNN model, \emph{confidence score masking}~\cite{confemasking1, ye2022one} was proposed as a defense. 
Confidence score masking defenses work by hiding true confidence values of each class returned by a DNN~\cite{confemasking1, ye2022one}. 
However, as noted in~\cite{icml2021, ccs2021}, confidence score masking is not adequate to prevent MIAs that do not rely on confidence scores for membership inference.
%

A new class of MIAs called label-based MIAs (LAB MIAs) was recently proposed in \cite{icml2021,ccs2021}. 
In carrying out a LAB MIA, an adversary only makes use of the predicted label of the DNN model without requiring knowledge of confidence scores associated to labels. 
In mounting a LAB MIA, an adversary first perturbs a data sample with additive noise of increasing magnitude 
to create new input data samples
until the newly generated sample is misclassified by the overfitted DNN model~\cite{icml2021, ccs2021}. 
Based on the magnitude of the added noise, the adversary estimates the distance of the sample to the classification decision boundary. 
Data samples farther away from the boundary are judged to be more likely to belong to the member set~\cite{icml2021,ccs2021}. 
This approach was shown to be highly effective in distinguishing between members and nonmembers in~\cite{icml2021,ccs2021}. 

\begin{figure}[h]
    \centering
     \includegraphics[scale=0.3]{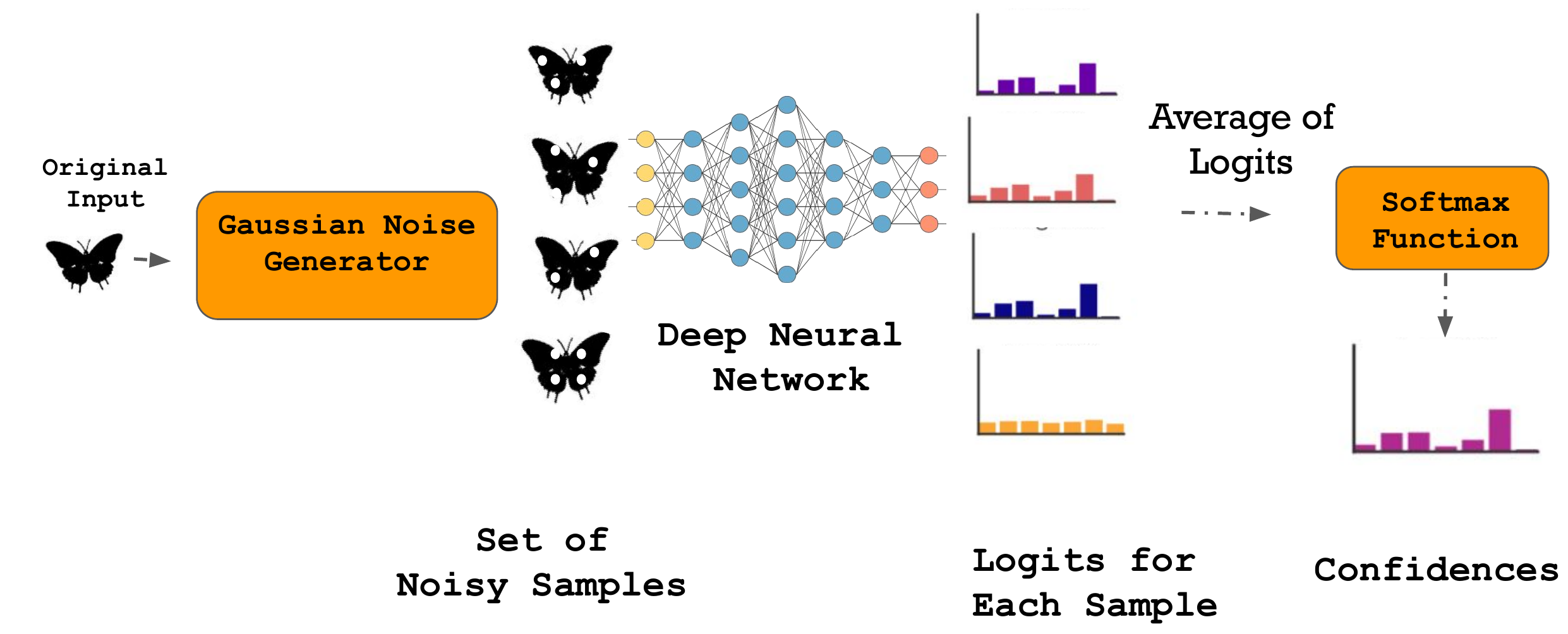} 
    
    \caption{Schematic of LDL. LDL constructs a high dimensional sphere of label invariance around each sample in a manner such that decisions of the model remain unchanged for noisy variants of the sample contained in the sphere. LDL computes the average of confidence scores obtained for each sample in the ensemble and returns the 
    label with highest confidence value.  
    }
    \label{fig:defensescheme}
\end{figure}
The demonstrated success of LAB MIAs~\cite{abadi2016deep, carlini2019secret, jia2019memguard} necessitates developing techniques to mitigate their impact. 
Since retraining DNNs is often impractical in many applications, as noted in~\cite{carlini2017adversarial}, a question is whether one can develop methods to mitigate the impact of LAB MIAs on DNN models that have already been deployed. 
We provide an affirmative answer to this question 
by proposing \textbf{LDL}, a Light weight Defense against LAB MIAs that can be incorporated into already-deployed DNN models without needing to retrain them. 
Fig.~\ref{fig:defensescheme} presents a schematic of LDL.

{\bf Our Contribution:} 
Our focus is on DNN models that perform a classification task whose objective is to identify the most relevant class out of a set of possible choices for each sample. 
We assume that an adversary can carry out a LAB MIA~\cite{icml2021,ccs2021} 
to identify samples that were used to train the model. 
We develop LDL, a defense against LAB MIAs, that does not require retraining of DNNs. 
LDL employs a randomization technique such that all data points within a high-dimensional sphere of a chosen radius parameter with the input sample at the center of the sphere return the same label. 
This type of label smoothing over the high-dimensional sphere 
aims to prevent an adversary carrying out a LAB MIA from identifying whether the sample is a member or not. 
%
%
Specifically, our contributions are: 
%
\begin{itemize}
\item 
We provide an analytical characterization of the effectiveness of an adversary carrying out a LAB MIA when LDL is deployed, and examine how it varies with 
the magnitude of noise that will be required to misclassify samples. 


\item We estimate the magnitude of perturbation required to misclassify samples 
using adversarial noise learning methods~\cite{goodfellow2014explaining}. 
We observe 
that deploying LDL ensures that misclassification rates accomplished by the addition of adversarial noise is similar for both members (training dataset) and nonmembers (testing dataset), making it difficult for an adversary to successfully carry out a LAB MIA. 

\item We evaluate LDL against state-of-the-art (SOTA) LAB MIAs~\cite{icml2021, ccs2021} on four image datasets-- CIFAR-10, CIFAR-100, GTSRB, Face-- and three non-image datasets-- Purchase, Texas, and Location. 
We consider two distinct adversary types: (i) a \emph{weak adversary} that only has knowledge of feature values of samples in its possession,  
and (ii) a \emph{strong adversary} that has knowledge of both, sample features and their corresponding labels. 
In each case, we observe that using LDL results in a significant reduction in the attack success rate of LAB MIAs.

\item We empirically demonstrate that despite not needing to retrain the DNN, using LDL achieves classification accuracy and adversary success rates that are comparable to other defenses against LAB MIAs~\cite{abadi2016deep, nasr2018machine, srivastava2014dropout} which require that the DNN model be retrained. 
\end{itemize}

The rest of the paper is organized as follows:  Section~\ref{sec:preliminaries} provides background on DNNs, MIAs, and metrics to evaluate the attack performance. Section~\ref{sec:systemmodels} describes the adversary model and gives an overview of our defense strategy. Section~\ref{sec:approach} introduces LDL, a defense against LAB MIAs that obviates a need to retrain models. Sections~\ref{sec:evaluation} and \ref{sec:results} discuss evaluation of LDL and reports our experimental results. 
Section~\ref{sec:discussion} discusses extensions of LDL to other domains and possible future directions of research. Section~\ref{sec:relatedwork} summarizes related work and Section~\ref{sec:conclusion} concludes the paper.

\section{Background}
\label{sec:preliminaries}

In this section, we provide the necessary background on DNNs, MIAs, and metrics used to evaluate attack performance.

\noindent{\bf Deep Neural Network (DNN) Classifiers:}  DNN classifiers  take an input $x_i$, and return a weight  for each class ($Z_i=[z_{i1},z_{i2},\cdots, z_{ic}]$ called {\it logits}, where $c$ is the number of classes).  
The $\texttt{softmax}$ function is used to translate these weights to probabilities, where:
\begin{equation}
    \texttt{softmax} (Z_i) = \frac{1}{\sum_k e^{z_{ik}}}[e^{z_{i1}},e^{z_{i2}},  \cdots, e^{z_{ic}}].
    \label{eq:softmax}
\end{equation}
The probability/confidence that a sample $x_i$ belongs to class $j$ is $p_{ij}= \frac{e^{z_{ij}}}{\sum_k e^{z_{ik}}}$. We call the class with highest confidence the decision of DNN classifier.  Let $Y^*_i$ be the true label of $x_i$ \ie $Y_i^*= [y_{i1}, y_{i2}, \cdots,y_{ic}]$ and $y_{ij}=1$  if $x_i$ belongs to class $j$, otherwise $y_{ij}=0$.
DNN classifiers 
return a non-zero probability for each class 
.  An input is assigned to the class with highest confidence value \ie $x_i$ is assigned to class $k$ if and only if $p_{ik}> p_{ij} \forall j\in \{1,\cdots,k-1,k+1,\cdots, c\}$.
Loss functions quantify the difference between the ideal output ($Y_i^*$) and the observed output ($P_i=[p_{i1},\cdots,p_{ic}]$). 
The cross entropy loss is widely used in ML and is defined as:
\begin{equation}
\mathit{l_{CE}}(P_i,Y_i^*)=-\sum_{j=1}^c y_{ij} \log p_{ij}.
\label{eq:celoss}
\end{equation}
%
A DNN classifier is called \emph{overfitted} if the model classifies members (samples in the training set) correctly with high confidence while having lower accuracy when identifying nonmembers~\cite{shokri2017membership}. 
Consequently, loss values for members would be low, while those for nonmembers will be high. 
In~\cite{jia2019memguard}, it was shown that overfitting of DNNs during training makes the model vulnerable to MIAs.


\noindent {\bf Membership Inference Attacks (MIAs):} MIAs determine whether a sample ($x$) was a member of the training set used to learn a DNN model ($\mathcal{M}$). The effectiveness 
of an adversary depends on the knowledge ($\mathcal{\phi}$) about the model \eg the hyper-parameters~\cite{carlini2019secret}. 
The adversary uses a function $\mathcal{A}$ that maps an input sample $x$ to  $\{0,1\}$, \ie the function returns $s=1$ if it decides that a given sample $x$ belonged to the training set (member), and $s=0$ otherwise. 

\noindent{\bf Metrics:}
We assume that samples belonging (not belonging) to the member set, denoted $S_m$ ($S_m'$), have membership label $s=1$ ($s=0$). 
MIAs aim to predict the membership label ($s_i$) of a sample ($x_i$).

\noindent{\it True Positive Rate (TPR) and True Negative Rate (TNR)}  indicate correct classification rates for member and nonmember samples, 
and are defined as ${TPR}=\Pr \{s=1;x\in S_m\}$ and ${TNR}=\Pr\{s=0;x\in S_m'\}$ respectively. 

\noindent{\it False Positive Rate (FPR) and False Negative Rate (FNR)}  indicate incorrect classification rates for nonmember and member samples, and are defined as ${FPR}=\Pr\{s=1;x\in S_m'\}$ and ${FNR}=\Pr\{s=0;x\in S_m\}$ respectively.

\noindent{\it Attack Success Rate (ASR)} measures the average accuracy of the MIA in distinguishing members from nonmembers, which is defined as:
\begin{equation}
    {ASR}= \frac{TPR+TNR}{2}.
    \label{eq:acc}
\end{equation}




\section{System Model}
\label{sec:systemmodels}
In this section we describe overfitted DNN models and label-based MIAs (LAB MIAs), then present the adversary model and provide an overview of our defense methodology. 

%

\subsection{Overfitted DNN Models}\label{subsec:overfitted}
Overfitting in DNN models can occur when member (training or seen) samples are learned 
extremely well by `memorizing' patterns in the data and noise fluctuations~\cite{hastie2009elements}. As a consequence, the model will be unable to classify nonmember (test or unseen) samples effectively. 
Overfitting is a major reason for the feasibility of MIAs. 
It has been shown in~\cite{yeom2018privacy} that an 
adversary will be able to use loss or confidence values to effectively distinguish between member and nonmember samples of an overfitted model. 
 
One method to quantify the information obtained by an adversary carrying out a MIA on overfitted DNN models 
is to evaluate a \emph{gap attack}, as defined in~\cite{yeom2018privacy, icml2021}. 
The gap attack is a MIA that infers membership using the fact that for an \emph{overfitted} model, a sample belonging to the training dataset is classified correctly with high confidence, whereas a nonmember sample is likely to be misclassified with high confidence~\cite{icml2021}. 
The $ASR$ of an adversary carrying out a gap attack is denoted $ASR_{gap}$. 
We note that $TPR$ in Eqn. (\ref{eq:acc}) is the accuracy of classification of member data samples, and denoted as $Acc_{mem}$. 
For nonmmembe samples, classification accuracy of the DNN is denoted as $Acc_{nonmem}$. 
When a nonmember sample provided as input to the DNN by the adversary is correctly classified, the adversary is led to conclude, albeit incorrectly, that this sample belongs to the member set. 
Hence, the $TNR$ in Eqn. (\ref{eq:acc}) is given by $1-Acc_{nonmem}$. 
Substituting these values in Eqn. (\ref{eq:acc}) yields the expression given in~\cite{yeom2018privacy, icml2021}:
\begin{equation}
{ASR}_{gap}=0.5+ \frac{Acc_{mem}-Acc_{nonmem}}{2}=0.5+ \frac{TPR-FPR}{2}
   \label{eq:asrmin}
\end{equation}

It has been noted in~\cite{icml2021} that the gap attack overestimates the value of $Acc_{nonmem}$ (since any misclassified sample is identified as a nonmember). 
Carrying out a more deliberate MIA, such as a LAB MIA, will typically enable an adversary to achieve 
$ASR \geq  ASR_{gap}$ through additional queries to the DNN model to obtain a better estimate of $Acc_{nonmem}$. 
The objective of any defense against MIAs is to 
reduce the value of $ASR$  
to the minimum possible, i.e., $\max\{ASR_{gap}, 0.5\}$. 
 In this paper, we 
 examine the impact of deploying LDL on the ASR of an adversary carrying out a LAB MIA on an overfitted DNN model with varying sizes of training datasets. 



\subsection{Label-Based MIAs (LAB MIAs) on DNNs}\label{sec:label-basedMIA}
Given the final label of a sample output by a DNN model, an adversary carrying out a LAB MIA~\cite{icml2021,ccs2021} uses 
adversarial learning methods~\cite{chen2020hopskipjumpattack} to estimate the amount of 
adversarial noise needed to be added to the sample to make the DNN misclassify the sample. 
The magnitude of noise will enable distinguishing members from nonmembers, as explained next. 

An overfitted DNN model will classify member samples with high confidence, since these samples are far away from a decision boundary.
For example, for a classifier that uses a binary logistic regression model $\rho(x):= \frac{1}{1+e^{-(\omega^T x +b)}}$, the distance of a sample $x$ to the decision boundary is $(\omega^T x +b)/\|\omega\|_2$.
Then, samples with higher confidence will have a larger distance to the decision boundary (since {$(\omega^T x + b) \rightarrow \infty \Rightarrow \rho(x)\rightarrow 1$}). 
As shown in~\cite{icml2021}, computing this distance will be equivalent to determining the smallest magnitude of additive noise required to make the DNN misclassify the sample. 
Member samples will require a larger magnitude of noise for misclassification compared to nonmember samples. 
It was noted in~\cite{icml2021, ccs2021} that LAB MIAs only required the final sample label, and did not require knowledge of confidence values associated to labels.

\subsection{Adversary Model for LAB MIAs}\label{subsec:AdvModel}
We follow the setting considered in~\cite{icml2021, ccs2021}, and assume that the adversary has black-box model access, 
and that the DNN model only returns a label for a given input. 
Thus, model hyper-parameters are not not known to the adversary, but the adversary can obtain the output for any sample provided to the model. 
The adversary aims to determine whether a sample belongs to the training set of a DNN model ($s_i=1$ if $x_i \in S_m$, otherwise $s_i=0$) by analyzing outputs of the model. 
Similar to~\cite{ccs2021}, we consider two types of adversaries: 
(i) an adversary with access to a small subset of member and nonmember samples. These samples can be used to assess the behavior of the DNN model by examining the magnitude of noise that needs to be added 
in order for the model to misclassify them; and (ii) an adversary that does not have any knowledge about samples used to train the DNN model. In this case, as noted in~\cite{ccs2021}, the LAB MIA adversary uses substitute models to estimate the magnitude of noise required to misclassify members and nonmembers of the substitute model. 

\noindent \textbf{Weak vs. Strong adversary:} 
Any adversary carrying out a LAB MIA will have full knowledge of features of samples that are in its possession. 
The adversary examines the output of the DNN model for a sample that it perturbed with additive noise. 
A \emph{strong} adversary has knowledge of sample features and the corresponding label of the sample. 
A \emph{weak} adversary only has knowledge of features of samples in its possession. 
The weak adversary will be able to obtain the label associated to the sample by examining the output of the DNN when the sample is given as input. 
The procedure to carry out a LAB MIA will then be identical for both types of adversaries. 
\subsection{Defense Strategy Against LAB MIAs}

We assume that the owner of the DNN model uses a defense mechanism to mitigate information leakage about samples used to train this model. The objective of the defense strategy is to ensure that the amount of perturbation required to misclassify members and nonmembers will be of a comparable magnitude. 
As noted in~\cite{icml2021, ccs2021}, 
the owner of the DNN model does not have any additional computational resources beyond those used for model inference, and will not be able to retrain the DNN model. 

In this paper we propose LDL, a light weight defense against LAB MIAs. LDL works by constructing a high-dimensional sphere around queried samples such that the model decision is unchanged for (noisy) variants of the sample within the sphere. 
%
Using LDL as a defense 
does not require retraining the DNN model or additional computational resources beyond those needed for inference.
We describe LDL in detail in Section~\ref{sec:approach}. 

%



\section{Our Proposed Defense}
\label{sec:approach}

In this section, we first describe the insight behind our method and then describe the implementation of LDL in detail. 

\subsection{Description of LDL}\label{sec:intution}
The objective of LDL is to ensure that magnitudes of noise required for misclassification of members and nonmembers are comparable. 
This will ensure indistinguishability of members and nonmembers to an adversary carrying out a LAB MIA. 
LDL uses randomization at inference time and constructs a label invariant sphere around samples.  
Following~\cite{cohen2019certified}, we term a classifier $f$ with randomization as a \emph{smoothed} classifier $g$.
The smoothed classifier $g$ returns the class that $f$ will most likely return for \emph{perturbed} variants of $x$ (as opposed to the original sample $x$).
Although the smoothed classifier $g$ may not return the true label of $x$,  it is guaranteed to return the same label for all $x'$ inside a sphere of radius $r$ centered at $x$~\cite{cohen2019certified}.
LDL achieves a reduction in the attack success rate of LAB MIAs that use additive noise to infer whether a sample belongs to the training set (member) or not (nonmember).

We note that the term $ASR_{gap}$ in Eqn. \eqref{eq:asrmin} is specifically 
defined for the gap attack~\cite{icml2021}, and will not be adequate to quantify the success rates of other classes of MIAs. 
Moreover, Eqn. \eqref{eq:asrmin} does not capture the effectiveness of defense mechanisms that might be used by a DNN model to mitigate the impact of a MIA. 
Therefore, the quantification of ASR when using a defense mechanism such as LDL should capture the following properties: 
(i) when LDL is not deployed by the DNN model, it should precisely model the attack success rate, 
(ii) when LDL is deployed, it should express the ASR as a function of randomization parameters, 
%
(iii) as the amount of randomization used by LDL increases, the attack success rate should approach $50\%$, 
since its strategy will then reduce to a fair coin-toss, and 
(iv) when LDL is not deployed as a defense and the adversary carries out a gap attack, we should recover Eqn. (\ref{eq:asrmin}) that was presented in~\cite{icml2021}. 
Following the above criteria, we propose that the attack success rate (in Eqn. (\ref{eq:asrmin})) when LDL is deployed takes the modified form: 
\begin{equation}
    ASR_{LDL} = 0.5 + (1-h(\sigma^2))\frac{{TPR} - {FPR}}{2},\label{eqn:asrldl}
\end{equation}
where ${TPR}$ and ${FPR}$ are the adversary's true positive rate and false positive rates as defined in Sec. \ref{sec:preliminaries}. Here, $h(\sigma^2)\in[0,1]$ is a function that has the following properties: 
(a) $h(\sigma^2)=0$ if $\sigma=0$, 
(b) $h(\sigma^2)$ is monotone increasing with respect to $\sigma$, 
(c) $h(\sigma^2)=1$ as $\sigma\rightarrow\infty$, and 
(d) if $h(\sigma^2)=0$ and ${TPR} - {FPR}=Acc_{mem}-Acc_{nonmem}$, $ASR_{LDL}$ reduces to $ASR_{gap}$. 
These properties model all the criteria (i) - (iv) identified above.  

Moreover, when LDL is deployed with $\sigma^2>0$, we observe that $ASR_{LDL}<ASR_{gap}$, since: 
(i) as $\sigma^2$ increases, the TPR of the adversary decreases while the FPR increases, and 
(ii) $h(\sigma^2)\geq 0$ for $\sigma^2>0$. 
This formulation is adequate to quantify the ASR of a wide range of MIAs, depending on the function $h(\sigma^2)$. 
Two pertinent choices 
that we will use in our experimentas in Sec.~\ref{sec:results} are $h_1(\sigma^2):=1-\exp{(-\alpha\sigma^2)}$, and $h_2(\sigma^2):= L(1+\exp{(-\beta(\sigma^2-c)))^{-1}}$, where $\alpha, \beta, L, c > 0$.

\subsection{Implementation of LDL}\label{sec:ldlimplementation}

As noted in~\cite{icml2021, ccs2021} and from the example in 
Sec.~\ref{sec:label-basedMIA}, samples with lower confidence will be located closer to a decision boundary of a DNN model. 
An adversary carrying out a LAB MIA can either use learning methods that require only black-box access to the model~\cite{chen2020hopskipjumpattack} or add an arbitrary amount of noise (\emph{random}) to the input data sample to estimate the distance to the decision boundary. 
In each case, LDL constructs a sphere centered at a sample in a manner that the model returns the same label for all perturbed variants of the sample contained in the sphere. 
This aims to create an ambiguity for the adversary who tries to distinguish between members and nonmembers. 
%
To construct the \emph{label-invariant} sphere around samples, LDL uses a smoothed classifier $g$ that aims to return the same label for 
variants of $x$ that satisfy 
($x+\delta, \:\: \|\delta\|_2\leq r$). 
We assume that a zero-mean Gaussian noise with variance $\sigma^2$ is used to perturb $x$. 
Algorithm~\ref{alg:noisyinputs} illustrates the working of LDL.  
For each input $x$, we obtain $K$ perturbed variants of $x$ 
and determine the logit of the model for each perturbed sample. We then calculate the average of the logits (\emph{ensemble of decisions}). 
We compute confidence scores for each class through a softmax function. 
Using a smoothed classifier increases the amount of noise required to change a decision of the original model for members and nonmembers. 
As a result, an adversary will not be able to accurately estimate the magnitude of the perturbation required for misclassification. 
 

\begin{algorithm}
\caption{LDL: Light weight Defense against LAB MIAs}\label{alg:noisyinputs}
\begin{algorithmic}[1]
\Require $x, f, {\sigma}^2 $;

\ForAll {$i=1:T$}
    \State $n_i \sim \mathcal{N}(0,\sigma^2 I); \:\:\: x_{noisy} \gets x+n_i$;
    \State predictions $ \gets$ predictions $\cup  f(x_{noisy})$;
\EndFor
\State prediction $\gets$ average(predictions);
\State   return prediction;
\end{algorithmic}
\end{algorithm} 
%

\begin{figure}[t]
    \centering
     \includegraphics[width=0.3\textwidth]{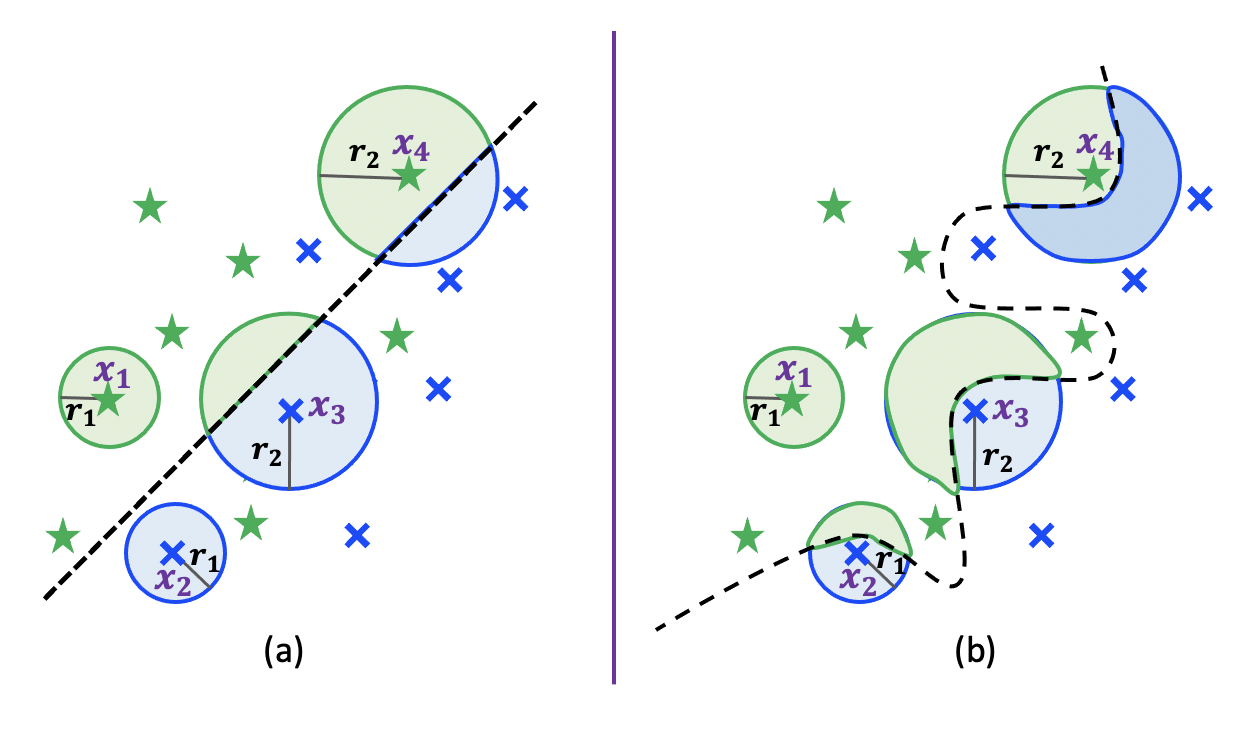} 
    \caption{ 
    Representation of high-dimensional spheres of label invariance of different radii on normal and overfitted models. 
    The figure shows two classes (green and blue) separated by a decision boundary (dashed black line). 
    In Fig. 2(a), we see that spheres of radius $r_1$ or $r_2$ ($>r_1$) result in correct classification of samples (\eg $x_3$ is classified to the blue class since area of the blue region within the sphere centered at $x_3$ is larger than that of the green region.). Thus, an adversary will require a perturbation $>r_2$ for the model to misclassify samples.  
    The shape of the decision boundary in Fig. 2(b) indicates an overfitted model. In this case, even a smaller value of $r$ (i.e., $r_1$) can result in samples being misclassified.}
    \label{fig:noisefigure}
    \vspace{-5mm}
\end{figure}

Figure~\ref{fig:noisefigure}(a) illustrates that a larger radius increases robustness of samples to noise. 
For example, samples in the sphere of radius $r_2$ around $x_3$ mostly belong to the blue class. 
LDL thus classifies $x_3$ and any perturbed variant of $x_3$ within this sphere to the blue class. 
As a result, an adversary will require a perturbation larger than $r_2$ for the model to misclassify $x_3$. 
However, when an adequate decision boundary is not learned, such as for overfitted DNNs, even a small amount of perturbation ($r_1 < r_2$) can result in misclassification of samples by the DNN. This is illustrated in Fig.~\ref{fig:noisefigure}(b), where samples $x_3$ and $x_4$, which would otherwise be correctly classified by the model, will be misclassified after deploying LDL with radius $r_2$. 
The radius of the label-invariant sphere is related 
to the value of the parameter $\sigma^2$. 
This has the following consequence. From Sec.~\ref{sec:intution} and Eqn.~(\ref{eqn:asrldl}), an increase in $\sigma^2$, and correspondingly, the radius of the sphere, will cause a reduction in the value of $ASR_{LDL}$. 
\section{Evaluation of LDL Performance}\label{sec:evaluation}
In this section, we describe the settings under which we evaluate LDL, and describe the datasets that we use. 

\subsection{Evaluation Setup}\label{subec:setup}
We experimentally evaluate the performance of LDL against SOTA LAB MIAs~\cite{icml2021, ccs2021}. In each case, for any input, the DNN model only returns the label with the highest confidence. 
LAB MIAs estimate the amount of noise required to make the DNN model change its decision in order to distinguish between members and nonmembers. 
When the DNN model does not incorporate/deploy any defense against a LAB MIA, we call it a \emph{defense-free model}.  
We compare the success rate of an adversary ($ASR$) carrying out a LAB MIA on a defense-free model and in the case when LDL is used as a defense.

We evaluate LDL against a weak and a strong adversary defined in Sec.~\ref{subsec:AdvModel}. As noted in Sec.~\ref{subsec:AdvModel}, a
strong adversary has knowledge of true labels. 
For a strong adversary, we hypothesize that the value of $ASR$ when LDL is deployed will be close to $ASR_{gap}$, since it has knowledge of values of $TPR$ and $FPR$ (Eqn. (\ref{eqn:asrldl}). 
Similarly, from Sec.~\ref{subsec:AdvModel}, a weak adversary has knowledge of only feature values of samples. 
For a weak adversary, we hypothesize that using LDL will reduce the $ASR$ of a LAB MIA to $50\%$, since a sample will be identified as a member or a nonmember from the outcome of a fair coin toss. 
As described in Sec. \ref{subsec:AdvModel}, we consider two settings: 
(i) when the adversary has a small set of member and nonmember samples~\cite{ccs2021}, 
it learns the optimum value of a threshold that separates members from nonmembers. 
(ii) when the adversary has no knowledge about which samples may have been used to train the DNN model, it locally learns a \emph{substitute model}~\cite{icml2021} to estimate the minimum amount of noise required to misclassify samples. 
The amount of noise required to be added to a sample to trigger misclassification 
can be estimated using adversarial learning methods (adversarial noise) as shown in
~\cite{chen2020hopskipjumpattack}, or by generating perturbed variants of each sample (random noise), as in~\cite{icml2021}. 

\noindent {\bf Adversarial Noise}: 
Since the adversary has only black-box access to the DNN model, 
it uses query-based methods to learn the minimum amount of noise required to cause the DNN model to misclassify the sample. 
In generating adversarial noise, to maintain consistency with~\cite{icml2021, ccs2021}, we use 
a query-based adversarial learning method called \emph{HopSkipJump}~\cite{chen2020hopskipjumpattack} to estimate the minimum magnitude of noise perturbation that has to be added to the input sample point for misclassification. 

\noindent {\bf Random Noise}: 
In this case, the LAB MIA adversary uses a random additive noise to estimate the distance of a sample to the decision boundary~\cite{icml2021}.
Varying amounts of noise are used to generate multiple perturbed samples and the adversary learns the smallest amount of noise required to trigger a misclassification by the DNN model.

\subsection{Datasets and Classifiers} 
We use all the datasets noted in \cite{ccs2021}, including
(i) CIFAR-10~\cite{krizhevsky2009CIFARs}, (ii) CIFAR-100~\cite{krizhevsky2009CIFARs}, (iii) GTSRB~\cite{GTSRB013}, and (iv) Face~\cite{Huang2012a} for our experiments. 
We remove classes with $<40$ images from the Face dataset. 
To train DNN models for 
classification tasks, 
we construct a convolutional neural network  with 4 convolutional layers and 4 pooling layers with 2 hidden layers, each containing at least 256 units. We train them for 200 epochs using the Adam optimizer.  

We also evaluate LDL on three non-image datasets, as noted in \cite{shokri2017membership}, which includes (i) Location, (ii) Purchase, and (iii) Texas.
Unlike image datasets that have continuous feature values, features of these three datasets take discrete values. 
We use the same settings as in~\cite{icml2021}, and train a fully connected neural network with one hidden layer of size of 128 for the Purchase dataset, a fully connected neural network with two hidden layers of size 128 and a Tanh activation function for the Location and Texas datasets.  
Our code is available on our Github page at \textcolor{blue}{\url{https://tinyurl.com/mpzb3j5y}}.

\section{Results of Experiments}\label{sec:results}

This section presents results of our experimental evaluations. 
We use the same datasets and experiment settings from SOTA LAB MIAs~\cite{icml2021, ccs2021}, and also compare LDL to other defenses against MIAs. 
In Sec.~\ref{sec:hopskipjump} and~\ref{sec:randomnoise}, we evaluate the performance of LDL against an adversary carrying out a LAB MIA using adversarial noise and random noise (described in Sec.~\ref{subec:setup}). 
In Sec.~\ref{sec:shadowmodels}, we examine the effect of deploying LDL against an adversary that uses substitute models when it does not have access to the set of member and nonmember samples. 
Section~\ref{sec:FGS} shows that deploying LDL ensures that misclassification rates accomplished by the addition of adversarial noise is similar for both members 
and nonmembers, 
making it difficult for an adversary to successfully carry out a LAB MIA. 
%
In Sec.~\ref{sec:defenses}, we demonstrate that despite not needing to retrain DNN models, LDL performs favorably in comparison to other defenses againt LAB MIAs that require DNN models to be retrained. 
We examine the consistency of observed values of $ASR_{LDL}$ with the analytical characterization from Eqn. (\ref{eqn:asrldl}) in Sec.~\ref{sec:selectingsigma}. 

\begin{figure*}
    \centering
    \begin{tabular}{>{\centering\arraybackslash} m{0.4cm} >{\centering\arraybackslash} m{8cm} >{\centering\arraybackslash} m{8cm}  }
     & {\bf Strong Adversary}&  {\bf Weak Adversary}\\
      \rotatebox{90}{ CIFAR-10} &
       \includegraphics[trim={2cm 0.5cm 2cm 2cm}, scale=0.15]{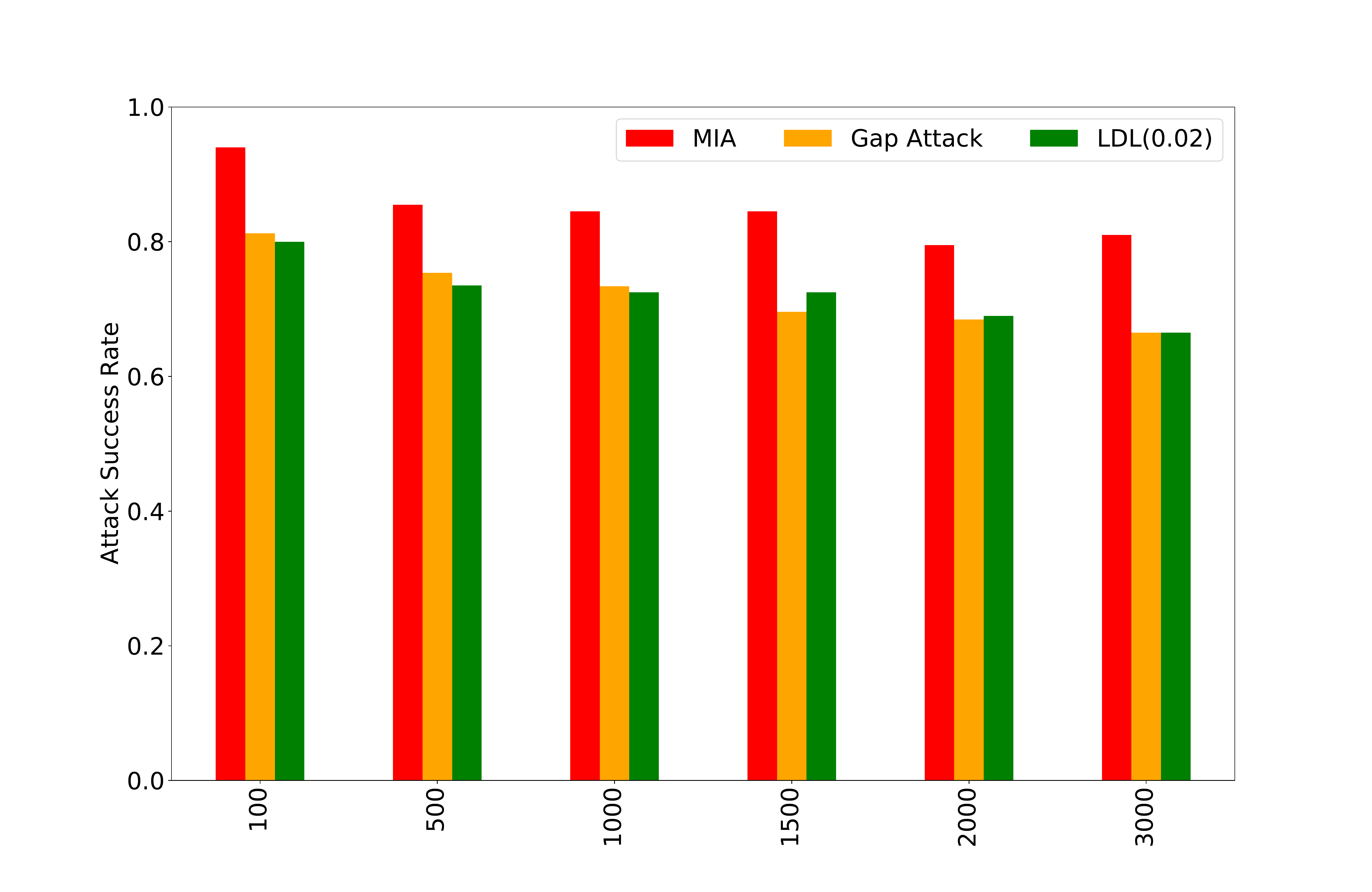} &    
       \includegraphics[trim={2cm 0.5cm 2cm 2cm}, scale=0.15]{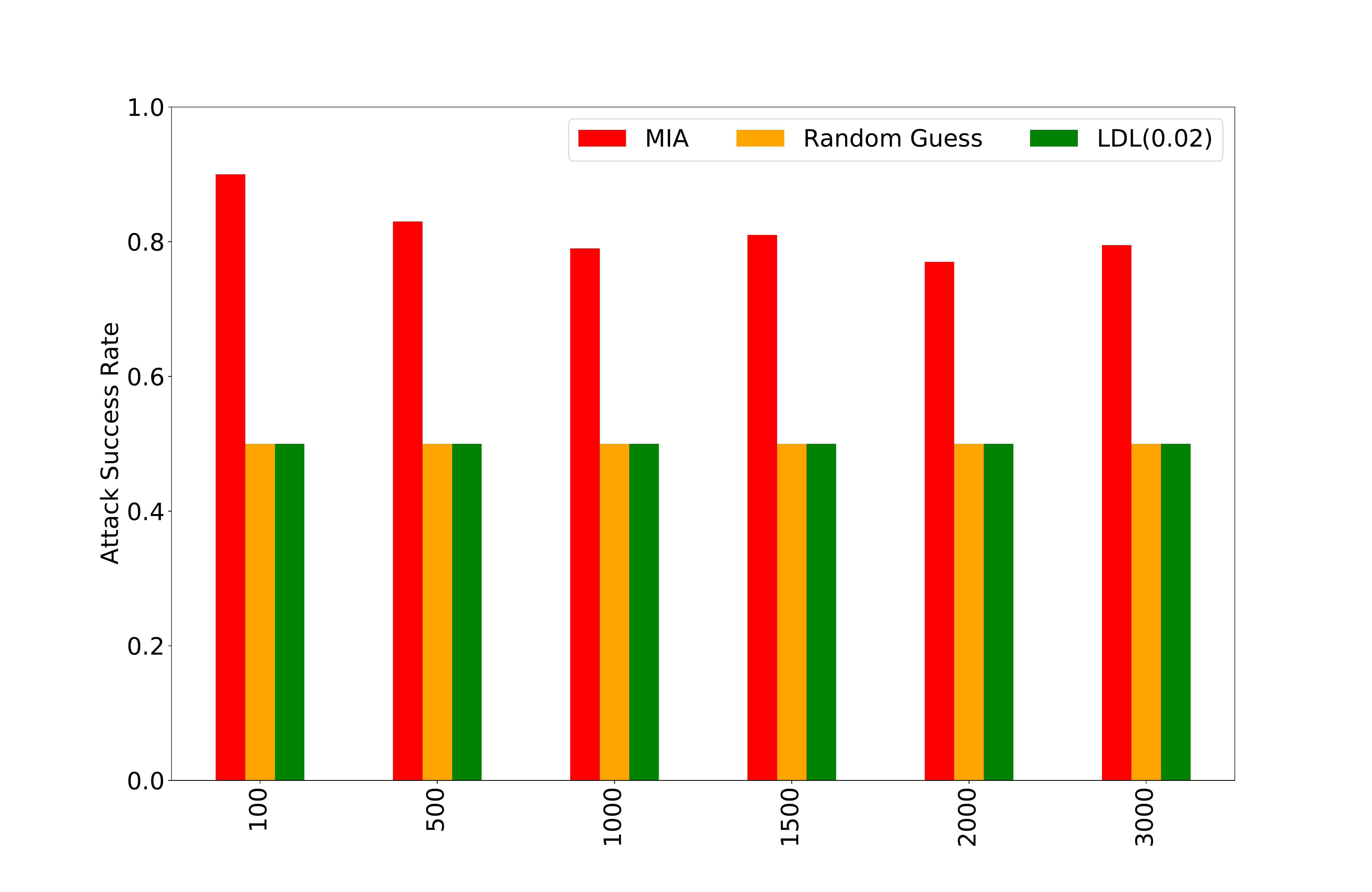}\\
        \rotatebox{90}{CIFAR-100}&
        \includegraphics[trim={2cm 0.5cm 2cm 2cm},scale=0.15]{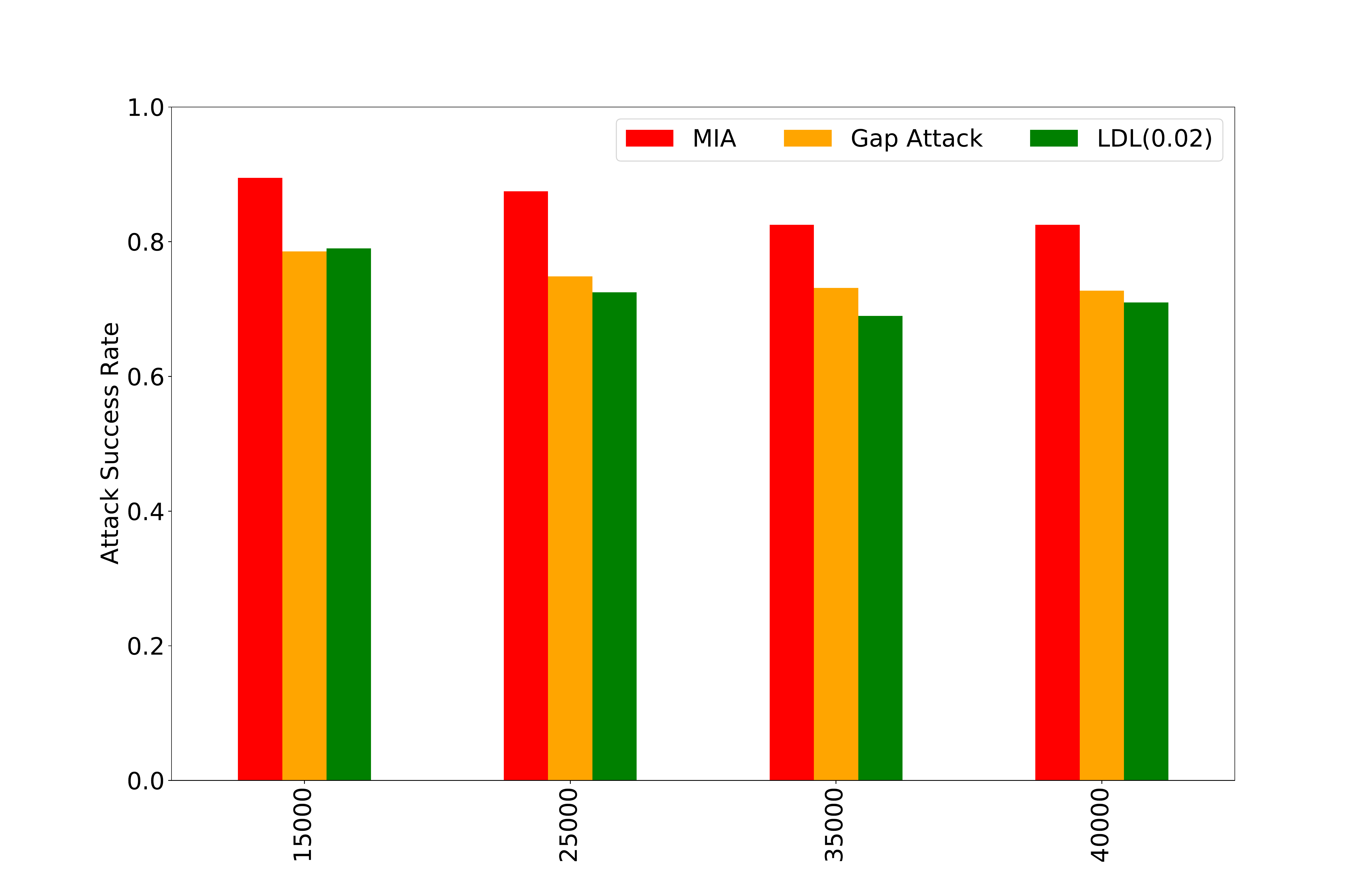}&  
        \includegraphics[trim={2cm 0.5cm 2cm 2cm},scale=0.15]{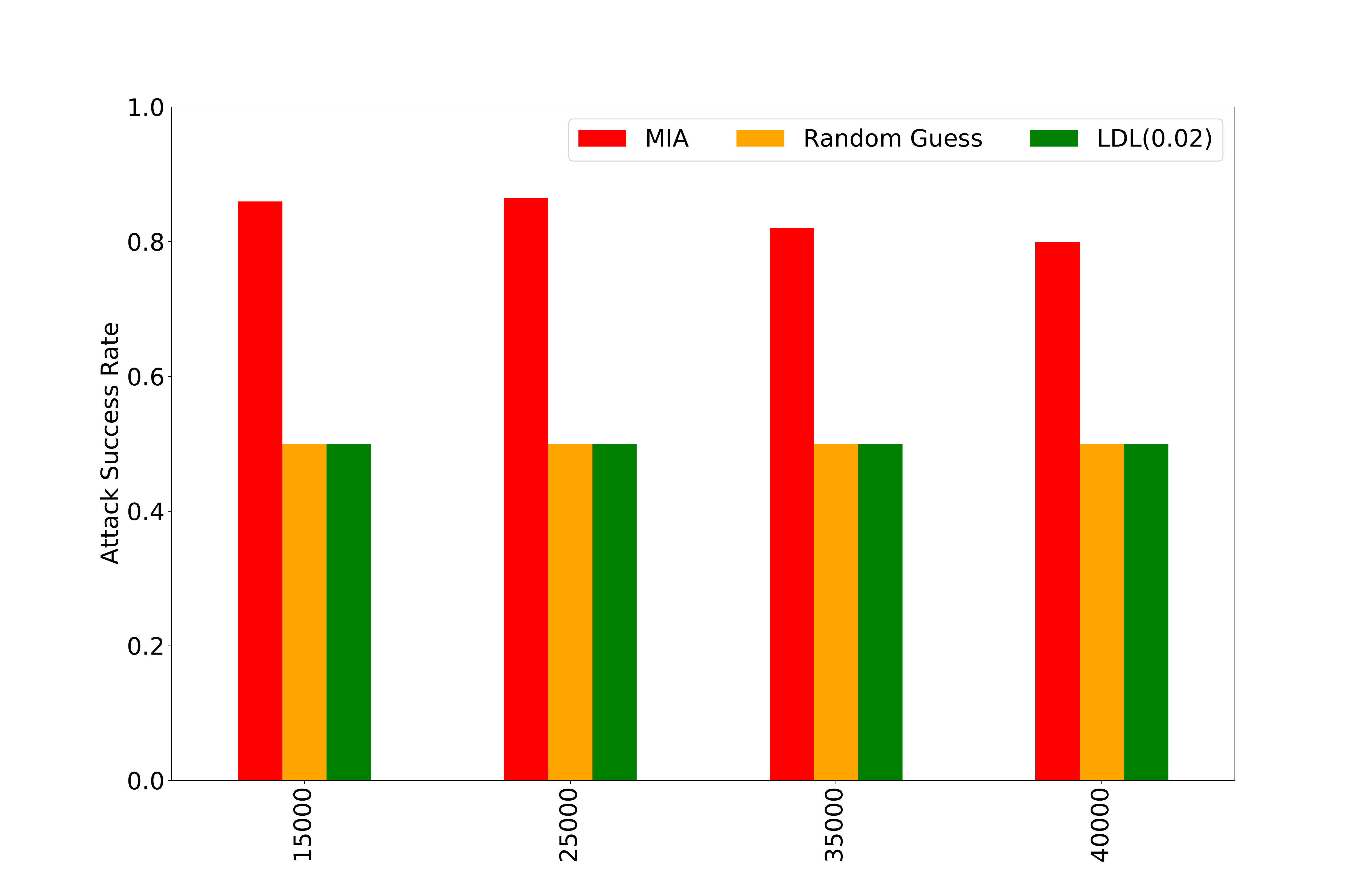}
%
    \end{tabular}
    \caption{Attack success rate (ASR) of adversary carrying out a LAB MIA using HopSkipJump adversarial noise~\cite{ccs2021} for defense-free models (red bars) and models with LDL deployed (green bars). We evaluate LDL on datasets with different sizes of training sets for a strong adversary (left column) and a weak adversary (right column). We show results for the CIFAR-10 and CIFAR-100 datasets here; results for the GTSRB and Face datasets are presented in the Appendix. 
    The value of $ASR_{gap}$ for a strong adversary carrying out a gap attack and that of a random guess for a weak adversary on a defense-free model are also shown (orange bars). 
    Deploying LDL successfully reduces the $ASR$ to a value close to $ASR_{gap}$ for a strong adversary and to $50\%$ for a weak adversary. 
    }
    \label{fig:strongadv}
\end{figure*}

\begin{figure*}[!h]
    \centering
    \begin{tabular}{  c c c c }
        \includegraphics[trim={3cm 0 5cm  2.5cm},clip, scale=0.13]{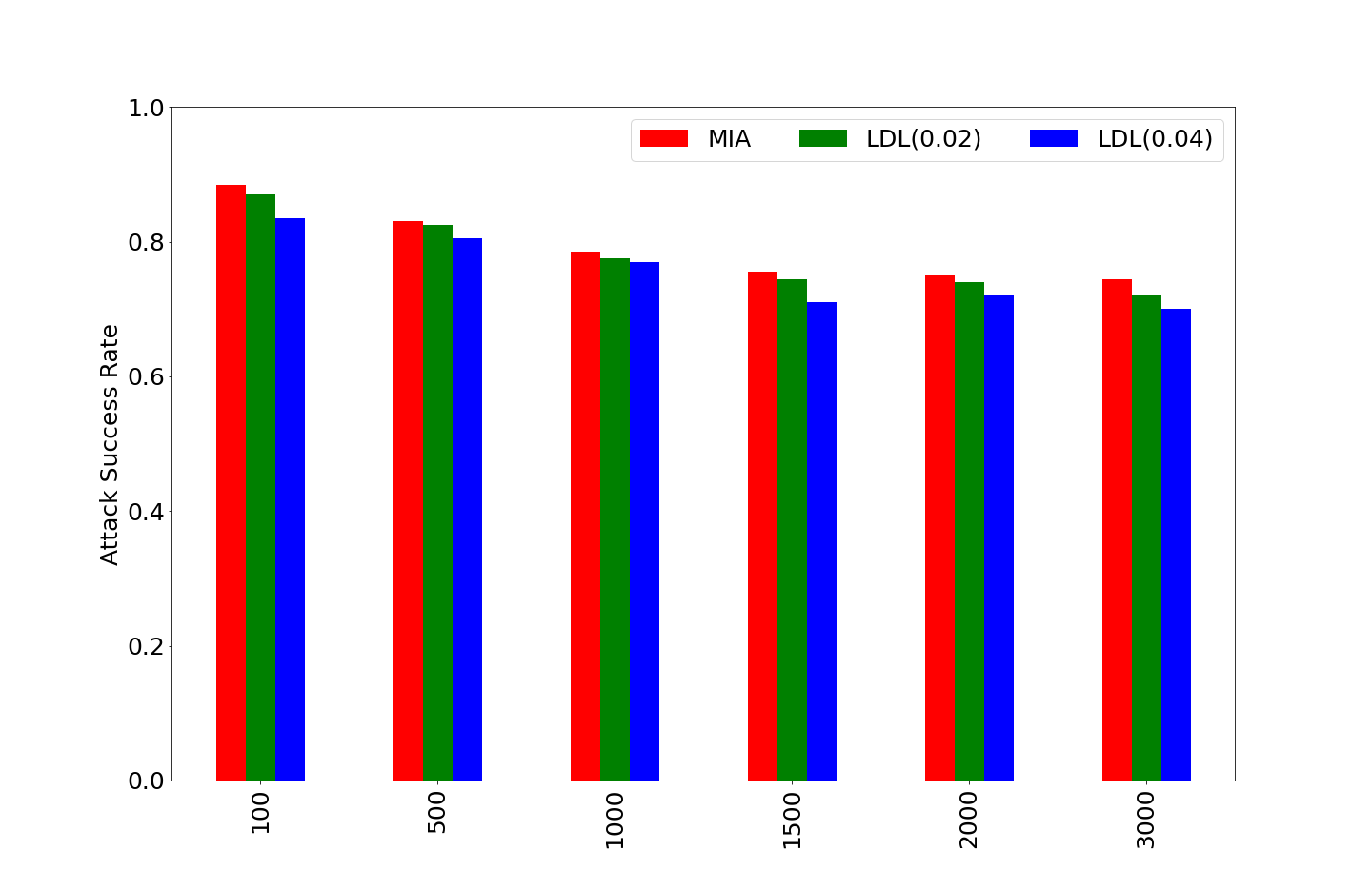} &  
        \includegraphics[trim={3cm 0 5cm  2.5cm},clip,scale=0.13]{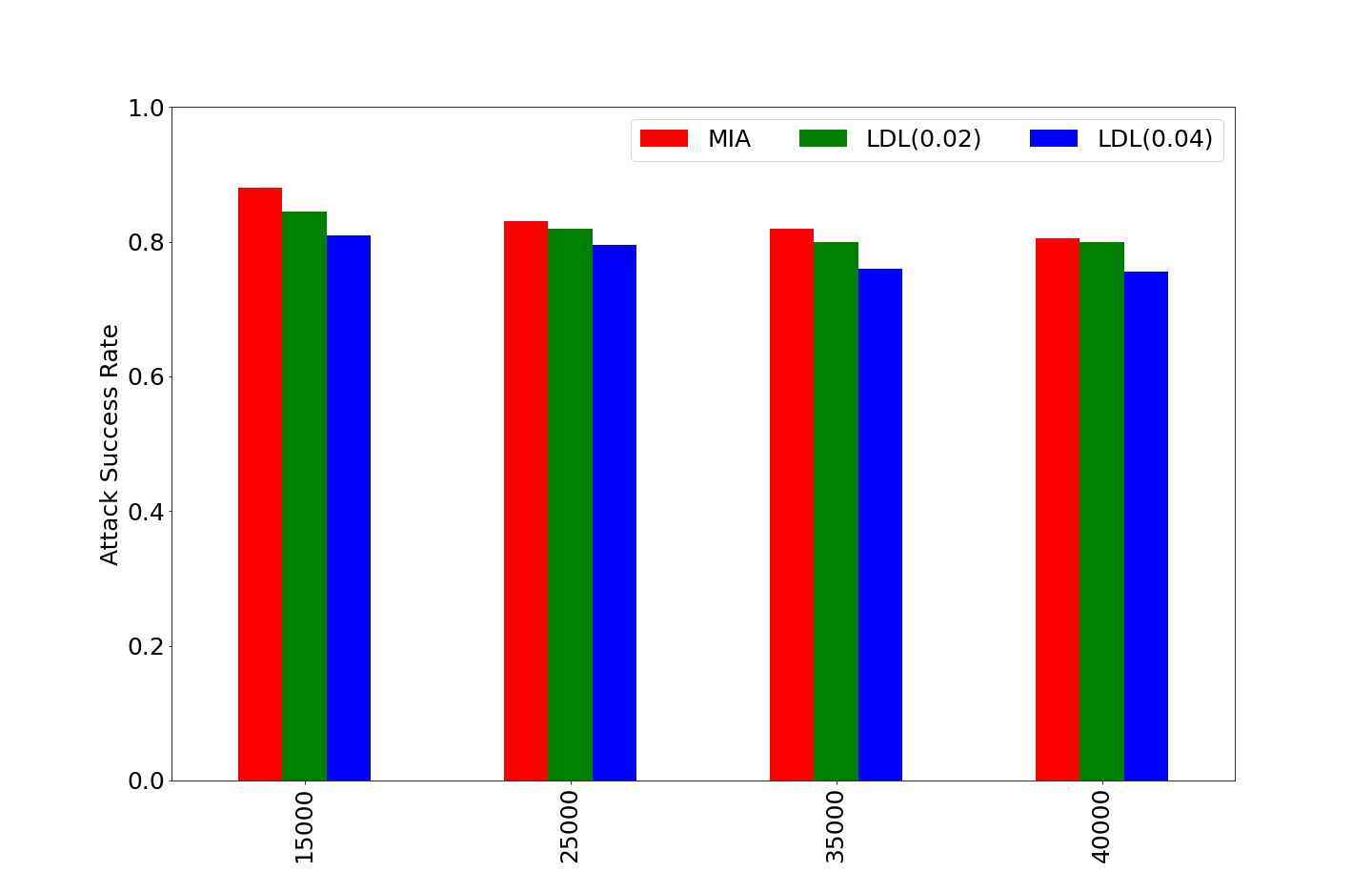}\\ 
         (a) CIFAR-10& (b) CIFAR-100 \\
         \includegraphics[trim={3cm 0 5cm  2.53cm},clip,scale=0.13]{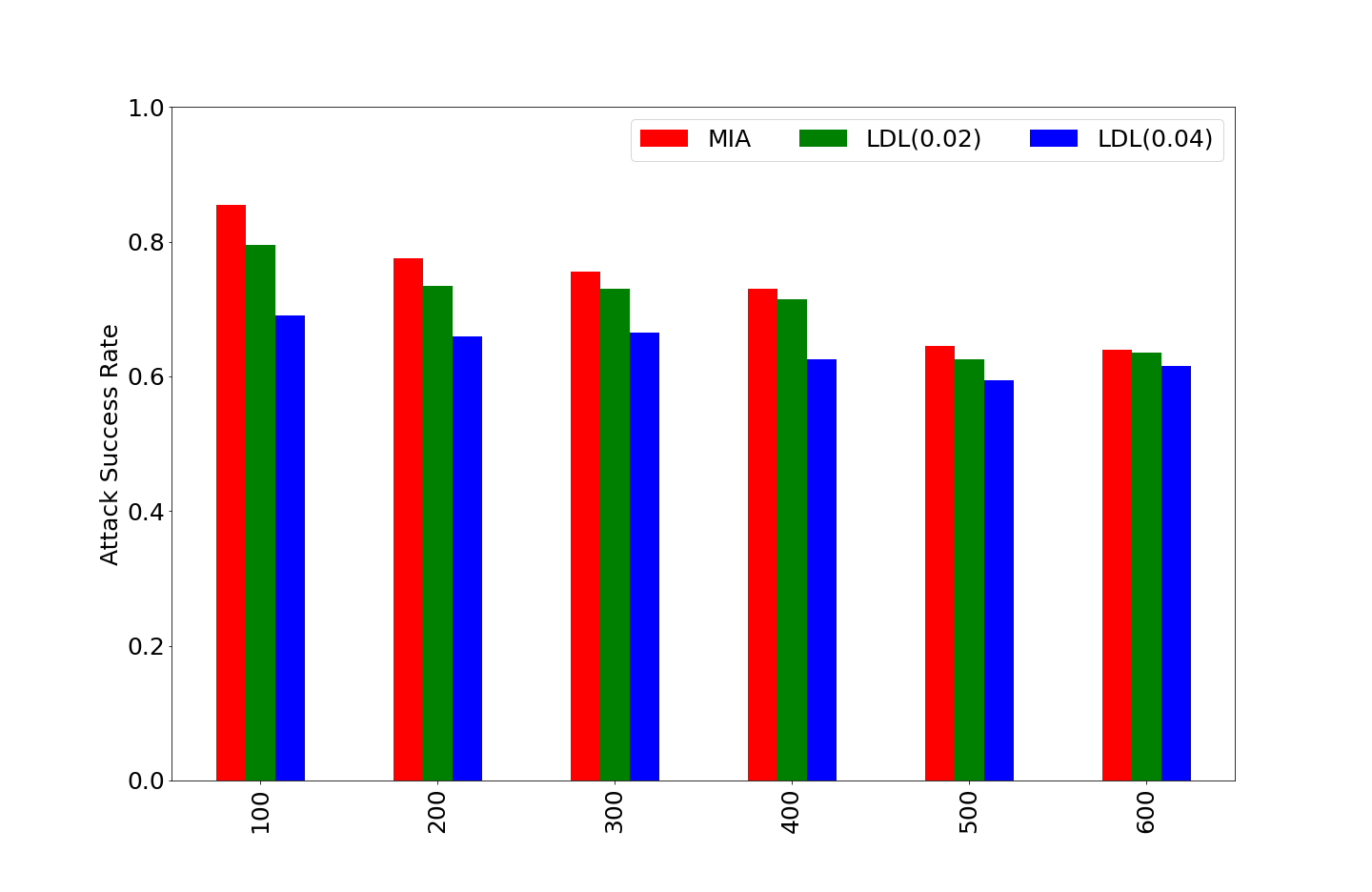}
        & \includegraphics[trim={3cm 0 5cm 2.5cm},clip,scale=0.13]{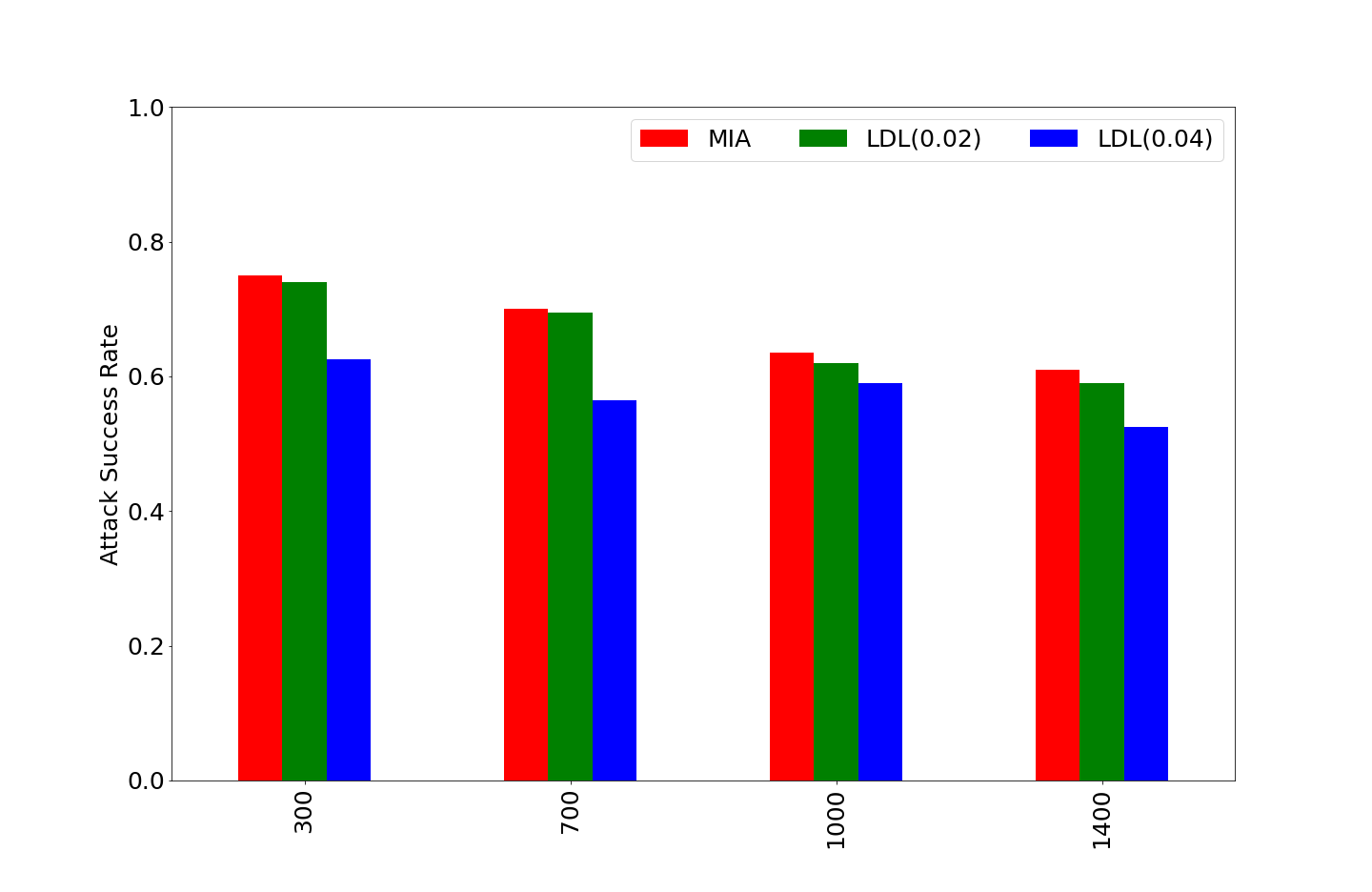}\\
      (c) GTSRB & (d) Face 
    \end{tabular}
    \caption{Success rate of a LAB MIA that uses random noise to estimate the distance of a sample to a decision boundary. 
    Values of the attack success rate ($ASR$) for a defense-free model (red bars) and model with LDL deployed using different values of $\sigma^2$ (green and blue bars) are shown. 
    Consistent with Eqn. (\ref{eqn:asrldl}), a larger value of $\sigma^2$ provides a greater reduction the value of $ASR$. 
    }
    \label{fig:randomnoise}
\end{figure*}

\begin{figure*}[!h]
    \centering
    \begin{tabular}{>{\centering\arraybackslash} m{0.14cm} >{\centering\arraybackslash} m{3.8cm} >{\centering\arraybackslash} m{3.8cm}>{\centering\arraybackslash} m{3.8cm}>{\centering\arraybackslash} m{3.8cm}}
    \rotatebox{90}{Defense-free} &   
    \includegraphics[trim={6cm 5.5cm 6cm 6cm},scale=0.18]{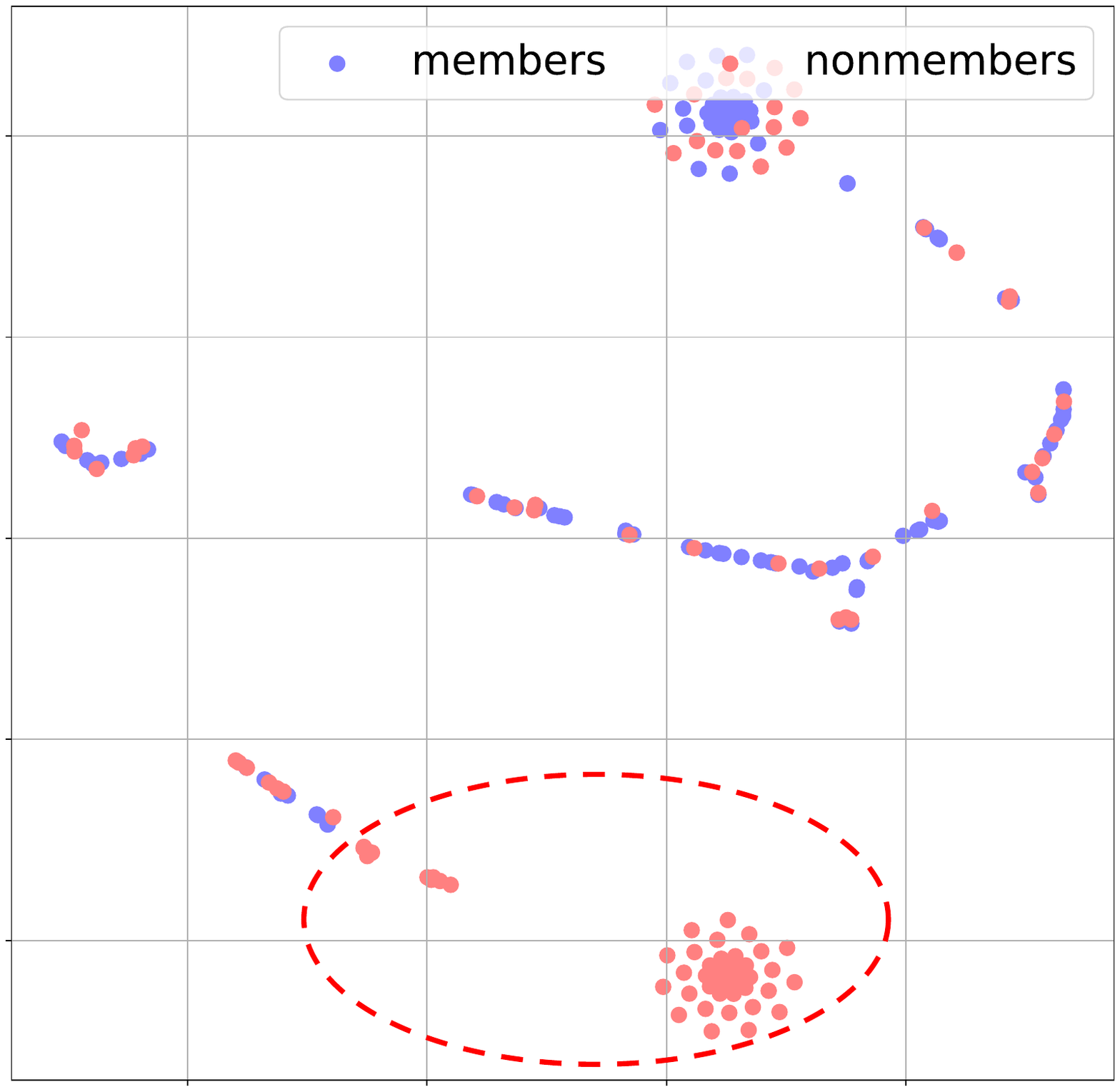}&  
        \includegraphics[trim={6cm 5.5cm 6cm 6cm},scale=0.18]{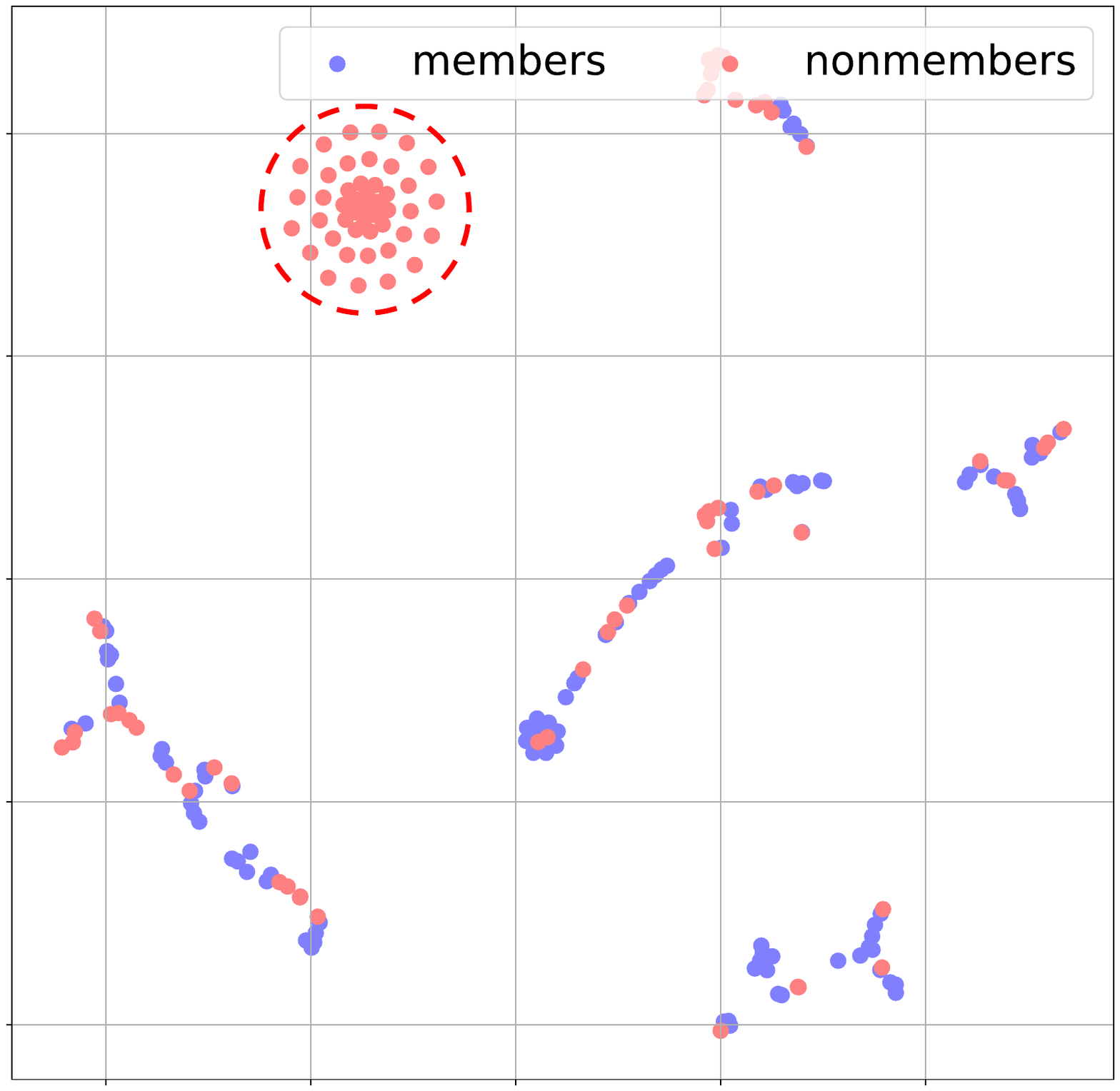}&  
        \includegraphics[trim={6cm 5.5cm 6cm 6cm},scale=0.18]{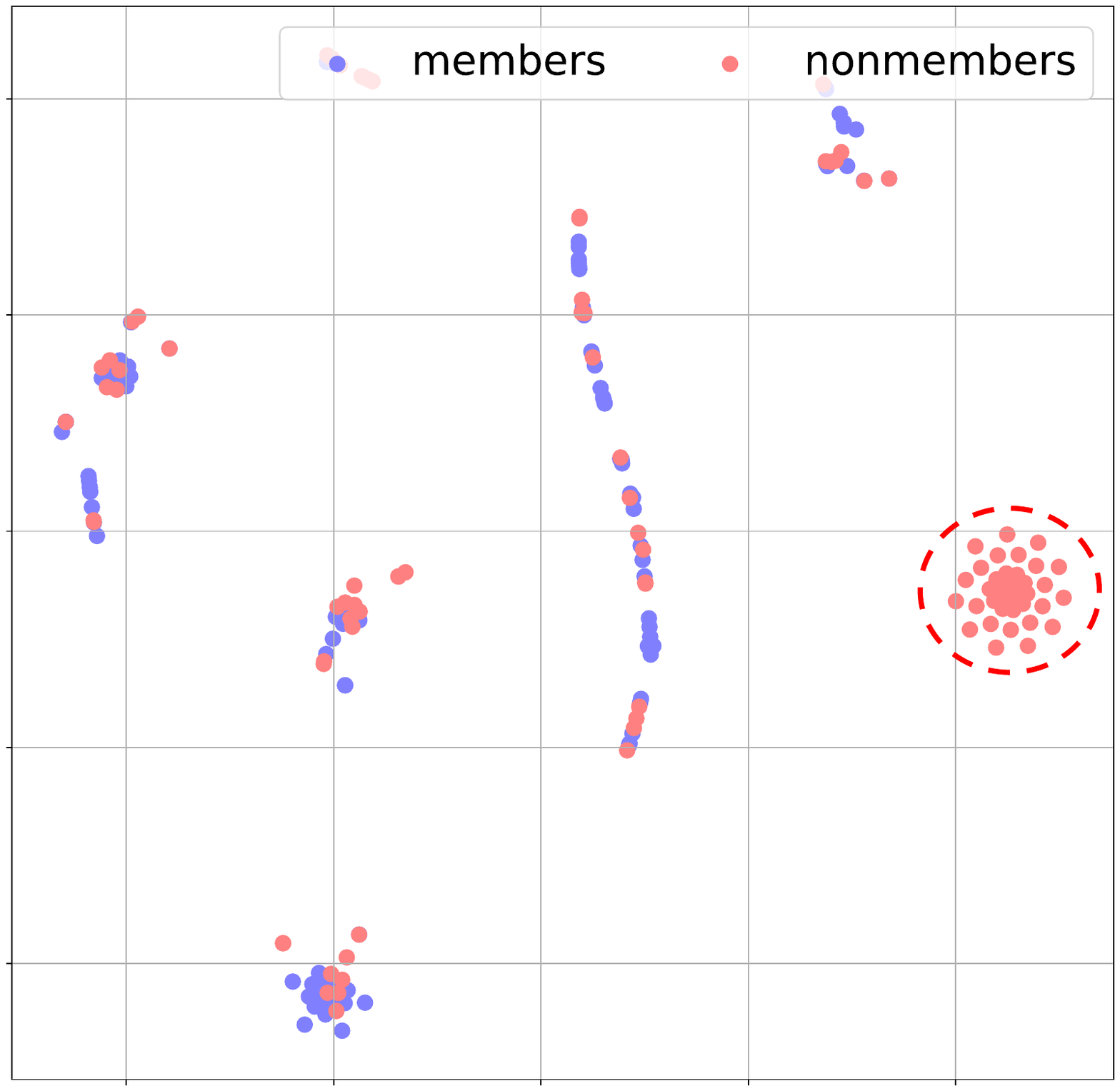}& 
        \includegraphics[trim={6cm 5.5cm 6cm 6cm},scale=0.18]{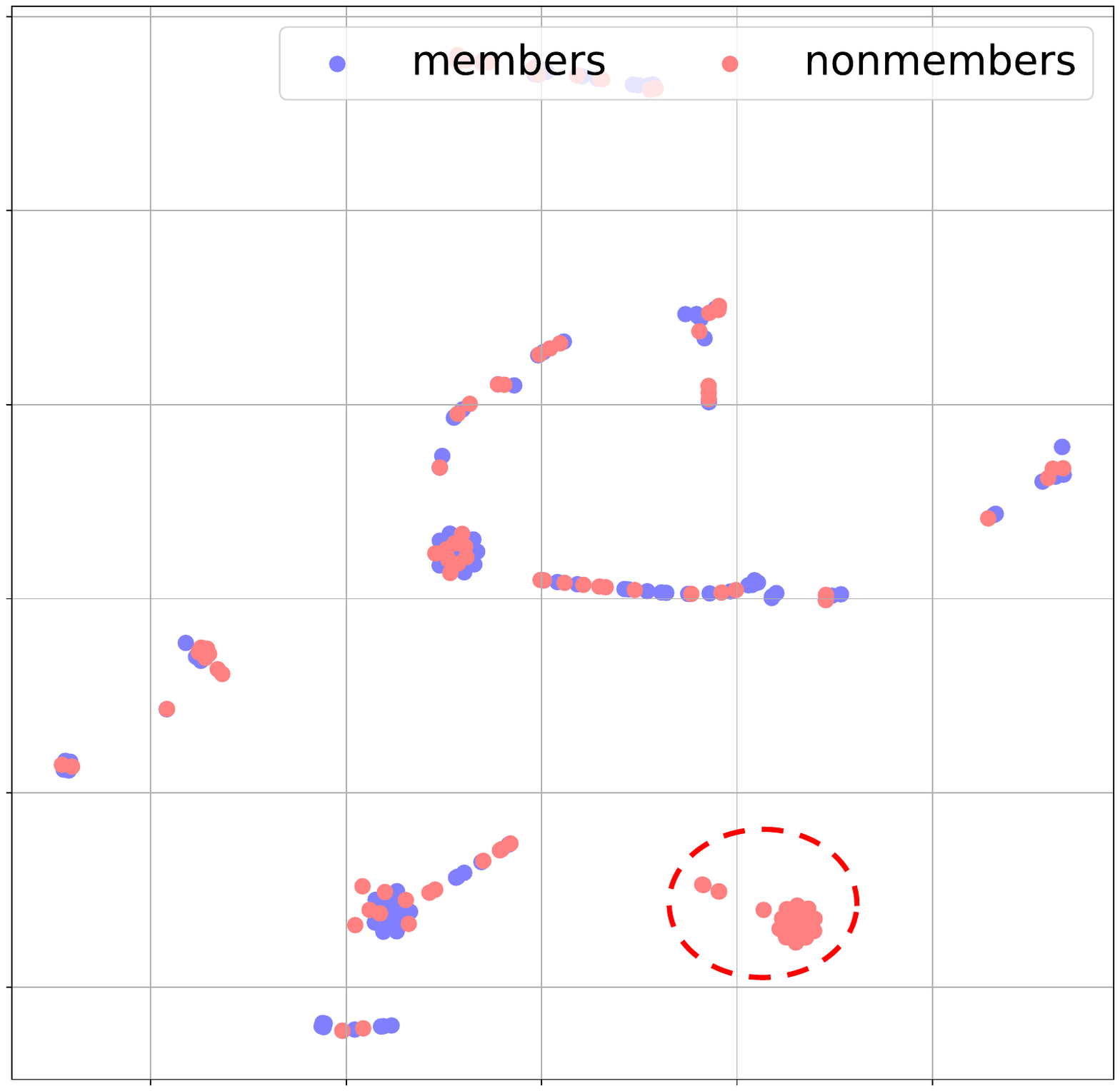}\\
        \\
       \rotatebox{90}{ LDL(0.02)}  & 
       \includegraphics[trim={6cm 3cm 6cm 6cm},scale=0.12]{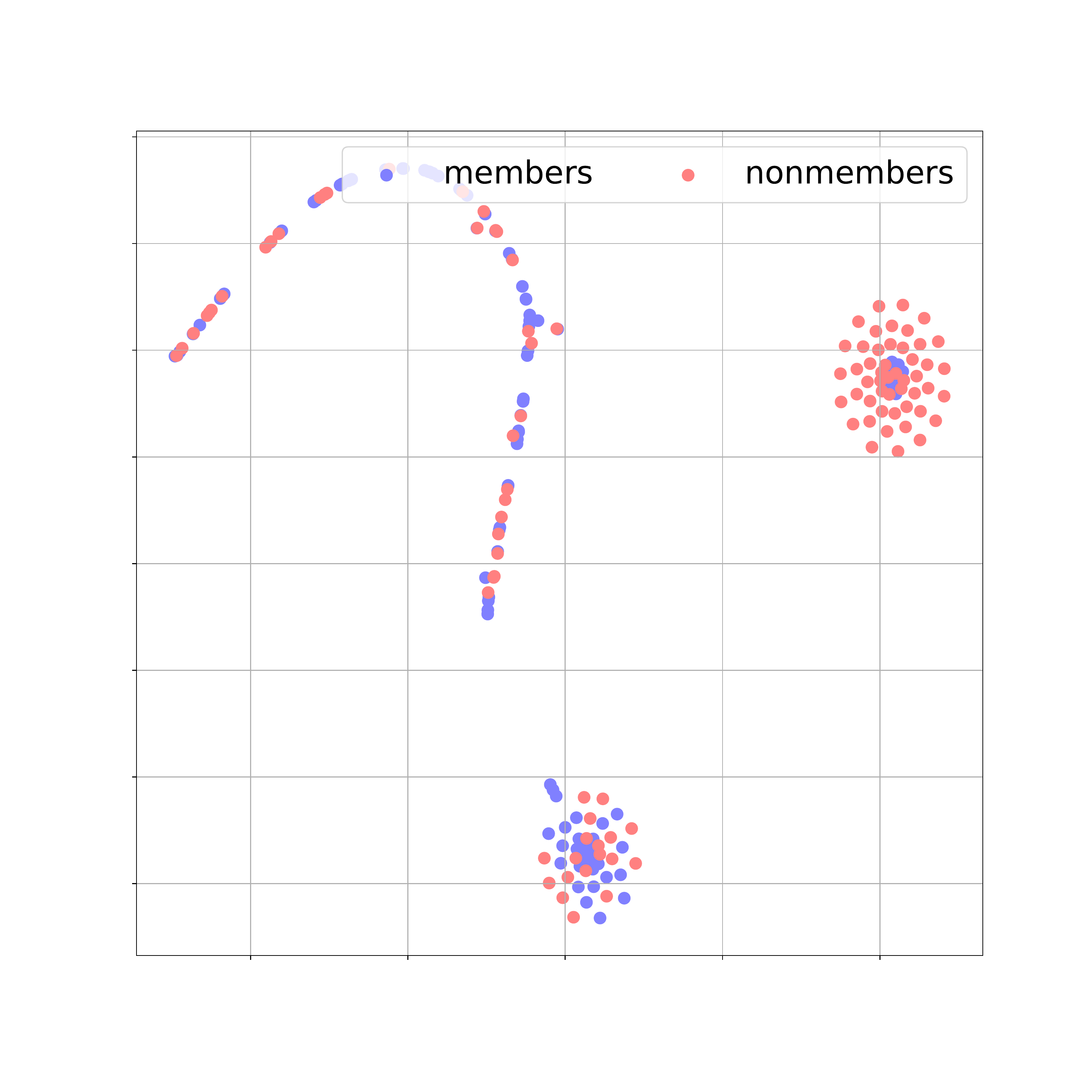} &  
        \includegraphics[trim={6cm 3cm 6cm 6cm},scale=0.12]{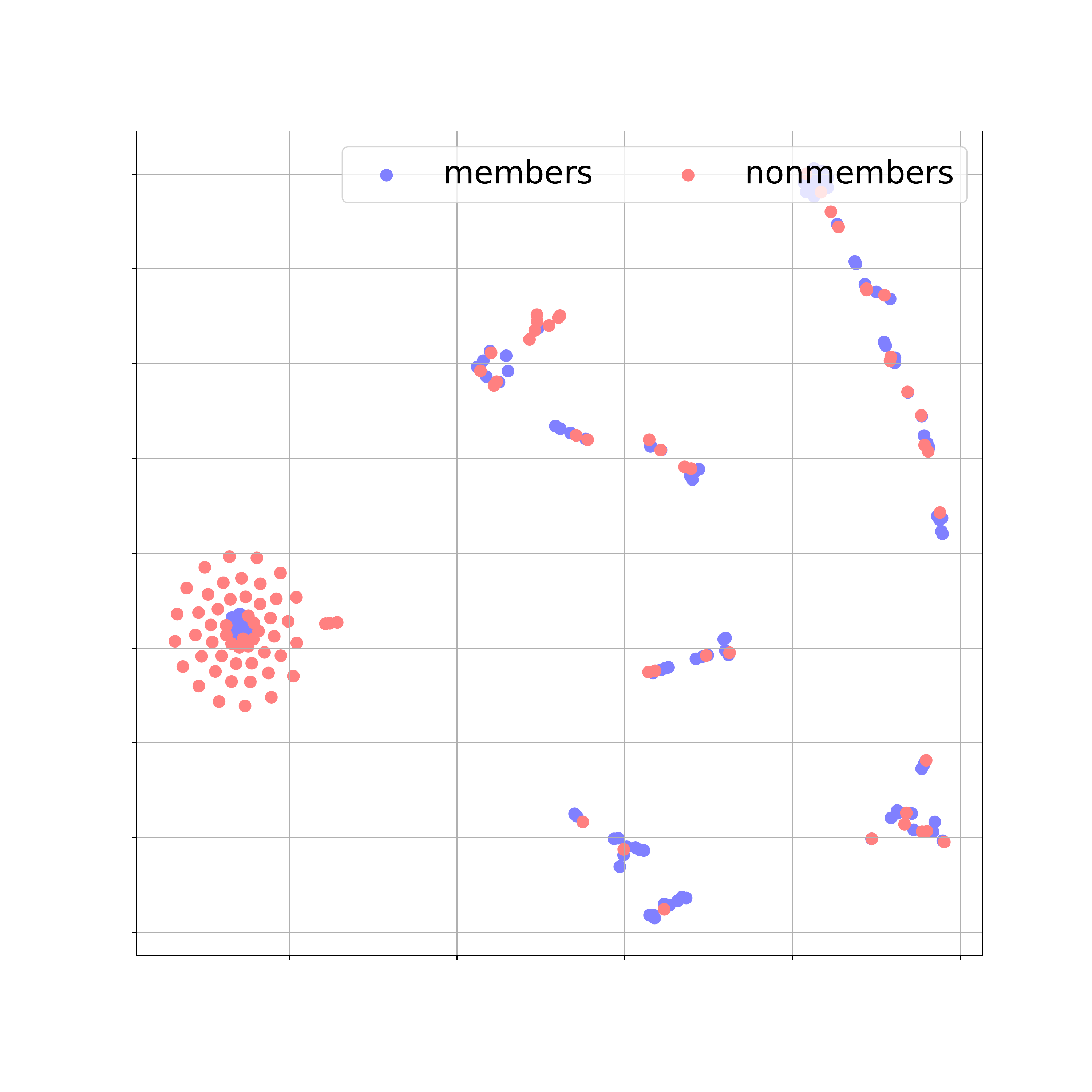} &  
         \includegraphics[trim={6cm 3cm 6cm 6cm},scale=0.12]{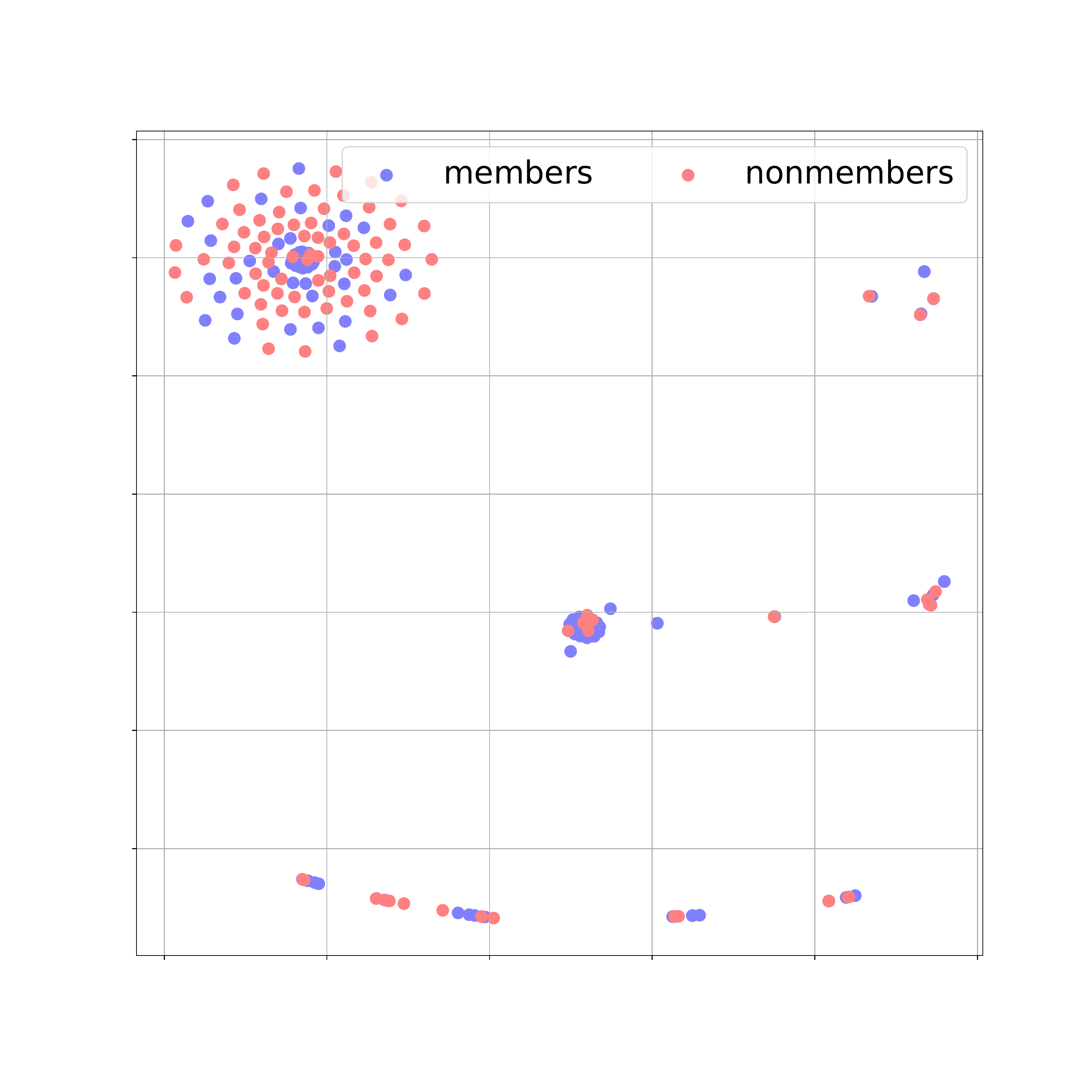} & 
         \includegraphics[trim={6cm 3cm 6cm 6cm},scale=0.12]{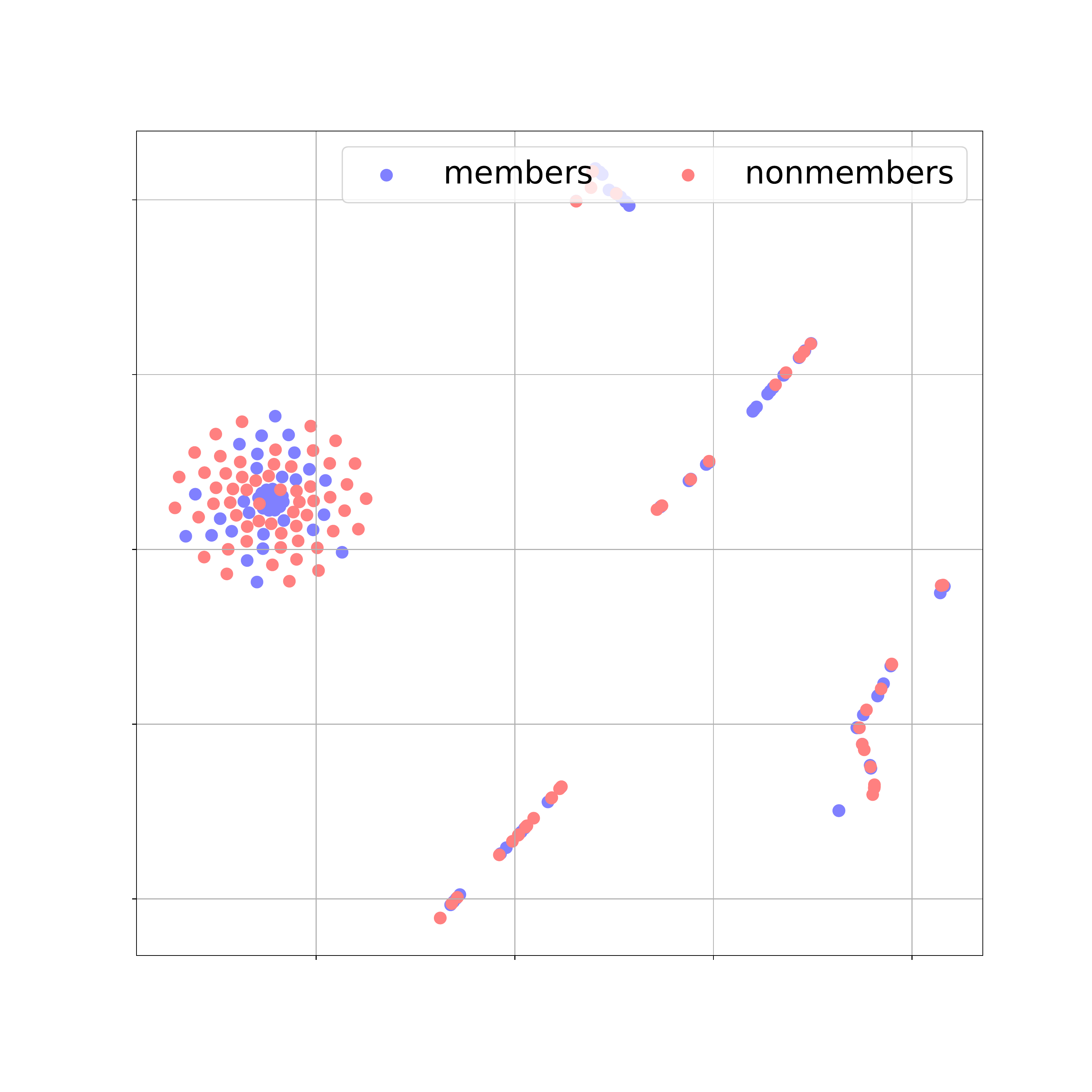} \\
        & (a) CIFAR-10 & (b) CIFAR-100 & (c) GTSRB & (d) Face 
    \end{tabular}
    \caption{t-SNE representation of separability of members and nonmembers for 4 datasets. We calculate ${score_G} (x)$ for 10 different values of $\sigma'$ 
    and use t-SNE to translate them to a $2D-$representation. 
    For defense-free models (top row), there is a region denoted by the dashed ellipse that enables an adversary to easily identify nonmembers. In comparison, LDL successfully prevents an adversary from distinguishing members from nonmembers since there is no region that clearly separates these two (blue and red dots on graphs in bottom row).}
    \label{fig:tsne}
\end{figure*}

\subsection{LDL vs. Adversary using Adversarial Noise
}\label{sec:hopskipjump}


We evaluate LDL against an adversary carrying out a LAB MIA. 
The adversary uses a black-box adversarial model \emph{HopSkipJump}~\cite{chen2020hopskipjumpattack} to estimate the minimum adversarial noise required to cause the DNN model to misclassify a sample. 
To distinguish members from nonmembers, the adversary estimates the best threshold on the minimum amount of noise needed to separate them
~\cite{ccs2021}. 
%
Our objective is to demonstrate that using LDL effectively reduces the $ASR$ value, i.e., the ability of the adversary to distinguish members from nonmembers. 
We examine whether LDL successfully reduces the value of $ASR$ to $ASR_{gap}$ for a strong adversary (Eqn. (\ref{eq:asrmin}) and (\ref{eqn:asrldl})). 
For a weak adversary, we examine if LDL successfully reduces the value of $ASR$ to $0.5$, which corresponds to tossing a fair coin.
%

Fig.~\ref{fig:strongadv} shows the $ASR$ for defense-free models (MIA, red bars) and models with LDL (green bars). 
Our experiments reveal that deploying LDL hinders an adversary using  \emph{HopSkipJump} to determine the minimum magnitude of noise required to misclassify samples. 

 \noindent{\it \underline{Strong adversary}:} 
 A strong adversary can carry out a \emph{gap attack}~\cite{icml2021} (Sec~\ref{subsec:overfitted}), which identifies a sample as a member if it is classified correctly by the DNN model, and a nonmember otherwise. 
The success rate of the adversary carrying out a gap attack is given by $ASR_{gap}$ in Eqn. (\ref{eq:asrmin}). 
Similar to~\cite{icml2021}, we use the value of $ASR_{gap}$ as a benchmark to evaluate how well LDL reduces the $ASR$ of LAB MIAs. 
Reducing the $ASR$ below $ASR_{gap}$ by a large amount will affect classification accuracy of the model ($Acc_{mem}, Acc_{nonmem}$ terms in Eqn.~(\ref{eq:asrmin})). 
The left column of Fig.~\ref{fig:strongadv} demonstrates that while the adversary has a high $ASR$ value when carrying out a LAB MIA on defense-free DNN models (red bars), using LDL reduces the $ASR$ to near $ASR_{gap}$ (green and orange bars are comparable).

\noindent{\it \underline{Weak adversary}:} 
A weak adversary assumes that the decision of the DNN model is the true label (Sec~\ref{subsec:AdvModel}). 
In this case, we use $50\%$ (i.e., a fair coin-toss) as a benchmark (orange bars in Fig.~\ref{fig:strongadv} right column) to evaluate the performance of LDL. 
The right column of Fig.~\ref{fig:strongadv} indicates that in the defense-free case, a weak adversary carrying out a LAB MIA can achieve high $ASR$ even when using the output of the DNN model as the true label (red bars). 
Using LDL reduces $ASR$ to $50\%$ for the weak adversary (orange and green bars are comparable), thus making members and nonmembers indistinguishable.  

\subsection{
LDL vs. Adversary using Random Noise
}\label{sec:randomnoise}
Instead of using adversarial noise to estimate the magnitude of noise for misclassification, the adversary can also use random noise to estimate distance to the decision boundary. 
Such an approach was shown to be successful in~\cite{icml2021}. We use a Gaussian noise, $n \sim \mathcal{N}(0, {\sigma'}^2I)$ and generate $K$ perturbed variants for each sample $x$. 
The adversary chooses the value of $K$ as the lower bound on the number of perturbed samples beyond which the ASR would not increase. This is indeed an empirical parameter, and for datasets that we use from SOTA LAB MIAs~\cite{icml2021, ccs2021}, we empirically found this value to be around $K=1000$. 
The fraction of samples that are classified correctly is computed as: 
\begin{equation}\label{eq:attacckG}
    \texttt{score}_G (x, {\sigma'}^2):= \frac{1}{K}\sum_i^K \mathbb{I} (g(x+n_i)==y),  
\end{equation}
where $\mathbb{I}(t)$ returns $1$ if $t$ is true and $0$ otherwise, $g(\cdot)$ returns the predicted label for the perturbed variant of input $x$, and $y$ is the true label of $x$. 
We observe that for smaller values of ${\sigma'}^2$, $\texttt{score}_G (x, {\sigma'}^2)$ is lower for members than for nonmembers. 
A justification for this observation is that members are less likely to be misclassified for small perturbations since they are typically located far from a decision boundary~\cite{shokri2017membership, ccs2021, icml2021}. 

Fig.~\ref{fig:randomnoise} shows the $ASR$ of an adversary that uses random noise to carry out a LAB MIA. 
We observe that the $ASR$ value for the defense-free setting 
is lower for an adversary carrying out a LAB MIA using random noise than when using adversarial noise (red bars in Fig.~\ref{fig:randomnoise} are lower than red bars of Fig.~\ref{fig:strongadv}). 
A reason for this is that adversarial learning techniques aim to optimize the minimum noise required for misclassification. In comparison, random noise is a heuristic used by the adversary. 
To 
reduce the value of $ASR$, 
we will require a larger value of $\sigma^2$ as shown by the blue bars of Fig.~\ref{fig:randomnoise}. 
A larger $\sigma^2$ corresponds to a label-invariant sphere of larger radius. 
Consequently, the adversary would require to add a noise of larger magnitude in order for samples to be misclassified. 
These observations are consistent with the insight underpinning LDL described in Sec. \ref{sec:intution} and Eqn. (\ref{eqn:asrldl}).  



In Eqn.~(\ref{eq:attacckG}), we had computed $\texttt{score}_G (x, {\sigma'}^2)$ for each sample $x$. 
We collect values of $\texttt{score}_G (x, {\sigma'}^2)$ for $10$ different ${\sigma'}^2$ into a vector, denoted by $\texttt{score}_G (x)$. 
We use a t-distributed stochastic neighbor embedding (t-SNE)~\cite{van2008visualizingtsne} to demonstrate that deploying LDL reduces the ability of an adversary to distinguish members from nonmembers. 
The t-SNE is a technique to visualize high-dimensional data through a two or three dimensional representation.

Fig.~\ref{fig:tsne} shows that score vectors for members and nonmembers before LDL is deployed can be easily distinguished from each other, since there is a region that completely consists of only nonmembers (dashed ellipse in top row). 
However, after deploying LDL, it becomes difficult for the adversary to distinguish members from nonmembers (bottom row). 

%
\begin{figure*}
    \centering
    \begin{tabular}{c c }
       \includegraphics[trim={6cm 0.5cm 2cm 3cm}, scale=0.13]{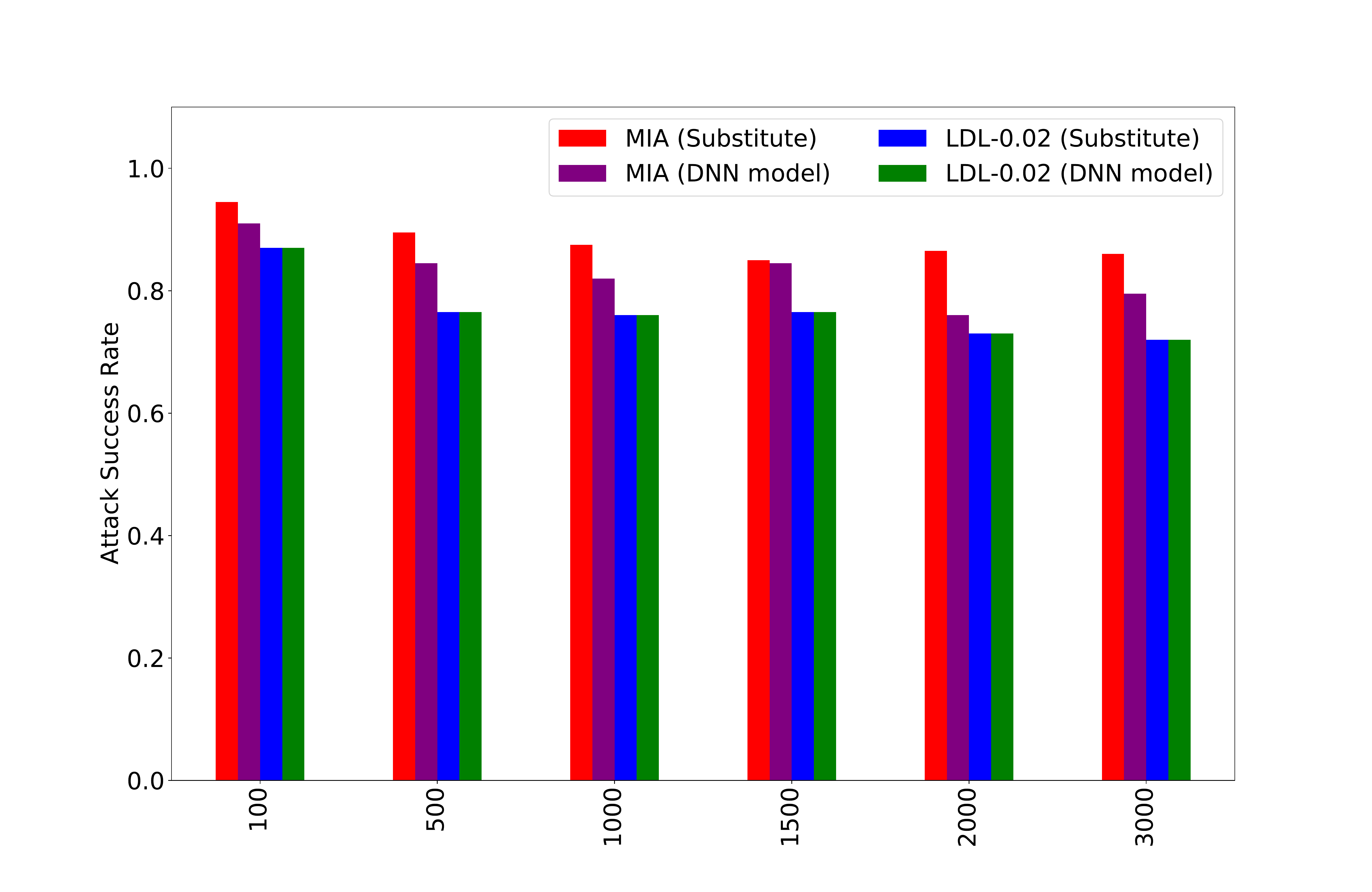} & 
    \includegraphics[trim={6cm 0.5cm 2cm 3cm}, scale=0.13]{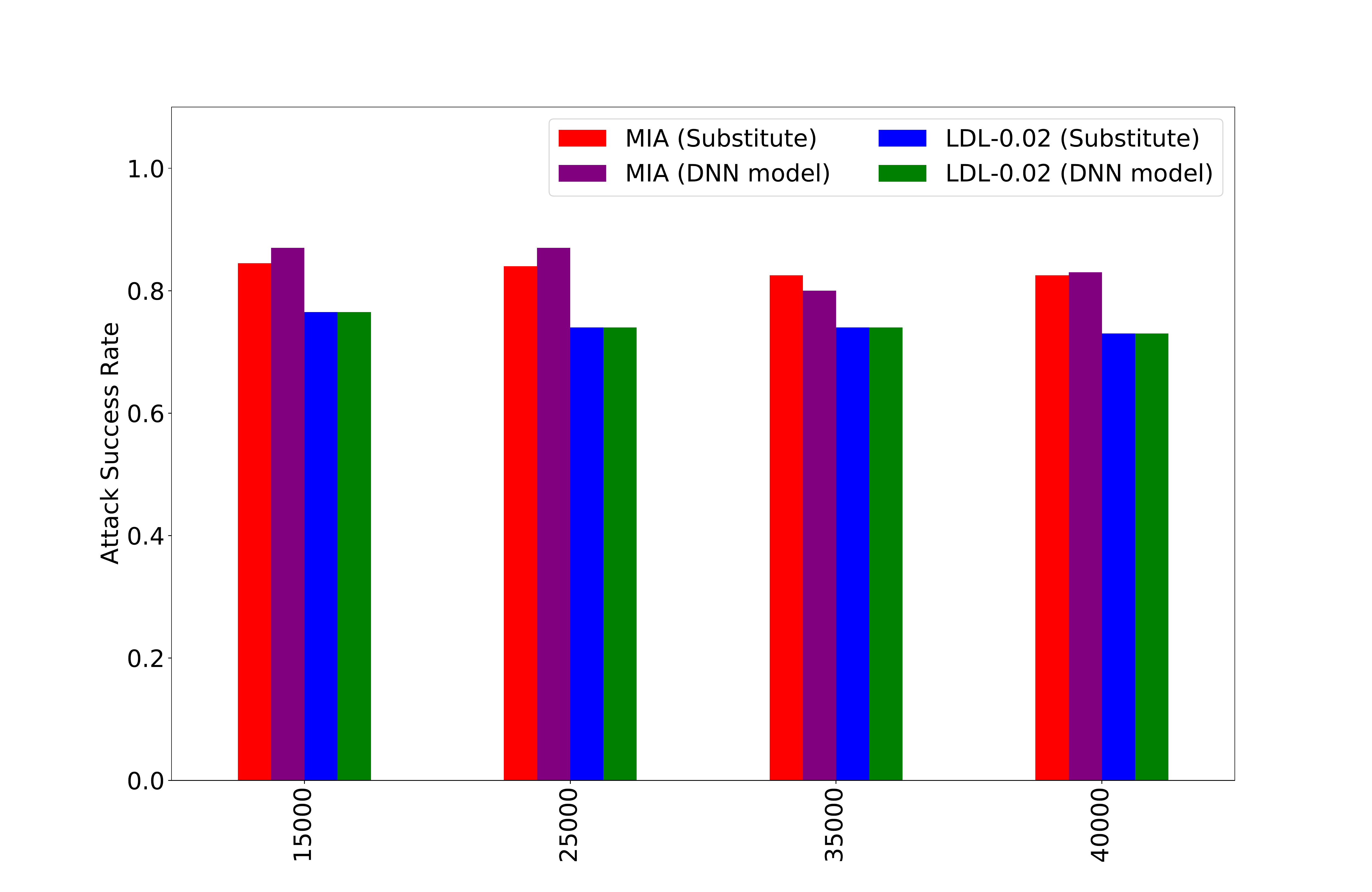} \\
    (a) CIFAR-10 &  (b) CIFAR-100 \\
        \includegraphics[trim={6cm 0.5cm 2cm 3cm}, scale=0.13]{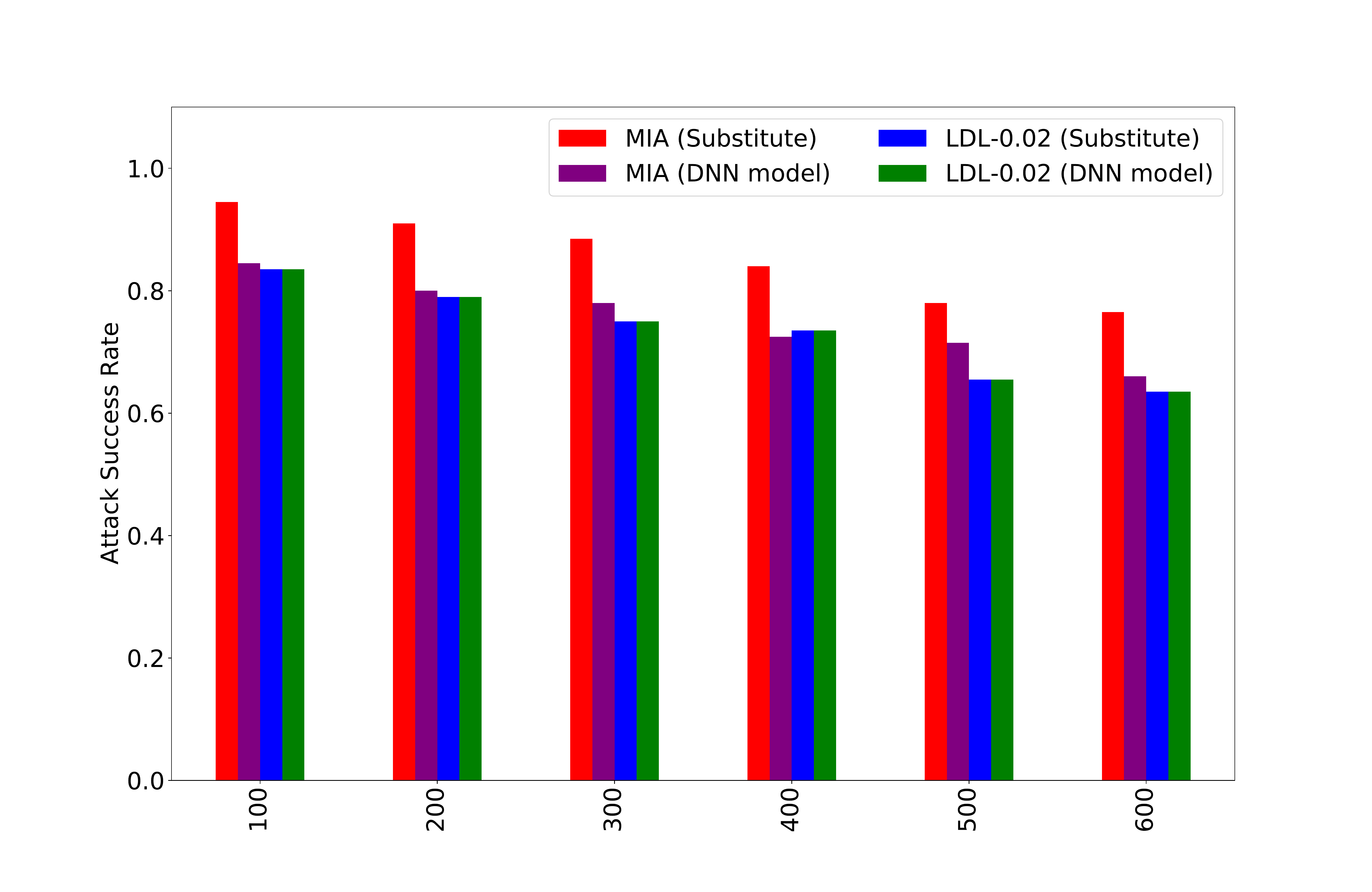}&  
         \includegraphics[trim={6cm 0.5cm 2cm 3cm},scale=0.13]{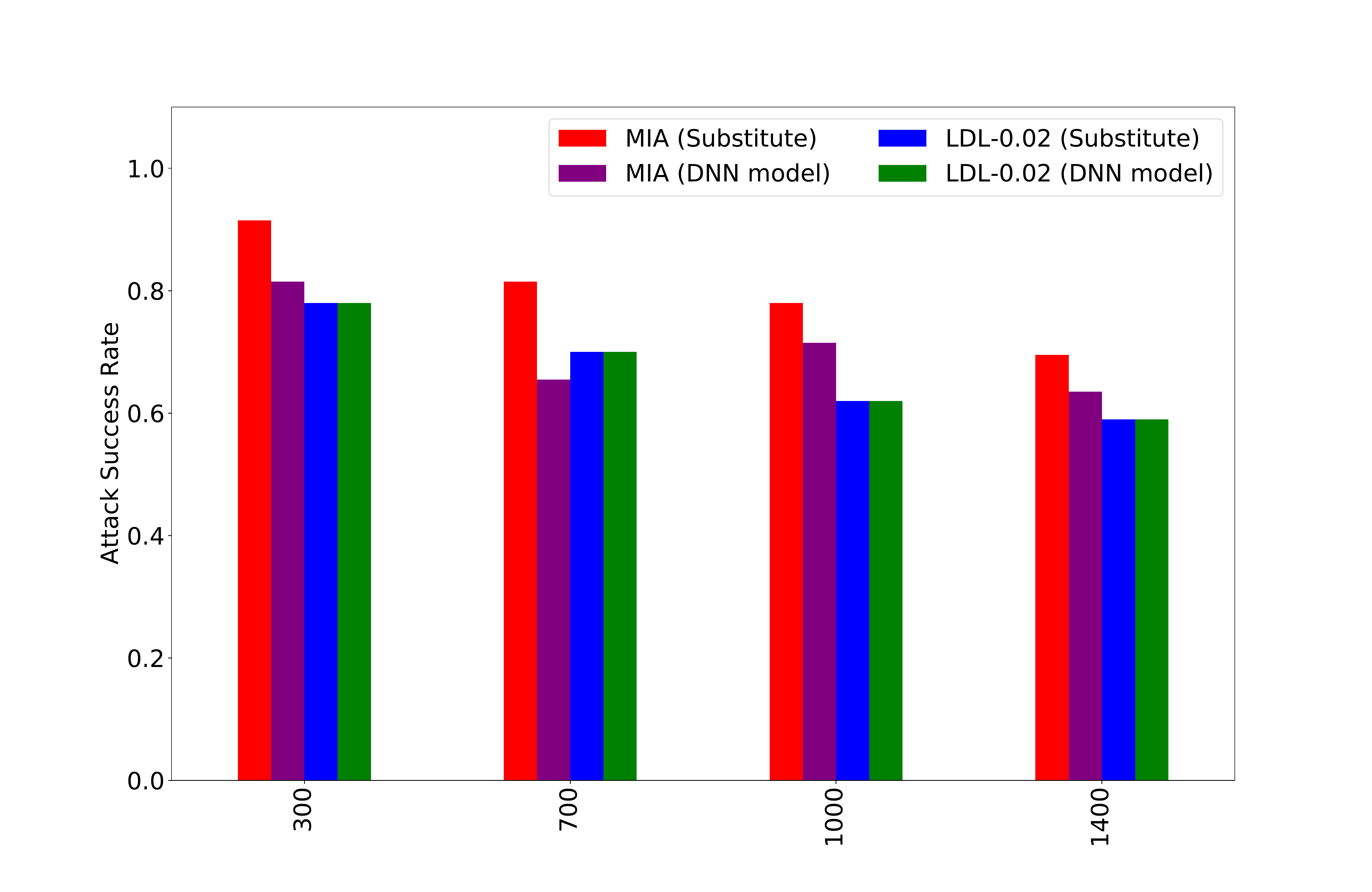}\\
        (c) GTSRB &  (d) Face
    \end{tabular}
    \caption{The success rate of LAB MIAs using HopSkipJump for adversary with a substitute classifier~\cite{ccs2021}. The adversary uses a local substitute classifier to estimate a threshold for distinguishing the minimum required perturbation for misclassification of members and nonmembers. The smallest amount of adversarial noise required to misclassify each sample by the DNN classifier is estimated using HopSkipJump. The threshold is then used to identify a sample as a member or nonmember. 
    Red bars show attack success rate (ASR) of an adversary carrying out a LAB MIA on a defense-free substitute classifier and purple bars show ASR on a defense-free variant of the DNN classifier using the threshold. Blue and green bars show the ASR of substitute and DNN classifiers after deploying LDL.
    We observe that deploying LDL reduces the ASR of LAB MIAs. Further, the substitute models learned by the adversary are a good representation of the DNN model (blue and green bars are almost equal). 
    }
    \label{fig:shadow}
\end{figure*}

\subsection{LDL vs. Adversary using Substitute Models}\label{sec:shadowmodels}
An adversary may not have access to 
member and nonmember samples to allow it to estimate the 
minimum amount of noise required to be added in order to misclassify the sample. 
However, it might have access to a dataset of other samples along with their labels. 
In such a scenario, the authors of~\cite{icml2021} suggested using a substitute model (also termed shadow model~\cite{icml2021}) trained locally by a strong adversary. 
They assumed that the adversary knew the structure of the DNN model and used this to train a substitute model with the same structure on their own dataset. 
%
This enabled the adversary to estimate the distance of members and nonmembers to a classification decision boundary. 
Using this information, the minimum adversarial noise required to misclassify each sample in the DNN model was computed using \emph{HopSkipJump}~\cite{chen2020hopskipjumpattack}. 
The adversary then applied a locally estimated threshold (from the substitute model) to distinguish members from nonmembers. 
Fig.~\ref{fig:shadow} indicates that an adversary using a substitute model to carry out a LAB MIA can effectively distinguish between members and nonmembers 
(red and purple bars). 
Deploying LDL with $\sigma^2 = 0.02$ (green and blue bars) successfully reduces the the value of $ASR$. 
Similar values of $ASR$ observed for substitute (blue bars in Fig.~\ref{fig:shadow}) and DNN models (green bars) indicate that the substitute models learned by the adversary are a good representation of the DNN model. 
The results in this case are also consistent with our observation in Fig.~\ref{fig:strongadv} where LDL was able to reduce the value of $ASR$ 
against a strong adversary employing \emph{HopSkipJump} to carry out a LAB MIA. 


\begin{figure*}
    \centering
    \begin{tabular}{>{\centering\arraybackslash} m{0.14cm} >{\centering\arraybackslash} m{3.8cm} >{\centering\arraybackslash} m{3.8cm}>{\centering\arraybackslash} m{3.8cm}>{\centering\arraybackslash} m{3.8cm} }
     
  \rotatebox{90}{Nonmembers} &\includegraphics[trim={4cm 1cm 8cm 4cm}, scale=0.15]{ 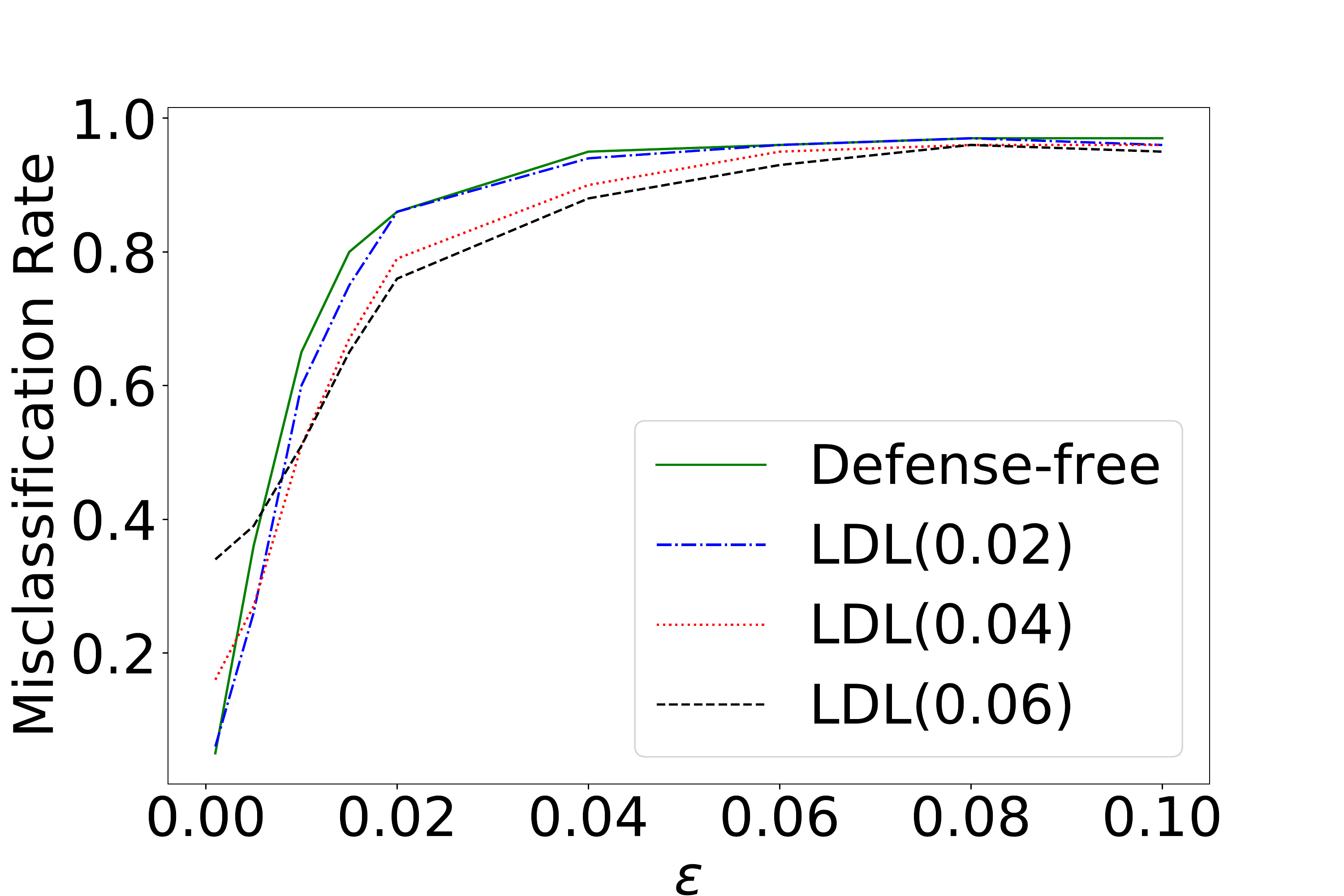}
    &\includegraphics[trim={4cm 1cm 8cm 4cm}, scale=0.15]{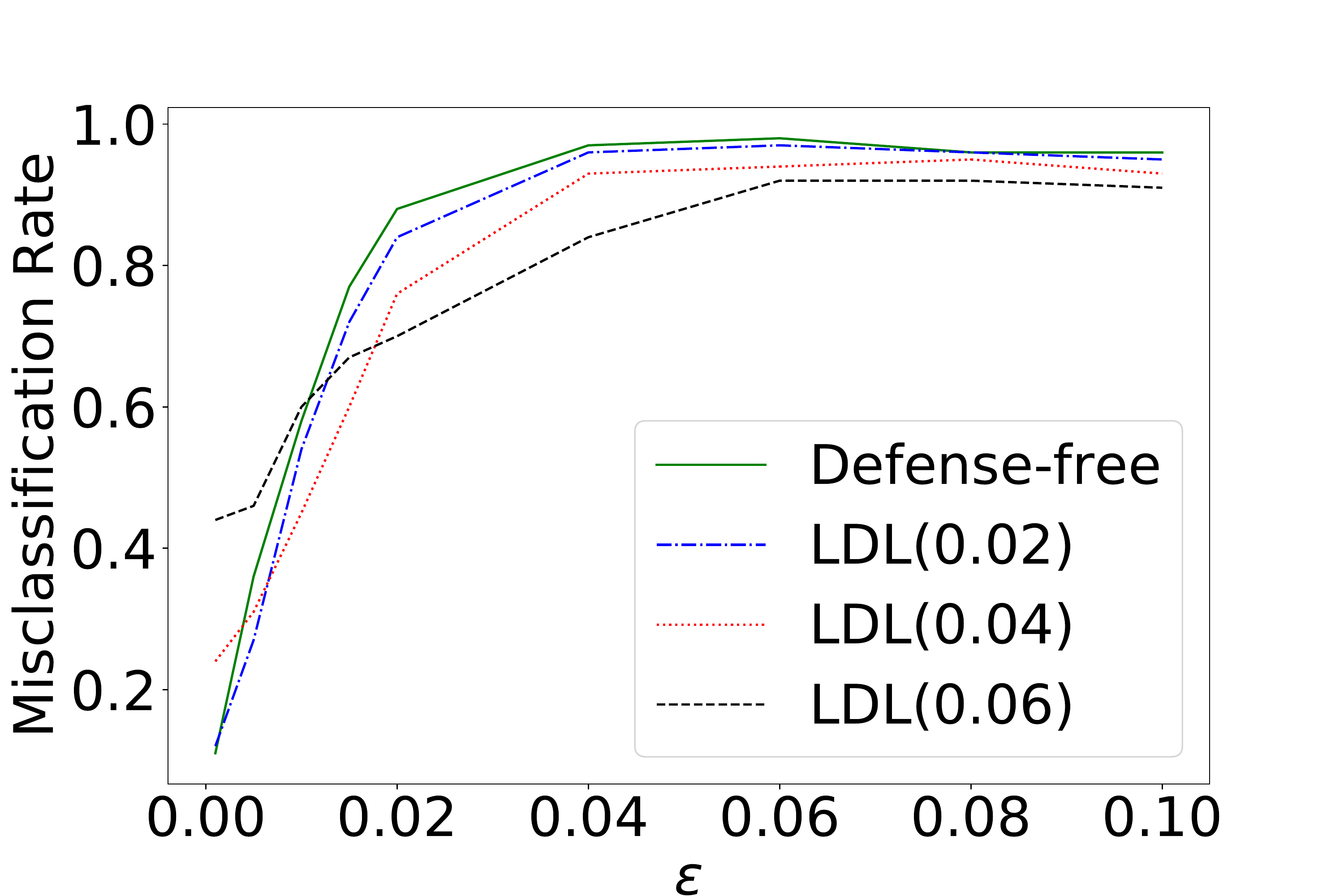}&
        \includegraphics[trim={4cm 1cm 8cm 4cm}, scale=0.15]{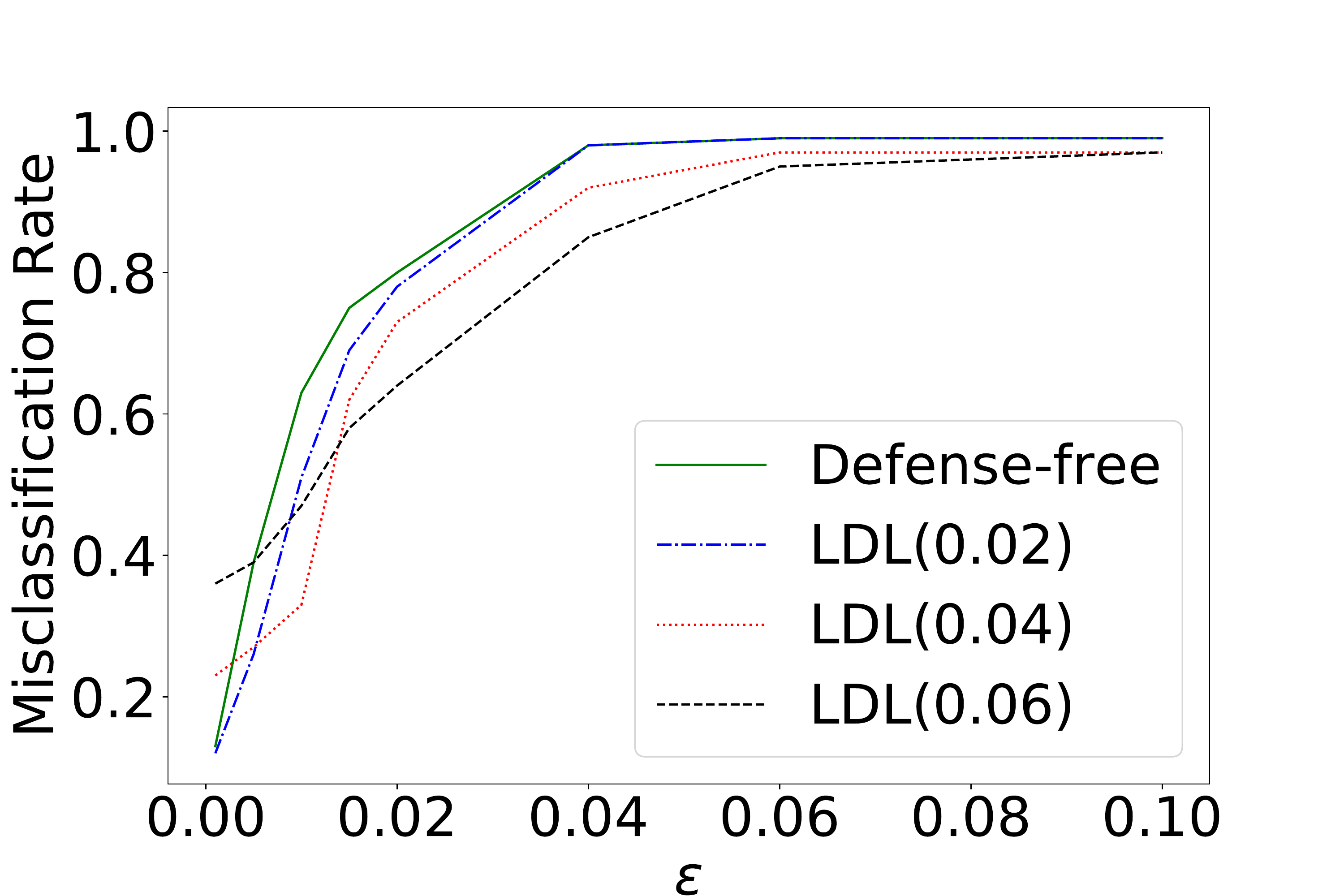}&
    \includegraphics[trim={4cm 1cm 8cm 4cm}, scale=0.15]{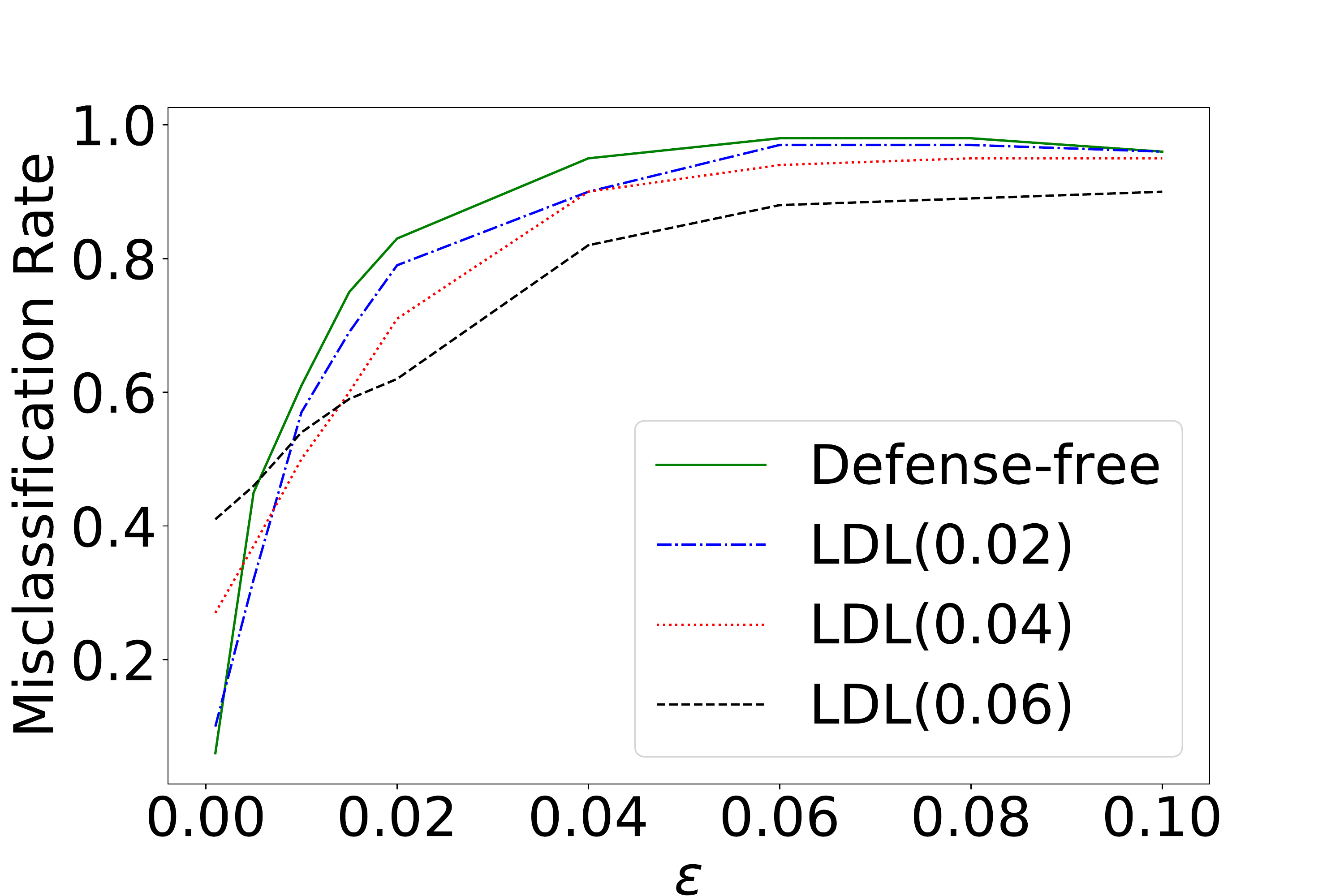}\\ \\
    
  \rotatebox{90}{Members} &\includegraphics[trim={4cm 1cm 8cm 4cm}, scale=0.15]{ 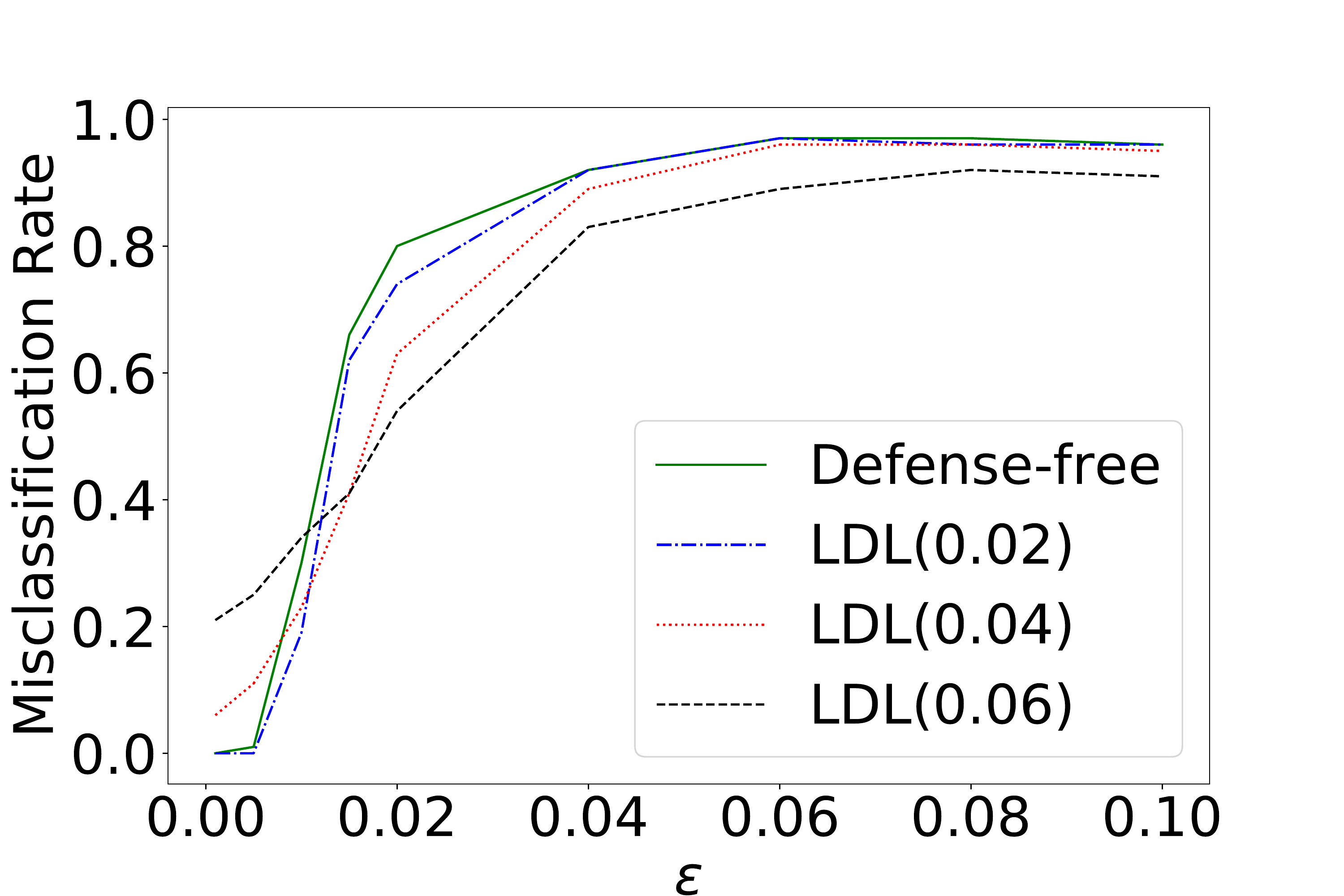}
    &\includegraphics[trim={4cm 1cm 8cm 4cm}, scale=0.15]{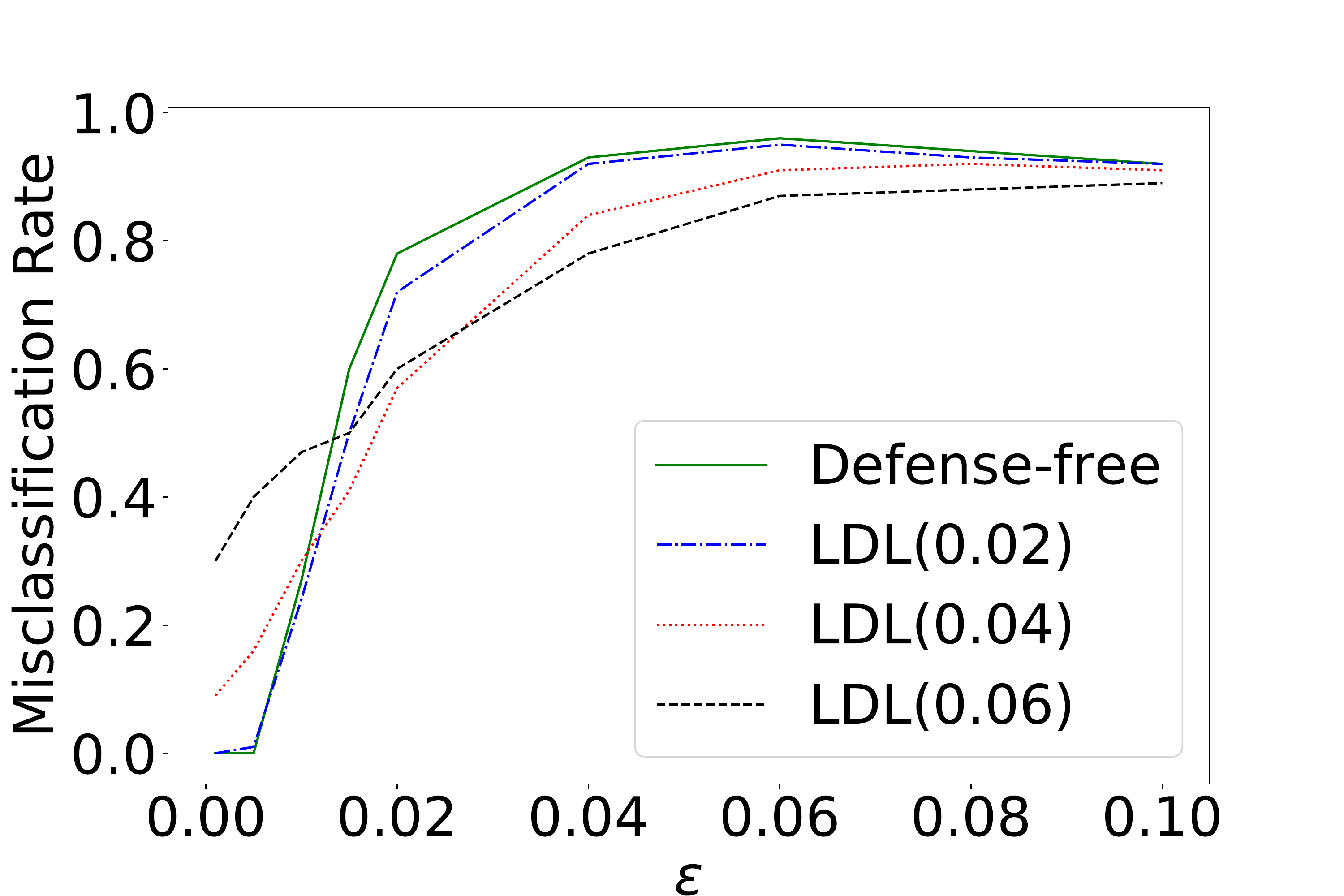}&
        \includegraphics[trim={4cm 1cm 8cm 4cm}, scale=0.15]{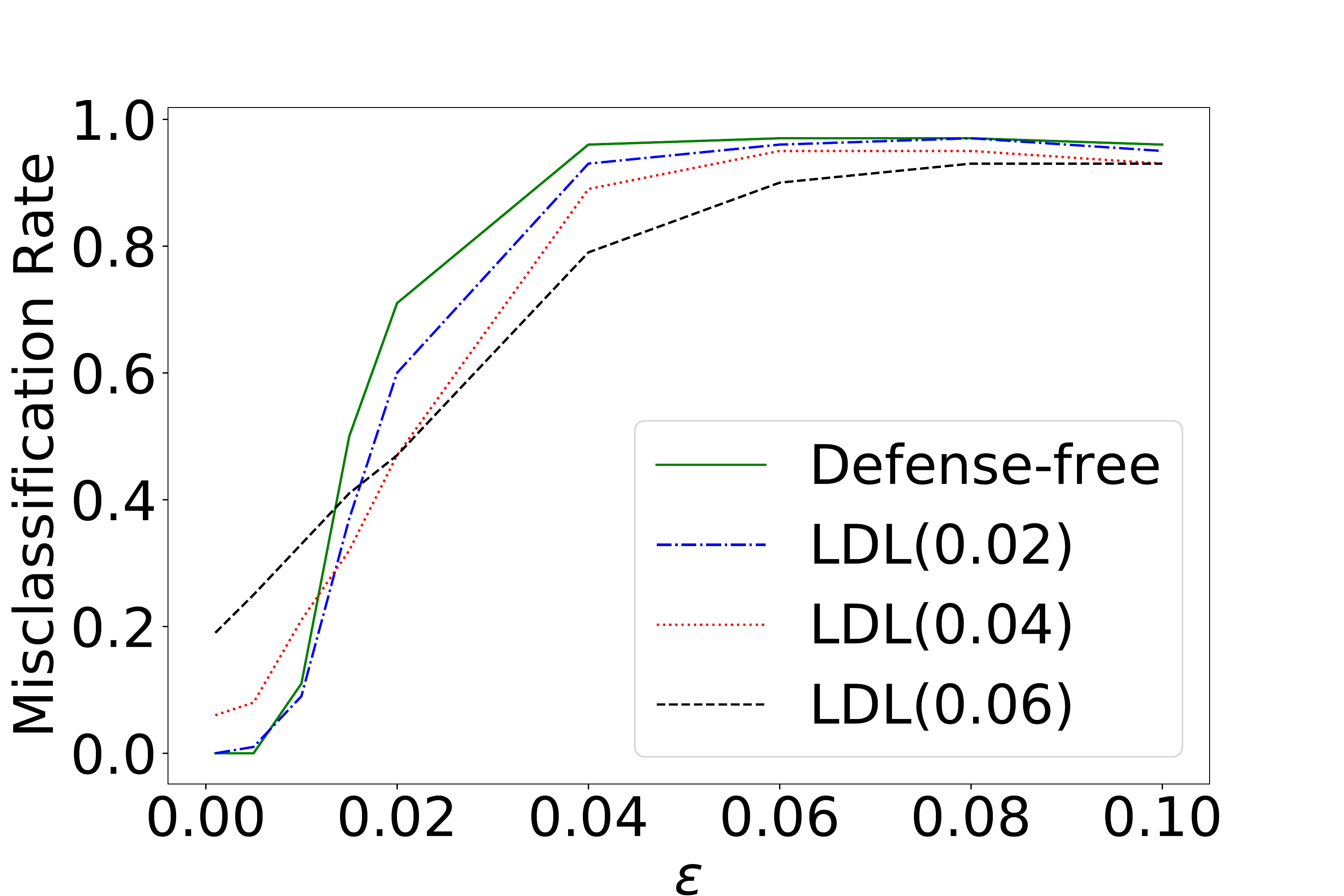}&
    \includegraphics[trim={4cm 1cm 8cm 4cm}, scale=0.15]{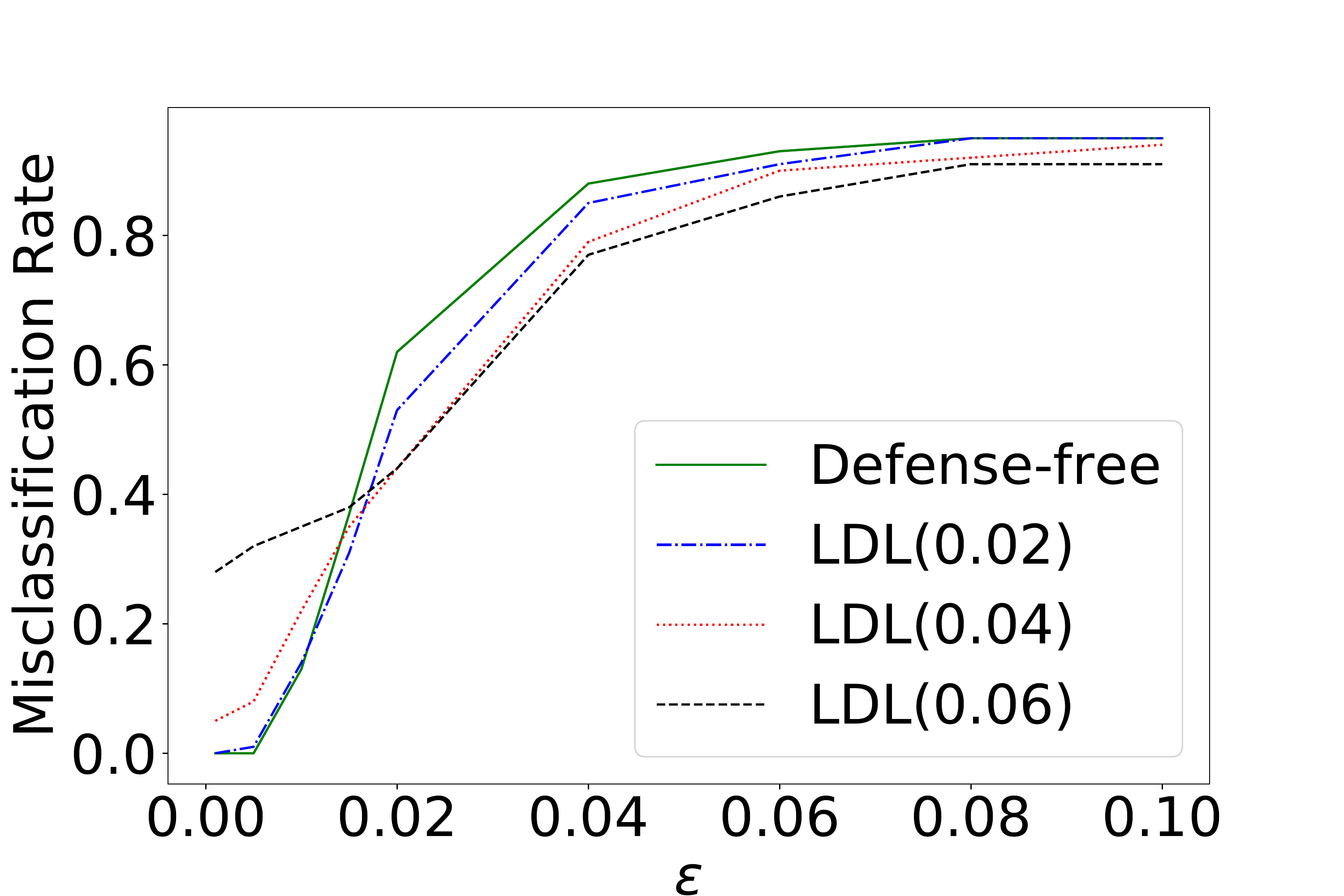}\\
         \\
      &  (a) CIFAR-10 (3000) & (b) CIFAR-10 (2000) & (c) CIFAR-10 (1500) & (d) CIFAR-10 (1000) \\
    \end{tabular}
    \caption{
Misclassification rates of samples of CIFAR-10 with different sizes of training sets with increasing amounts of noise perturbation $\epsilon$. 
%
Deploying LDL ensures that the misclassification rate accomplished by the addition of adversarial noise is similar for both members (training dataset) and nonmembers (testing dataset) in each case. 
}    
    \label{fig:fgs}
\end{figure*}
\begin{figure*}
    \centering
    \begin{tabular} {c c c c}
         \includegraphics[trim={0.6cm 0.1cm 3cm 1cm},clip,scale=0.14]{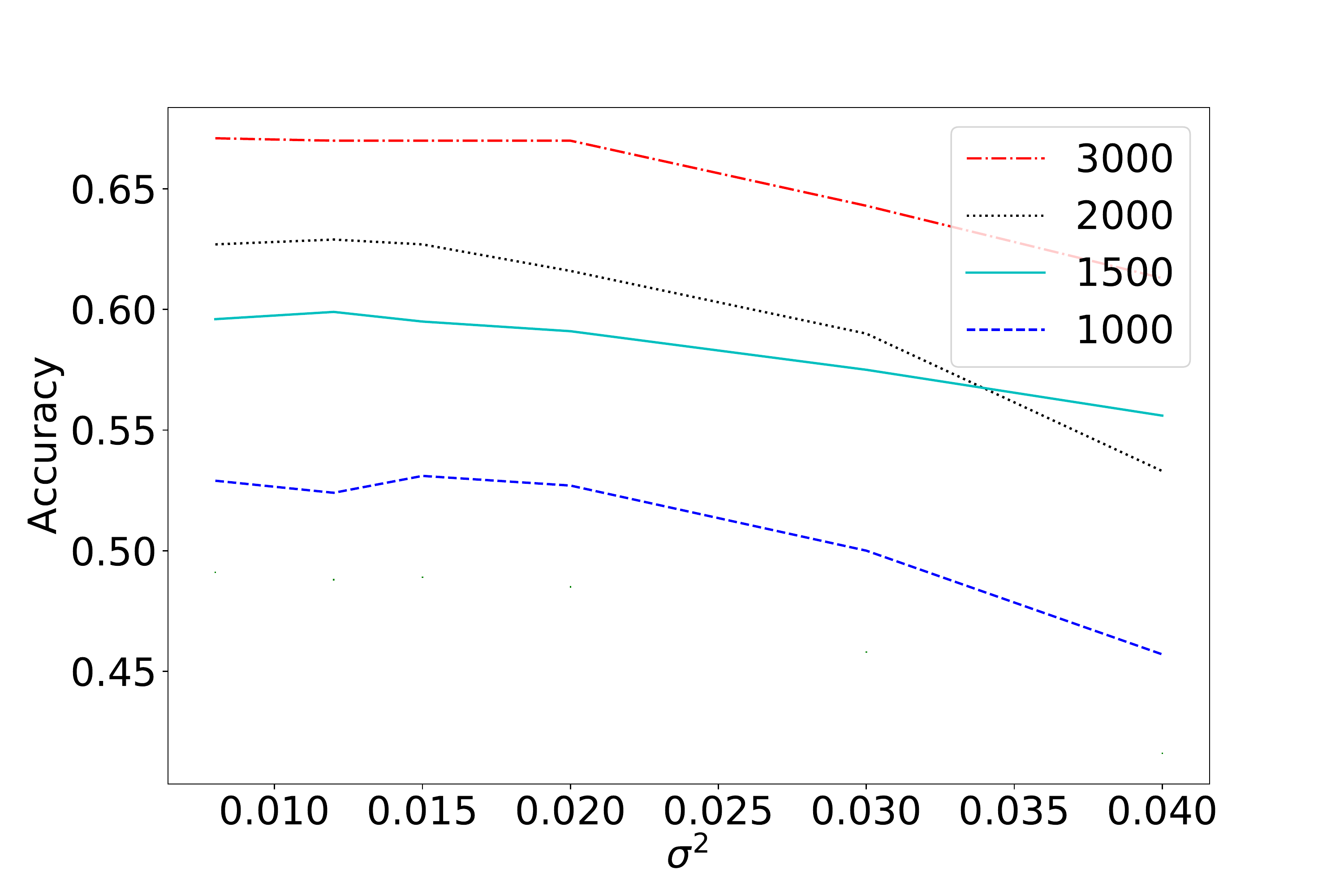} &  
         \includegraphics[trim={0.6cm 0.1cm 3cm 1cm},clip,scale=0.14]{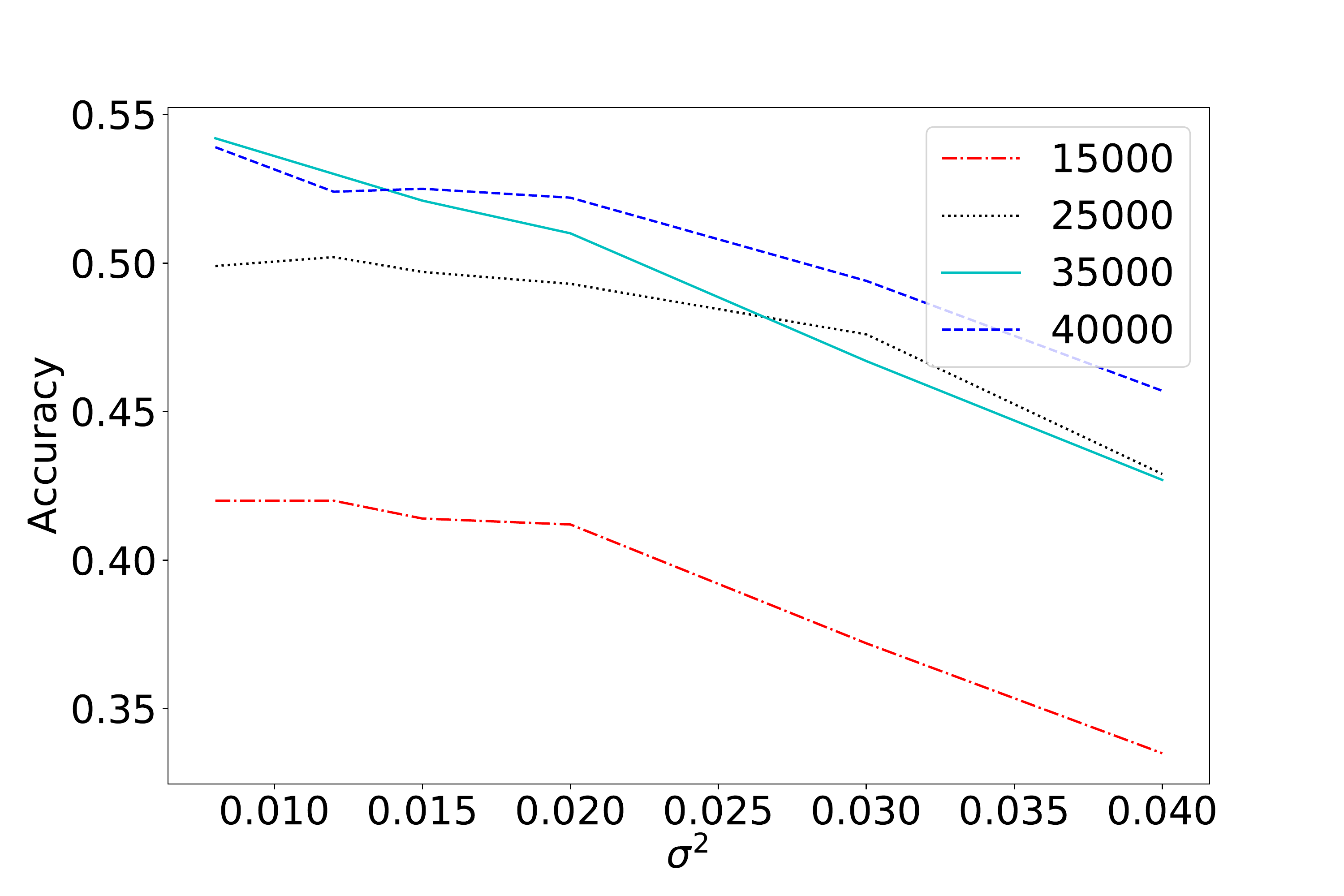} &
         \includegraphics[trim={0.6cm 0.1cm 3cm 1cm},clip,scale=0.14]{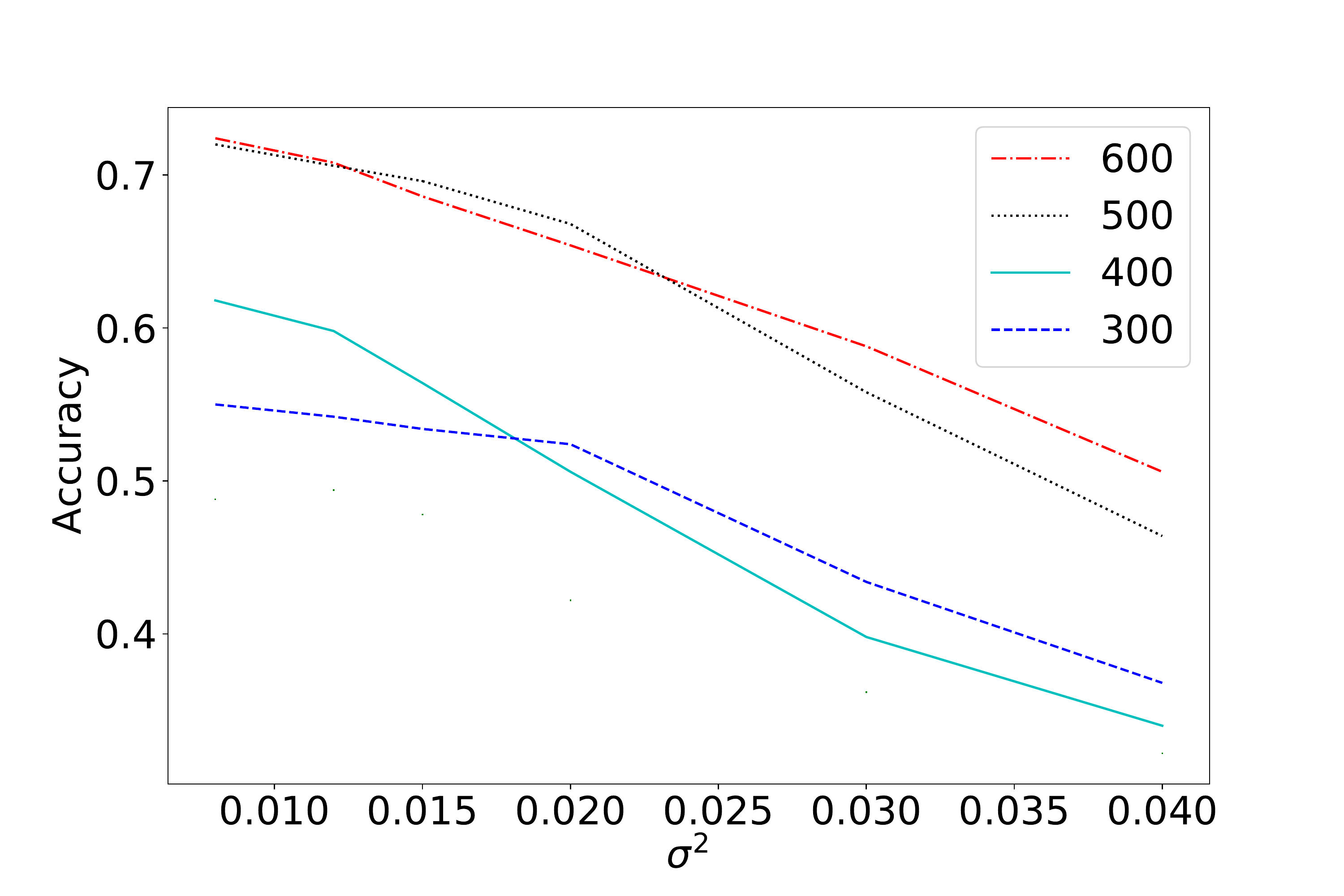} & 
         \includegraphics[trim={0.6cm 0.1cm 3cm 1cm},clip,scale=0.14]{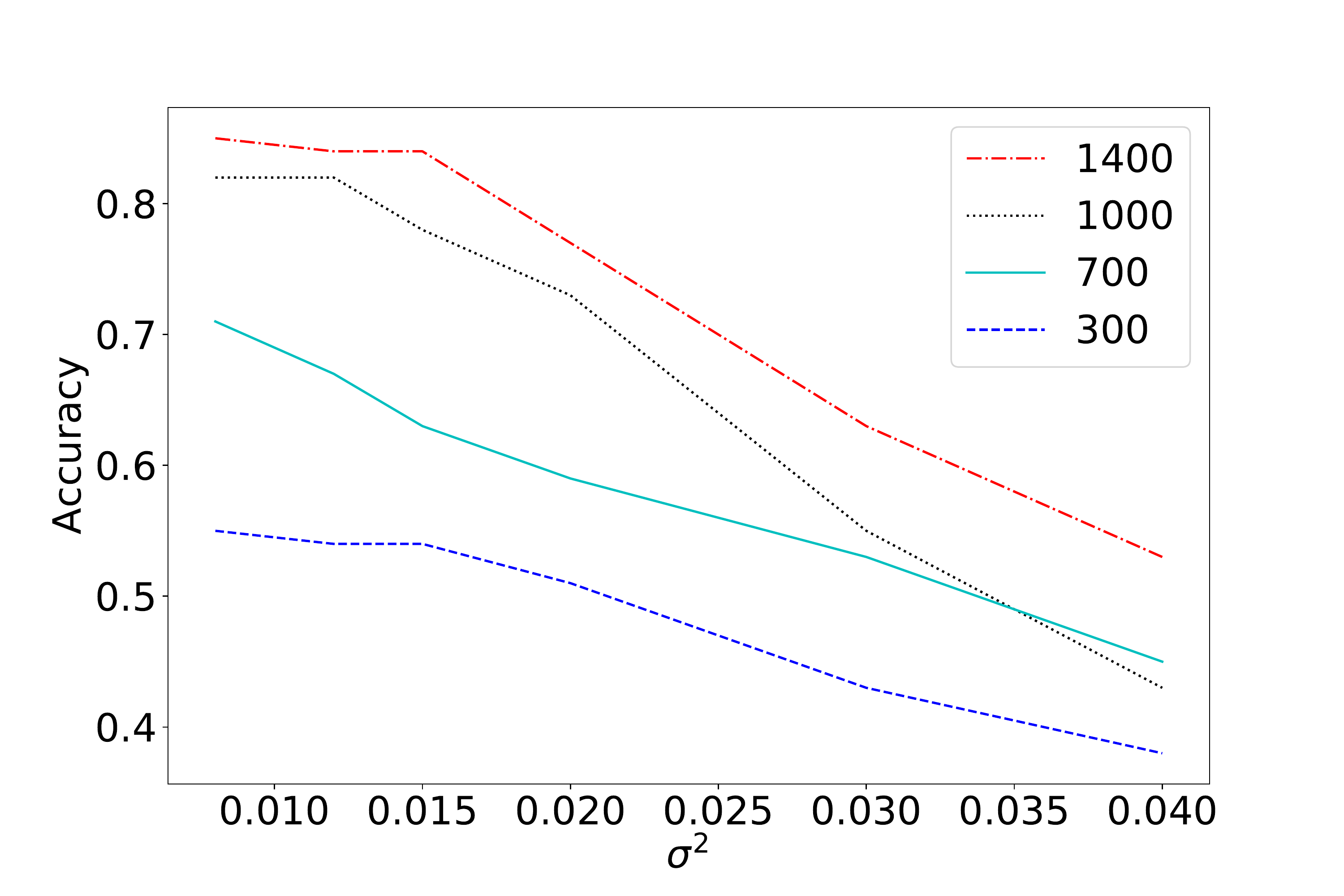} \\
           (a) CIFAR-10 & (b) CIFAR-100 & (c) GTSRB & (d) Face 
    \end{tabular}
    \caption{Classification accuracy on nonmember samples vs. magnitude of perturbation $\sigma^2$ when LDL is deployed on 4 datasets- CIFAR10, CIFAR100, GTSRB, Face- with different sizes of training data. The classification accuracy reduces with increasing $\sigma^2$.}
    \label{fig:accvssigma}
\end{figure*}
\begin{figure}
    \centering
    \includegraphics[trim={2cm 2cm 2cm 2cm},scale=0.2]{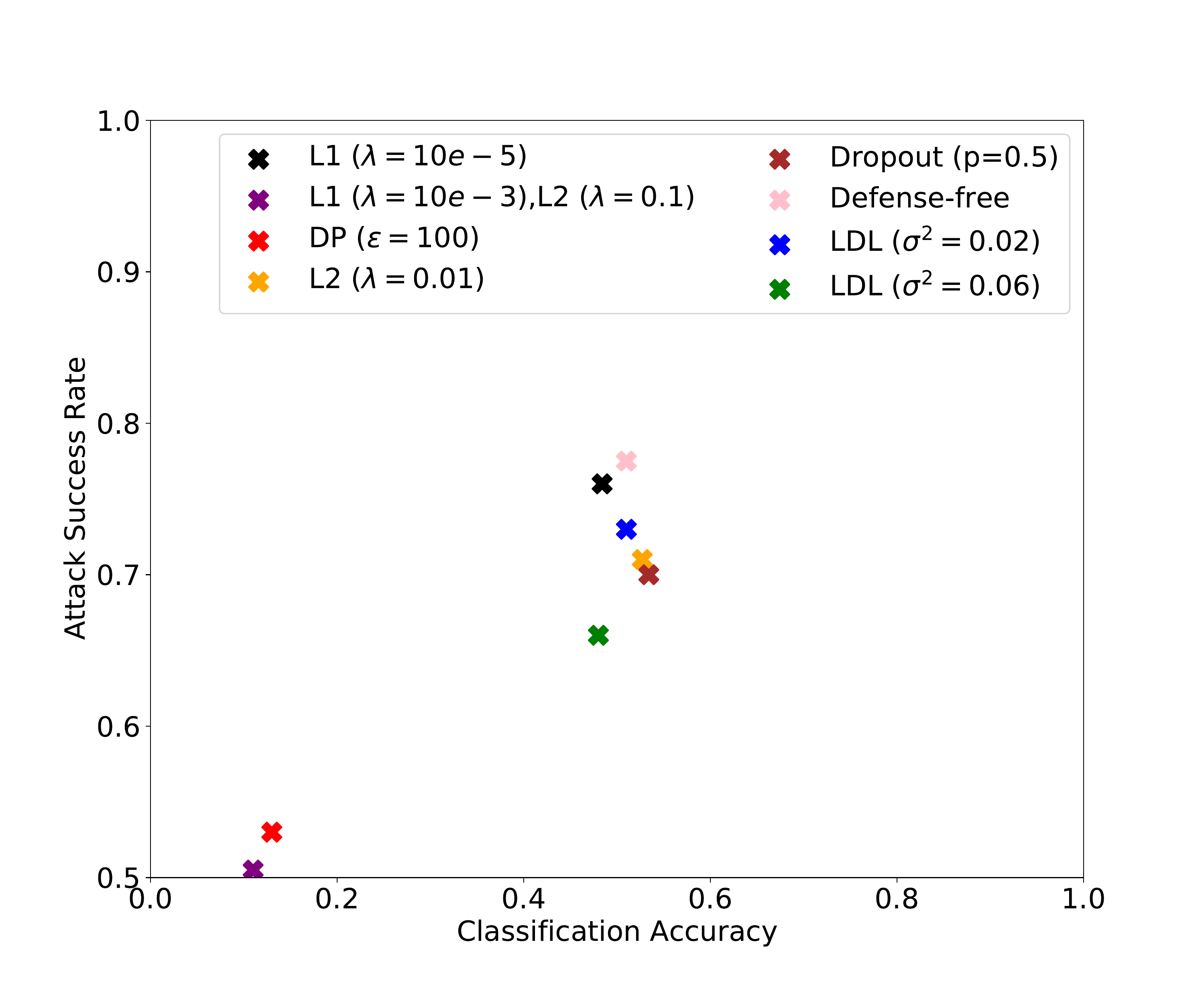}
    \caption{Classification accuracy and $ASR$ for defenses used against a LAB MIA on CIFAR-10 with training set size of $2000$. We observe that LDL with a small value of $\sigma=0.02$ results in a lower $ASR$ while maintaining classification accuracy compared to 
    defenses that aim to reduce model overfitting at the training stage. Using LDL with a larger value of $\sigma=0.06$ reduces the $ASR$, but also lowers classification accuracy. 
    LDL achieves these values without retraining the DNN model.}
    \label{fig:defenses}
\end{figure}

\noindent{\bf Non-Image Datasets: } 
We evaluate LDL on three non-image datasets~\cite{shokri2017membership, icml2021, ccs2021} (Purchase, Texas, and Location) 
characterized by binary-valued features. 
Discrete feature values prevent addition of Gaussian noise to obtain noisy variants of samples as we have done for experiments on image-based datasets in this section. 
Instead, we use a Bernoulli distribution with parameter $p \in (0,1)$ to generate $10000$ noisy variants of each sample. 
The choice of the value of $p$ is informed by the number of features of samples in the dataset. 
%
%
%
Table~\ref{tab:nonimage} indicates that a higher value of the parameter $p$ of the Bernoulli distribution will cause a reduction in classification accuracy and $ASR$. 
A possible reason for this is that the addition of Bernoulli noise can `flip' the values of features from $0$ to $1$ or vice-versa. 
This has a more drastic effect on samples when features are discrete compared to the case when they are continuous valued. 
\begin{table}[htb]
\vspace{-5mm}
    \begin{tabular}{|c|c|c|c|} \hline
      Dataset & Defense & \shortstack{ Classification\\ Accuracy } & ASR \\ \hline
      \multirow{2}{*}{\rotatebox{0}{Purchase}} & Defense-free&  $100\% (65.5\%)$  & $85.75\%$\\ \cline{2-4}
      & LDL ($p=0.0166$)& $59 (13.5 \%)$& $69\%$  \\ \hline
    
      \multirow{3}{*}{\rotatebox{0}{Location}} & Defense-free &$100\%  (55.0\%)$  & $92.5\%$\\ \cline{2-4}
      & LDL ($p=0.0022$)& $100 (30\%)$& $91.25\%$  \\ \cline{2-4}
      & LDL ($p=0.011$)& $98.5\%(15.5)$ & $80.25\%$ \\ \hline

      \multirow{3}{*}{\rotatebox{0}{Texas}} & Defense-free&  $100\% (54.5\%)$  & $83.5\%$\\ \cline{2-4}
      & LDL ($p=0.00016$)& $54.5\% (26.5\%)$ & $69.5\%$  \\ \cline{2-4}
      & LDL ($p=0.00032$)&$36\% (17\%)$ &  $62\%$\\ \hline
    \end{tabular}
    \caption{Classification accuracy for members (nonmembers) and $ASR$ when an adversary carries out a LAB MIA on the Purchase, Texas, and Location datasets with binary-valued features. 
    Perturbed variants of each sample is generated by adding a $Bernoulli(p)$ noise, where the values of the parameter $p$ are determined by the number of features corresponding to samples in the respective dataset. We observe that the ASR is largest when no defense is used against the MIA. When LDL is deployed as a defense, larger values of $p$ are more effective in reducing the value of $ASR$. However, this comes at the cost of simultaneously reducing the classification accuracy. \label{tab:nonimage}}
    \vspace{-10mm}
\end{table}

%

\subsection{Evaluation of Robustness of LDL to Noise Perturbations
}\label{sec:FGS}

In mounting LAB MIAs~\cite{icml2021, ccs2021}, an adversary uses the insight that 
nonmember samples require a smaller amount of additive noise in order to trigger a misclassification by the DNN model compared to members. 
The 
sphere of label-invariance constructed by LDL ensures that a sample and all its perturbed variants 
within this sphere will have the same label. 
In this section, we empirically evaluate the robustness of LDL to different levels of noise perturbations for varying values of radius of label-invariance $\sigma^2$. 
%
%

We use a SOTA adversarial example generation technique called Fast Gradient Sign (FGS)~\cite{goodfellow2014explaining} 
to study 
the 
perturbation required to misclassify samples after LDL is deployed. 
FGS is known to typically require only a single step 
%
to determine the 
magnitude of perturbation required for misclassification~\cite{goodfellow2014explaining}. 
FGS adds noise perturbation to a `clean' sample $x$ which has the true label $y$ so that a classifier will identify that $x$ belongs to a class $y' \neq y$. 
FGS accomplishes this goal by using 
the sign of the gradient of a loss function $\mathcal{L}((F(x,\theta),y)$ to compute the needed adversarial input $x_{adv}:=x+\epsilon \times sign(\nabla\mathcal{L}(F(x,\theta),y))$, 
%
%
 where 
 $\epsilon$ is the magnitude of perturbation noise added to $x$. 
 
Figure~\ref{fig:fgs} presents misclassification rates of samples of CIFAR-10 with different sizes of training sets with increasing amounts of noise perturbation $\epsilon$. 
%
We compare the defense-free case and when LDL is deployed using $\sigma^2 = \{0.02, 0.04, 0.06\}$. 
We observe that deploying LDL ensures that the misclassification rate accomplished by the addition of adversarial noise is similar for both members (training dataset) and nonmembers (testing dataset). 
%


The effect of deploying LDL on classification accuracy will depend on the radius of the sphere constructed by LDL (that depends on $\sigma^2$) and the original model itself. 
Figure~\ref{fig:accvssigma} shows the change in classification accuracy of nonmember samples as $\sigma^2$ increases. We observe that, as expected, the classification accuracy decreases with increasing $\sigma^2$. 
The accuracy drop is sharper for models that are known to be more prone to overfitting, such as GTSRB and Face. 

\subsection{LDL vs. SOTA Defenses against MIAs} \label{sec:defenses}

Defenses against MIAs can be divided into two broad categories: (i) distorting outputs of models that are already overfitted, and (ii) training models in a manner to prevent overfitting. 

In (i), the impact of MIAs is mitigated by distorting outputs of an overfitted DNN model. 
Recent methods that have successfully used this approach include Memguard~\cite{jia2019memguard} and confidence masking~\cite{confemasking1}. 
Although such methods do not require the DNN model to be retrained, it has been demonstrated in~\cite{icml2021,ccs2021} that they are not effective against an adversary employing a LAB MIA. 

In (ii), overfitting of the model is reduced at the training stage; however, this could result in a significant reduction in accuracy. 
Overfitting can be reduced through methods such as dropout~\cite{srivastava2014dropout}, L1- and L2- regularization~\cite{nasr2018machine} or differential privacy (DP) techniques~\cite{abadi2016deep}. 
We compare the classification accuracy and ASR of LDL with DP-SGD with a privacy budget $\epsilon=100$~\cite{abadi2016deep}
, dropout regularization with dropout probability $p = 0.8$~\cite{srivastava2014dropout}, and L1-regularization with weights $\lambda = 10^{-5},10^{-3}$ and L2-regularization with weights $\lambda = 0.01, 0.1$~\cite{nasr2018machine}. 
Fig.~\ref{fig:defenses} shows that LDL is able to reduce the value of $ASR$ while maintaining classification accuracy compared to defense mechanisms that aim to reduce overfitting. 
LDL achieves these performance levels \emph{without retraining the DNN model}.

\begin{figure}
    \centering
    \includegraphics[trim={2cm 2cm 2cm 2cm},scale=0.23]{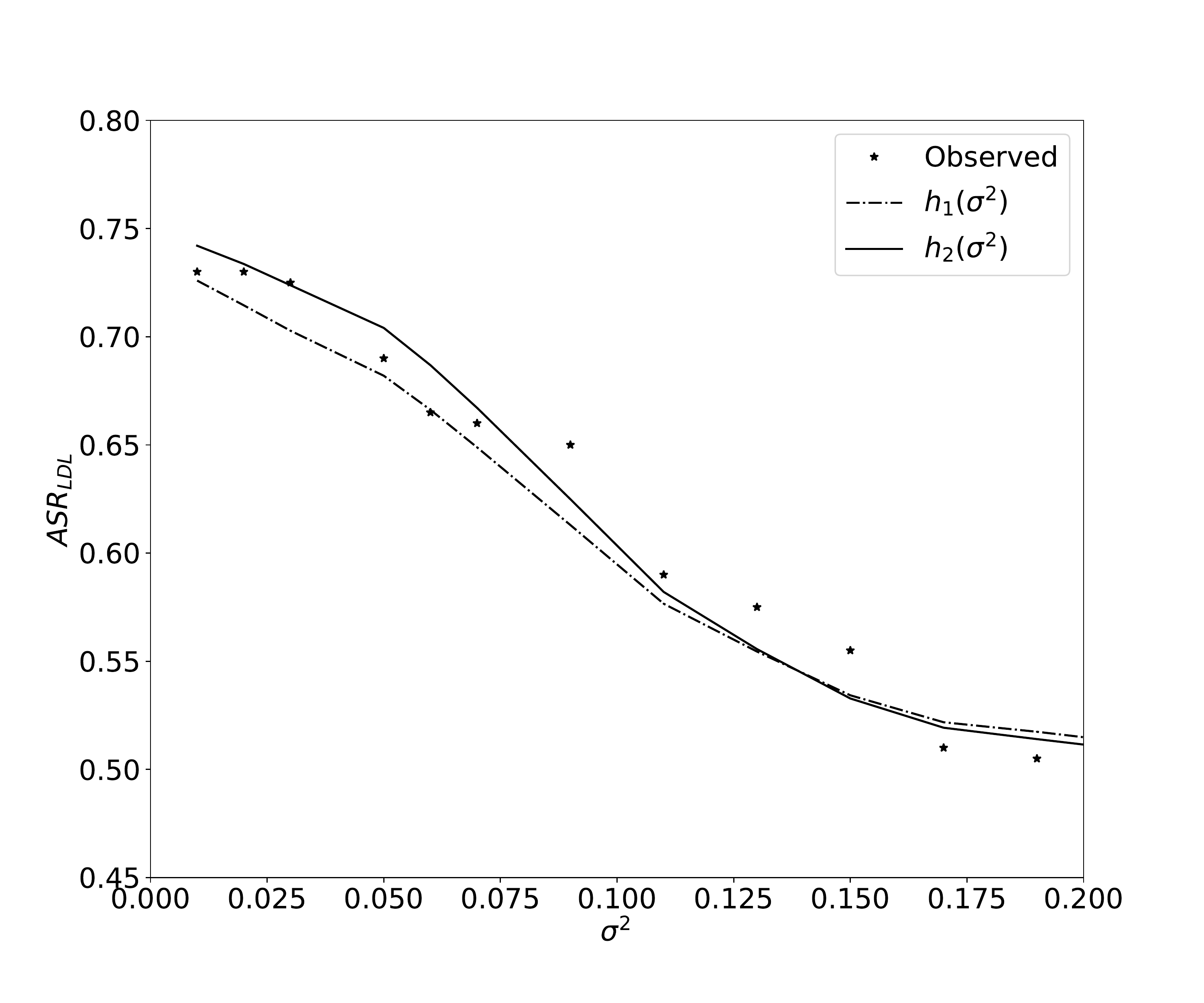}
    \caption{
    Change in the value of $ASR_{LDL}$ with increasing $\sigma^2$ for 
    $h_1(\sigma^2):=1-\exp{(-4.88\sigma^2)}$, and $h_2(\sigma^2):= 1.34(1+\exp{(-12.93(\sigma^2-0.1))^{-1}}$. The black dots indicate the value of $ASR_{LDL}$ from our experimental evaluations. 
    We observe that both choices of the function $h(\sigma^2)$ to compute $ASR_{LDL}$ in Eqn. (\ref{eqn:asrldl}) are consistent with the observed values.
    } \vspace{-5mm}
    \label{fig:asrvsclassification}
\end{figure}

\subsection{Observed vs. Analytical $ASR_{LDL}$ Values
} \label{sec:selectingsigma}

We examine the correctness of Eqn. (\ref{eqn:asrldl}), which provides an analytical characterization of the success rate of an adversary carrying out a LAB MIA when LDL is deployed as a defense. 
Fig.~\ref{fig:asrvsclassification} illustrates the effect of the choice of the parameter $\sigma^2$ on the value of $ASR_{LDL}$ obtained using: 
(i) our experiments (black dots), 
(ii) curve fitting to the functions $h_1(\sigma^2):=1-\exp{(-4.88\sigma^2)}$ (dashed line) and $h_2(\sigma^2):= 1.34(1+\exp{(-12.93(\sigma^2-0.1)))^{-1}}$ (solid line) in Eqn. (\ref{eqn:asrldl}). 
We observe that results of our experiments agree with both choices of $h(\cdot)$ for values of $\sigma^2 \in (0, 0.2)$.

\section{Discussion}
\label{sec:discussion}

\noindent{\bf Lack of Formal Guarantee:} 
LDL provides a way to protect against LAB MIAs without retraining DNN models, and compares favorably against other defenses that require DNNs to be retrained (Fig.~\ref{fig:defenses}). 
An adversary carrying out a LAB MIA only has black-box (input and output) access to the DNN model. 
It remains an open problem if additional assumptions on DNN structure can be used to provide a precise characterization on the level of protection offered by LDL against LAB MIAs. 
A potential approach is to leverage recent results that used polynomial functions to provide bounds on the error in estimating nonlinear activation functions of DNN layers~\cite{hesamifard2019deep}.

\noindent{\bf LDL on Discrete Datasets:} 
We evaluate LDL on three datasets--Texas, Purchase, and Location-- with discrete-valued features. 
In these cases, we used a $Ber(p)$ noise to generate perturbed variants of samples (Table~\ref{tab:nonimage}). 
Although using LDL was effective in reducing the value of $ASR$ for these datasets, it resulted in a simultaneous reduction in classification accuracy. 
The lower classification accuracy could occur as a result of the additive noise \emph{flipping} the label of the sample. 
In contrast, adding a Gaussian noise to samples in the four datasets with continuous-valued features (CIFAR-10, CIFAR-100, GTSRB, Face) did not change the label of samples as significantly. 
A possible way to maintain classification accuracy when LDL is used on datasets with discrete-valued features is to investigate the use of a noise generated from a discrete Gaussian distribution~\cite{canonne2020discrete}. 


\noindent {\bf Other Choices of $h(\sigma^2)$:} 
In this paper, we defined the set of properties that $h(sigma^2)$ must possess and presented two choices
that defined the value of $ASR_{LDL}$ in Eqn.~(\ref{eqn:asrldl}), and showed that they were both consistent with our experimental observations (Fig.~\ref{fig:asrvsclassification}). 
There can be a class of solutions of $h(\sigma^2)$ 
which satisfy the properties identified in Sec.~\ref{sec:intution}. 
Examining and characterizing the effect of the types and sizes of datasets used in experiments on the choice of $h(\sigma^2)$ remains an open research problem. 


\section{Related Work}
\label{sec:relatedwork}


\noindent {\bf Membership Inference Attacks (MIAs) and Defenses:} MIAs aim to determine whether a sample belongs to the training set of a classifier or not. 
There are three different approaches that an adversary with  black-box access to the outputs of its target model can use to distinguish members and non-members. 
First, the adversary can directly examine the output of the model by studying confidence values or calculating loss values returned by the model~\cite{yeom2018privacy} . 
Second, the adversary can learn a model to analyze the difference between outputs of the model for members and nonmembers~\cite{shokri2017membership,ye2021shokri2} . 
Third, the adversary can examine outputs of the model for perturbed variants of each input sample~\cite{PETS2021}. 
The MIAs of this type presume that models behave differently for perturbed variants of members and nonmembers. 
However, all the above three types of MIAs are confidence-score based. 
A new class of label-based MIAs was recently proposed in~\cite{icml2021, ccs2021} where an adversary only required knowledge of labels of samples rather than the associated confidence scores. 
These works demonstrated that defenses against confidence-score based MIAs will not be adequate. 
Thus, there is a need to design effective defenses against label-based MIAs, which is a gap that our work in this paper seeks to bridge. 

Recently, two defenses called Memguard and Attriguard~\cite{jia2018attriguard, jia2019memguard, yang2020defending} were designed to protect against MIAs that use confidence scores of outputs of a target model. 
These methods used the insight that perturbing the output of target models will be effective in mitigating the impact of confidence score-based MIAs~\cite{jia2018attriguard, jia2019memguard, yang2020defending}. 
However, they will not be effective against label-based MIAs, since these attacks use only predicted labels from the target classifier and do not require confidence scores.

\noindent{\bf Learning Differentially Private Models:} 
DNN models are known to suffer from a problem of overfitting~\cite{yeom2018privacy}. 
While this shortcoming can be mitigated through the use of regularization and dropout layers~\cite{li2021membership,nasr2018machine, srivastava2014dropout}, these methods do not provide formal guarantees of privacy. 
One line of research has proposed to use differential privacy to learn models that have provable guarantees on privacy~\cite{abadi2016deep, bassily2014private, bu2020deep,mcmahan2017learning,chaudhuri2011differentially, iyengar2019towards}. 
Although such techniques have successfully overcome the challenge of overfitting, differentially private algorithms require more computational resources, and often demonstrate slow convergence~\cite{mcsherry2009privacy}. 
A recent work~\cite{papernot2018scalable} proposed a model called PATE, which provided strong privacy guarantees through learning a model that mimicked the output of the target model. 
However, all these methods require retraining the target model, which is computationally expensive and often impractical. LDL, on the other hand, is a module that can be incorporated into already-deployed DNN models without needing to retrain them.

\section{Conclusion}
\label{sec:conclusion}
In this paper, 
we proposed LDL, a light weight defense against SOTA~\cite{icml2021, ccs2021}  label-based MIAs (LAB MIAs) which does not require retraining of DNNs. 
LDL constructed a sphere of label invariance that created ambiguity which prevented an adversary from correctly identifying a sample as member or nonmember. 
We provided an analytical characterization of the effectiveness of an adversary carrying out a LAB MIA when LDL is deployed, and showed that its variation with the magnitude of noise required to misclassify samples was consistent with observations in our experiments. 
Extensive evaluations of LDL on seven datasets (CIFAR-10, CIFAR-100, GTSRB, Face, Purchase, Location, Texas) demonstrated that LDL successfully nullified SOTA~\cite{icml2021, ccs2021} LAB MIAs in each case. 
Our experiments also revealed that LDL compared favorably against other defenses against LAB MIAs that required the DNN to be retrained. 
\section*{Acknowledgment}

This work was supported by the US National Science Foundation and the Office of Naval Research via grants CNS-2153136 and N00014-20-1-2636 respectively. 
We acknowledge Prof. Andrew Clark from Washington University in St. Louis and Prof. Sukarno Mertoguno from Georgia Tech for insightful discussions.


\bibliographystyle{ACM-Reference-Format}
\bibliography{ref.bib}


\begin{thebibliography}{41}


\ifx \showCODEN    \undefined \def \showCODEN     #1{\unskip}     \fi
\ifx \showDOI      \undefined \def \showDOI       #1{#1}\fi
\ifx \showISBNx    \undefined \def \showISBNx     #1{\unskip}     \fi
\ifx \showISBNxiii \undefined \def \showISBNxiii  #1{\unskip}     \fi
\ifx \showISSN     \undefined \def \showISSN      #1{\unskip}     \fi
\ifx \showLCCN     \undefined \def \showLCCN      #1{\unskip}     \fi
\ifx \shownote     \undefined \def \shownote      #1{#1}          \fi
\ifx \showarticletitle \undefined \def \showarticletitle #1{#1}   \fi
\ifx \showURL      \undefined \def \showURL       {\relax}        \fi
\providecommand\bibfield[2]{#2}
\providecommand\bibinfo[2]{#2}
\providecommand\natexlab[1]{#1}
\providecommand\showeprint[2][]{arXiv:#2}

\bibitem[Abadi et~al\mbox{.}(2016)]%
        {abadi2016deep}
\bibfield{author}{\bibinfo{person}{Martin Abadi}, \bibinfo{person}{Andy Chu},
  \bibinfo{person}{Ian Goodfellow}, \bibinfo{person}{H~Brendan McMahan},
  \bibinfo{person}{Ilya Mironov}, \bibinfo{person}{Kunal Talwar}, {and}
  \bibinfo{person}{Li Zhang}.} \bibinfo{year}{2016}\natexlab{}.
\newblock \showarticletitle{Deep learning with differential privacy}. In
  \bibinfo{booktitle}{\emph{Proceedings of the 2016 ACM SIGSAC Conference on
  Computer and Communications Security}}. \bibinfo{publisher}{ACM},
  \bibinfo{address}{New York, NY, USA}, \bibinfo{pages}{308--318}.
\newblock


\bibitem[Bassily et~al\mbox{.}(2014)]%
        {bassily2014private}
\bibfield{author}{\bibinfo{person}{Raef Bassily}, \bibinfo{person}{Adam Smith},
  {and} \bibinfo{person}{Abhradeep Thakurta}.} \bibinfo{year}{2014}\natexlab{}.
\newblock \showarticletitle{Private empirical risk minimization: Efficient
  algorithms and tight error bounds}. In \bibinfo{booktitle}{\emph{IEEE 55th
  Annual Symposium on Foundations of Computer Science}}. \bibinfo{address}{New
  York, NY, USA}, \bibinfo{pages}{464--473}.
\newblock


\bibitem[Bu et~al\mbox{.}(2020)]%
        {bu2020deep}
\bibfield{author}{\bibinfo{person}{Zhiqi Bu}, \bibinfo{person}{Jinshuo Dong},
  \bibinfo{person}{Qi Long}, {and} \bibinfo{person}{Weijie~J Su}.}
  \bibinfo{year}{2020}\natexlab{}.
\newblock \showarticletitle{Deep learning with {G}aussian differential
  privacy}.
\newblock \bibinfo{journal}{\emph{Harvard Data Science Review}}
  \bibinfo{volume}{2020}, \bibinfo{number}{23} (\bibinfo{year}{2020}).
\newblock


\bibitem[Canonne et~al\mbox{.}(2020)]%
        {canonne2020discrete}
\bibfield{author}{\bibinfo{person}{Cl{\'e}ment~L Canonne},
  \bibinfo{person}{Gautam Kamath}, {and} \bibinfo{person}{Thomas Steinke}.}
  \bibinfo{year}{2020}\natexlab{}.
\newblock \showarticletitle{The discrete gaussian for differential privacy}.
\newblock \bibinfo{journal}{\emph{Advances in Neural Information Processing
  Systems}}  \bibinfo{volume}{33} (\bibinfo{year}{2020}),
  \bibinfo{pages}{15676--15688}.
\newblock


\bibitem[Carlini et~al\mbox{.}(2019)]%
        {carlini2019secret}
\bibfield{author}{\bibinfo{person}{Nicholas Carlini}, \bibinfo{person}{Chang
  Liu}, \bibinfo{person}{{\'U}lfar Erlingsson}, \bibinfo{person}{Jernej Kos},
  {and} \bibinfo{person}{Dawn Song}.} \bibinfo{year}{2019}\natexlab{}.
\newblock \showarticletitle{The secret sharer: Evaluating and testing
  unintended memorization in neural networks}. In
  \bibinfo{booktitle}{\emph{USENIX Security}}. \bibinfo{publisher}{USENIX
  Association}, \bibinfo{address}{Berkeley, CA, USA},
  \bibinfo{pages}{267--284}.
\newblock


\bibitem[Carlini and Wagner(2017)]%
        {carlini2017adversarial}
\bibfield{author}{\bibinfo{person}{Nicholas Carlini} {and}
  \bibinfo{person}{David Wagner}.} \bibinfo{year}{2017}\natexlab{}.
\newblock \showarticletitle{Adversarial examples are not easily detected:
  Bypassing ten detection methods}. In \bibinfo{booktitle}{\emph{Proceedings of
  the 10th ACM workshop on Artificial Intelligence and Security}}.
  \bibinfo{publisher}{ACM}, \bibinfo{address}{New York, NY, USA},
  \bibinfo{pages}{3--14}.
\newblock


\bibitem[Chaudhuri et~al\mbox{.}(2011)]%
        {chaudhuri2011differentially}
\bibfield{author}{\bibinfo{person}{Kamalika Chaudhuri}, \bibinfo{person}{Claire
  Monteleoni}, {and} \bibinfo{person}{Anand~D Sarwate}.}
  \bibinfo{year}{2011}\natexlab{}.
\newblock \showarticletitle{Differentially private empirical risk
  minimization}.
\newblock \bibinfo{journal}{\emph{Journal of Machine Learning Research}}
  \bibinfo{volume}{12}, \bibinfo{number}{3} (\bibinfo{year}{2011}).
\newblock


\bibitem[Chen et~al\mbox{.}(2020)]%
        {chen2020hopskipjumpattack}
\bibfield{author}{\bibinfo{person}{Jianbo Chen}, \bibinfo{person}{Michael~I
  Jordan}, {and} \bibinfo{person}{Martin~J Wainwright}.}
  \bibinfo{year}{2020}\natexlab{}.
\newblock \showarticletitle{Hop{S}kip{J}ump{A}ttack: A query-efficient
  decision-based attack}. In \bibinfo{booktitle}{\emph{IEEE Symposium on
  Security and Privacy}}. \bibinfo{publisher}{IEEE}, \bibinfo{address}{New
  York, NY, USA}, \bibinfo{pages}{1277--1294}.
\newblock


\bibitem[Choquette-Choo et~al\mbox{.}(2021)]%
        {icml2021}
\bibfield{author}{\bibinfo{person}{Christopher~A Choquette-Choo},
  \bibinfo{person}{Florian Tramer}, \bibinfo{person}{Nicholas Carlini}, {and}
  \bibinfo{person}{Nicolas Papernot}.} \bibinfo{year}{2021}\natexlab{}.
\newblock \showarticletitle{Label-only membership inference attacks}. In
  \bibinfo{booktitle}{\emph{International Conference on Machine Learning}}.
  \bibinfo{publisher}{PMLR}, \bibinfo{pages}{1964--1974}.
\newblock


\bibitem[Cohen et~al\mbox{.}(2019)]%
        {cohen2019certified}
\bibfield{author}{\bibinfo{person}{Jeremy Cohen}, \bibinfo{person}{Elan
  Rosenfeld}, {and} \bibinfo{person}{Zico Kolter}.}
  \bibinfo{year}{2019}\natexlab{}.
\newblock \showarticletitle{Certified adversarial robustness via randomized
  smoothing}. In \bibinfo{booktitle}{\emph{International Conference on Machine
  Learning}}. \bibinfo{publisher}{PMLR}, \bibinfo{pages}{1310--1320}.
\newblock


\bibitem[Goodfellow et~al\mbox{.}(2015)]%
        {goodfellow2014explaining}
\bibfield{author}{\bibinfo{person}{Ian Goodfellow}, \bibinfo{person}{Jonathon
  Shlens}, {and} \bibinfo{person}{Christian Szegedy}.}
  \bibinfo{year}{2015}\natexlab{}.
\newblock \showarticletitle{Explaining and Harnessing Adversarial Examples}. In
  \bibinfo{booktitle}{\emph{Proc. Int. Conf. on Learning Representations}}.
\newblock


\bibitem[Hastie et~al\mbox{.}(2009)]%
        {hastie2009elements}
\bibfield{author}{\bibinfo{person}{Trevor Hastie}, \bibinfo{person}{Robert
  Tibshirani}, \bibinfo{person}{Jerome~H Friedman}, {and}
  \bibinfo{person}{Jerome~H Friedman}.} \bibinfo{year}{2009}\natexlab{}.
\newblock \bibinfo{booktitle}{\emph{The elements of statistical learning:
  {D}ata mining, inference, and prediction}}. Vol.~\bibinfo{volume}{2}.
\newblock \bibinfo{publisher}{Springer}.
\newblock


\bibitem[Hesamifard et~al\mbox{.}(2019)]%
        {hesamifard2019deep}
\bibfield{author}{\bibinfo{person}{Ehsan Hesamifard}, \bibinfo{person}{Hassan
  Takabi}, {and} \bibinfo{person}{Mehdi Ghasemi}.}
  \bibinfo{year}{2019}\natexlab{}.
\newblock \showarticletitle{Deep neural networks classification over encrypted
  data}. In \bibinfo{booktitle}{\emph{Proceedings of the ACM Conference on Data
  and Application Security and Privacy}}. \bibinfo{publisher}{ACM},
  \bibinfo{address}{New York, NY, USA}, \bibinfo{pages}{97--108}.
\newblock


\bibitem[Houben et~al\mbox{.}(2013)]%
        {GTSRB013}
\bibfield{author}{\bibinfo{person}{Sebastian Houben}, \bibinfo{person}{Johannes
  Stallkamp}, \bibinfo{person}{Jan Salmen}, \bibinfo{person}{Marc Schlipsing},
  {and} \bibinfo{person}{Christian Igel}.} \bibinfo{year}{2013}\natexlab{}.
\newblock \showarticletitle{Detection of Traffic Signs in Real-World Images:
  The {G}erman {T}raffic {S}ign {D}etection {B}enchmark}. In
  \bibinfo{booktitle}{\emph{International Joint Conference on Neural
  Networks}}. \bibinfo{publisher}{IEEE}, \bibinfo{address}{New York, NY, USA}.
\newblock


\bibitem[Hu et~al\mbox{.}(2020)]%
        {DNNsCapabilty}
\bibfield{author}{\bibinfo{person}{Junyan Hu}, \bibinfo{person}{Hanlin Niu},
  \bibinfo{person}{Joaquin Carrasco}, \bibinfo{person}{Barry Lennox}, {and}
  \bibinfo{person}{Farshad Arvin}.} \bibinfo{year}{2020}\natexlab{}.
\newblock \showarticletitle{Voronoi-Based Multi-Robot Autonomous Exploration in
  Unknown Environments via Deep Reinforcement Learning}.
\newblock \bibinfo{journal}{\emph{IEEE Transactions on Vehicular Technology}}
  \bibinfo{volume}{69}, \bibinfo{number}{12} (\bibinfo{year}{2020}),
  \bibinfo{pages}{14413--14423}.
\newblock
\urldef\tempurl%
\url{https://doi.org/10.1109/TVT.2020.3034800}
\showDOI{\tempurl}


\bibitem[Huang et~al\mbox{.}(2012)]%
        {Huang2012a}
\bibfield{author}{\bibinfo{person}{Gary~B. Huang}, \bibinfo{person}{Marwan
  Mattar}, \bibinfo{person}{Honglak Lee}, {and} \bibinfo{person}{Erik
  Learned-Miller}.} \bibinfo{year}{2012}\natexlab{}.
\newblock \showarticletitle{Learning to Align from Scratch}. In
  \bibinfo{booktitle}{\emph{Advances in Neural Information Processing
  Systems}}, Vol.~\bibinfo{volume}{25}. \bibinfo{publisher}{Curran Associates},
  \bibinfo{address}{Red Hook, NY, USA}.
\newblock


\bibitem[Iyengar et~al\mbox{.}(2019)]%
        {iyengar2019towards}
\bibfield{author}{\bibinfo{person}{Roger Iyengar}, \bibinfo{person}{Joseph~P
  Near}, \bibinfo{person}{Dawn Song}, \bibinfo{person}{Om Thakkar},
  \bibinfo{person}{Abhradeep Thakurta}, {and} \bibinfo{person}{Lun Wang}.}
  \bibinfo{year}{2019}\natexlab{}.
\newblock \showarticletitle{Towards practical differentially private convex
  optimization}. In \bibinfo{booktitle}{\emph{IEEE Symposium on Security and
  Privacy (SP)}}. \bibinfo{address}{New York, NY, USA},
  \bibinfo{pages}{299--316}.
\newblock


\bibitem[Jayaraman et~al\mbox{.}(2021)]%
        {PETS2021}
\bibfield{author}{\bibinfo{person}{Bargav Jayaraman}, \bibinfo{person}{Lingxiao
  Wang}, \bibinfo{person}{Katherine Knipmeyer}, \bibinfo{person}{Quanquan Gu},
  {and} \bibinfo{person}{David Evans}.} \bibinfo{year}{2021}\natexlab{}.
\newblock \showarticletitle{Revisiting membership inference under realistic
  assumptions}.
\newblock \bibinfo{journal}{\emph{Privacy Enhancing Technology Symposium
  (PETS)}}  \bibinfo{volume}{2021} (\bibinfo{year}{2021}),
  \bibinfo{pages}{348--468}.
\newblock


\bibitem[Jia and Gong(2018)]%
        {jia2018attriguard}
\bibfield{author}{\bibinfo{person}{Jinyuan Jia} {and}
  \bibinfo{person}{Neil~Zhenqiang Gong}.} \bibinfo{year}{2018}\natexlab{}.
\newblock \showarticletitle{Attriguard: {A} practical defense against attribute
  inference attacks via adversarial machine learning}. In
  \bibinfo{booktitle}{\emph{USENIX Security}}. \bibinfo{publisher}{USENIX
  Association}, \bibinfo{address}{Berkeley, CA, USA},
  \bibinfo{pages}{513--529}.
\newblock


\bibitem[Jia et~al\mbox{.}(2019)]%
        {jia2019memguard}
\bibfield{author}{\bibinfo{person}{Jinyuan Jia}, \bibinfo{person}{Ahmed Salem},
  \bibinfo{person}{Michael Backes}, \bibinfo{person}{Yang Zhang}, {and}
  \bibinfo{person}{Neil~Zhenqiang Gong}.} \bibinfo{year}{2019}\natexlab{}.
\newblock \showarticletitle{Memguard: Defending against black-box membership
  inference attacks via adversarial examples}. In
  \bibinfo{booktitle}{\emph{Proceedings of the ACM SIGSAC Conference on
  Computer and Communications Security}}. \bibinfo{publisher}{ACM},
  \bibinfo{address}{New York, NY, USA}, \bibinfo{pages}{259--274}.
\newblock


\bibitem[Krizhevsky and Hinton(2009)]%
        {krizhevsky2009CIFARs}
\bibfield{author}{\bibinfo{person}{Alex Krizhevsky} {and}
  \bibinfo{person}{Geoffrey Hinton}.} \bibinfo{year}{2009}\natexlab{}.
\newblock \bibinfo{booktitle}{\emph{Learning multiple layers of features from
  tiny images}}.
\newblock \bibinfo{type}{{T}echnical {R}eport}~0.
  \bibinfo{institution}{University of Toronto}, \bibinfo{address}{Toronto,
  Ontario}.
\newblock


\bibitem[Li et~al\mbox{.}(2021)]%
        {li2021membership}
\bibfield{author}{\bibinfo{person}{Jiacheng Li}, \bibinfo{person}{Ninghui Li},
  {and} \bibinfo{person}{Bruno Ribeiro}.} \bibinfo{year}{2021}\natexlab{}.
\newblock \showarticletitle{Membership Inference Attacks and Defenses in
  Classification Models}. In \bibinfo{booktitle}{\emph{Proceedings of the ACM
  Conference on Data and Application Security and Privacy}}.
  \bibinfo{publisher}{ACM}, \bibinfo{address}{New York, NY, USA},
  \bibinfo{pages}{5--16}.
\newblock


\bibitem[Li and Zhang(2021)]%
        {ccs2021}
\bibfield{author}{\bibinfo{person}{Zheng Li} {and} \bibinfo{person}{Yang
  Zhang}.} \bibinfo{year}{2021}\natexlab{}.
\newblock \showarticletitle{Membership leakage in label-only exposures}. In
  \bibinfo{booktitle}{\emph{Proceedings of the ACM SIGSAC Conference on
  Computer and Communications Security}}. \bibinfo{publisher}{ACM},
  \bibinfo{address}{New York, NY, USA}, \bibinfo{pages}{880--895}.
\newblock


\bibitem[Liu et~al\mbox{.}(2021)]%
        {liu2021machine}
\bibfield{author}{\bibinfo{person}{Bo Liu}, \bibinfo{person}{Ming Ding},
  \bibinfo{person}{Sina Shaham}, \bibinfo{person}{Wenny Rahayu},
  \bibinfo{person}{Farhad Farokhi}, {and} \bibinfo{person}{Zihuai Lin}.}
  \bibinfo{year}{2021}\natexlab{}.
\newblock \showarticletitle{When machine learning meets privacy: {A} survey and
  outlook}.
\newblock \bibinfo{journal}{\emph{ACM Computing Surveys (CSUR)}}
  \bibinfo{volume}{54}, \bibinfo{number}{2} (\bibinfo{year}{2021}),
  \bibinfo{pages}{1--36}.
\newblock


\bibitem[Liu et~al\mbox{.}(2018)]%
        {liu2018survey}
\bibfield{author}{\bibinfo{person}{Qiang Liu}, \bibinfo{person}{Pan Li},
  \bibinfo{person}{Wentao Zhao}, \bibinfo{person}{Wei Cai},
  \bibinfo{person}{Shui Yu}, {and} \bibinfo{person}{Victor~CM Leung}.}
  \bibinfo{year}{2018}\natexlab{}.
\newblock \showarticletitle{A survey on security threats and defensive
  techniques of machine learning: {A} data driven view}.
\newblock \bibinfo{journal}{\emph{IEEE Access}}  \bibinfo{volume}{6}
  (\bibinfo{year}{2018}), \bibinfo{pages}{12103--12117}.
\newblock


\bibitem[Liu et~al\mbox{.}(2020)]%
        {liu2020privacy}
\bibfield{author}{\bibinfo{person}{Ximeng Liu}, \bibinfo{person}{Lehui Xie},
  \bibinfo{person}{Yaopeng Wang}, \bibinfo{person}{Jian Zou},
  \bibinfo{person}{Jinbo Xiong}, \bibinfo{person}{Zuobin Ying}, {and}
  \bibinfo{person}{Athanasios~V Vasilakos}.} \bibinfo{year}{2020}\natexlab{}.
\newblock \showarticletitle{Privacy and security issues in deep learning: {A}
  survey}.
\newblock \bibinfo{journal}{\emph{IEEE Access}}  \bibinfo{volume}{9}
  (\bibinfo{year}{2020}), \bibinfo{pages}{4566--4593}.
\newblock


\bibitem[McMahan et~al\mbox{.}(2017)]%
        {mcmahan2017learning}
\bibfield{author}{\bibinfo{person}{H~Brendan McMahan}, \bibinfo{person}{Daniel
  Ramage}, \bibinfo{person}{Kunal Talwar}, {and} \bibinfo{person}{Li Zhang}.}
  \bibinfo{year}{2017}\natexlab{}.
\newblock \bibinfo{title}{Learning differentially private recurrent language
  models}.
\newblock
\newblock
\newblock
\shownote{arXiv preprint arXiv:1710.06963}.


\bibitem[McSherry(2009)]%
        {mcsherry2009privacy}
\bibfield{author}{\bibinfo{person}{Frank~D McSherry}.}
  \bibinfo{year}{2009}\natexlab{}.
\newblock \showarticletitle{Privacy integrated queries: {A}n extensible
  platform for privacy-preserving data analysis}. In
  \bibinfo{booktitle}{\emph{Proceedings of the ACM SIGMOD International
  Conference on Management of Data}}. \bibinfo{publisher}{ACM},
  \bibinfo{address}{New York, NY, USA}, \bibinfo{pages}{19--30}.
\newblock


\bibitem[Murphy(2012)]%
        {murphy2012machine}
\bibfield{author}{\bibinfo{person}{Kevin~P Murphy}.}
  \bibinfo{year}{2012}\natexlab{}.
\newblock \bibinfo{booktitle}{\emph{Machine learning: A probabilistic
  perspective}}.
\newblock \bibinfo{publisher}{MIT Press}, \bibinfo{address}{Cambridge, MA,
  USA}.
\newblock


\bibitem[Nasr et~al\mbox{.}(2018)]%
        {nasr2018machine}
\bibfield{author}{\bibinfo{person}{Milad Nasr}, \bibinfo{person}{Reza Shokri},
  {and} \bibinfo{person}{Amir Houmansadr}.} \bibinfo{year}{2018}\natexlab{}.
\newblock \showarticletitle{Machine learning with membership privacy using
  adversarial regularization}. In \bibinfo{booktitle}{\emph{ACM SIGSAC
  Conference on Computer and Communications Security}}. \bibinfo{address}{New
  York, NY, USA}, \bibinfo{pages}{634--646}.
\newblock


\bibitem[Papernot et~al\mbox{.}(2018)]%
        {papernot2018scalable}
\bibfield{author}{\bibinfo{person}{Nicolas Papernot}, \bibinfo{person}{Shuang
  Song}, \bibinfo{person}{Ilya Mironov}, \bibinfo{person}{Ananth Raghunathan},
  \bibinfo{person}{Kunal Talwar}, {and} \bibinfo{person}{{\'U}lfar
  Erlingsson}.} \bibinfo{year}{2018}\natexlab{}.
\newblock \bibinfo{title}{Scalable private learning with PATE}.
\newblock
\newblock
\newblock
\shownote{arXiv preprint arXiv:1802.08908}.


\bibitem[Rigaki and Garcia(2020)]%
        {rigaki2020survey}
\bibfield{author}{\bibinfo{person}{Maria Rigaki} {and}
  \bibinfo{person}{Sebastian Garcia}.} \bibinfo{year}{2020}\natexlab{}.
\newblock \bibinfo{title}{A survey of privacy attacks in machine learning}.
\newblock
\newblock
\newblock
\shownote{arXiv preprint arXiv:2007.07646}.


\bibitem[Shokri et~al\mbox{.}(2017)]%
        {shokri2017membership}
\bibfield{author}{\bibinfo{person}{Reza Shokri}, \bibinfo{person}{Marco
  Stronati}, \bibinfo{person}{Congzheng Song}, {and} \bibinfo{person}{Vitaly
  Shmatikov}.} \bibinfo{year}{2017}\natexlab{}.
\newblock \showarticletitle{Membership inference attacks against machine
  learning models}. In \bibinfo{booktitle}{\emph{IEEE Symposium on Security and
  Privacy (SP)}}. \bibinfo{publisher}{IEEE}, \bibinfo{address}{New York, NY,
  USA}, \bibinfo{pages}{3--18}.
\newblock


\bibitem[Srivastava et~al\mbox{.}(2014)]%
        {srivastava2014dropout}
\bibfield{author}{\bibinfo{person}{Nitish Srivastava},
  \bibinfo{person}{Geoffrey Hinton}, \bibinfo{person}{Alex Krizhevsky},
  \bibinfo{person}{Ilya Sutskever}, {and} \bibinfo{person}{Ruslan
  Salakhutdinov}.} \bibinfo{year}{2014}\natexlab{}.
\newblock \showarticletitle{Dropout: A simple way to prevent neural networks
  from overfitting}.
\newblock \bibinfo{journal}{\emph{The Journal of Machine Learning Research}}
  \bibinfo{volume}{15}, \bibinfo{number}{1} (\bibinfo{year}{2014}),
  \bibinfo{pages}{1929--1958}.
\newblock


\bibitem[Van~der Maaten and Hinton(2008)]%
        {van2008visualizingtsne}
\bibfield{author}{\bibinfo{person}{Laurens Van~der Maaten} {and}
  \bibinfo{person}{Geoffrey Hinton}.} \bibinfo{year}{2008}\natexlab{}.
\newblock \showarticletitle{Visualizing data using t-{SNE}}.
\newblock \bibinfo{journal}{\emph{Journal of Machine Learning Research}}
  \bibinfo{volume}{9}, \bibinfo{number}{11} (\bibinfo{year}{2008}).
\newblock


\bibitem[Veale et~al\mbox{.}(2018)]%
        {veale2018awhyMIA}
\bibfield{author}{\bibinfo{person}{Michael Veale}, \bibinfo{person}{Reuben
  Binns}, {and} \bibinfo{person}{Lilian Edwards}.}
  \bibinfo{year}{2018}\natexlab{}.
\newblock \showarticletitle{Algorithms that remember: Model inversion attacks
  and data protection law}.
\newblock \bibinfo{journal}{\emph{Philosophical Transactions of the Royal
  Society A}} \bibinfo{volume}{376}, \bibinfo{number}{2133}
  (\bibinfo{year}{2018}).
\newblock


\bibitem[Yang et~al\mbox{.}(2020)]%
        {yang2020defending}
\bibfield{author}{\bibinfo{person}{Ziqi Yang}, \bibinfo{person}{Bin Shao},
  \bibinfo{person}{Bohan Xuan}, \bibinfo{person}{Ee-Chien Chang}, {and}
  \bibinfo{person}{Fan Zhang}.} \bibinfo{year}{2020}\natexlab{}.
\newblock \bibinfo{title}{Defending model inversion and membership inference
  attacks via prediction purification}.
\newblock
\newblock
\newblock
\shownote{arXiv preprint arXiv:2005.03915}.


\bibitem[Ye et~al\mbox{.}(2022a)]%
        {confemasking1}
\bibfield{author}{\bibinfo{person}{Dayong Ye}, \bibinfo{person}{Sheng Shen},
  \bibinfo{person}{Tianqing Zhu}, \bibinfo{person}{Bo Liu}, {and}
  \bibinfo{person}{Wanlei Zhou}.} \bibinfo{year}{2022}\natexlab{a}.
\newblock \showarticletitle{One Parameter Defense - Defending against Data
  Inference Attacks via Differential Privacy}.
\newblock \bibinfo{journal}{\emph{IEEE Transactions on Information Forensics
  and Security}}  \bibinfo{volume}{17} (\bibinfo{year}{2022}),
  \bibinfo{pages}{1466--1480}.
\newblock
\urldef\tempurl%
\url{https://doi.org/10.1109/TIFS.2022.3163591}
\showDOI{\tempurl}


\bibitem[Ye et~al\mbox{.}(2022b)]%
        {ye2022one}
\bibfield{author}{\bibinfo{person}{Dayong Ye}, \bibinfo{person}{Sheng Shen},
  \bibinfo{person}{Tianqing Zhu}, \bibinfo{person}{Bo Liu}, {and}
  \bibinfo{person}{Wanlei Zhou}.} \bibinfo{year}{2022}\natexlab{b}.
\newblock \showarticletitle{One Parameter Defense: {D}efending against data
  inference attacks via differential privacy}.
\newblock \bibinfo{journal}{\emph{IEEE Trans. on Information Forensics and
  Security}}  \bibinfo{volume}{17} (\bibinfo{year}{2022}),
  \bibinfo{pages}{1466--1480}.
\newblock


\bibitem[Ye et~al\mbox{.}(2021)]%
        {ye2021shokri2}
\bibfield{author}{\bibinfo{person}{Jiayuan Ye}, \bibinfo{person}{Aadyaa Maddi},
  \bibinfo{person}{Sasi~Kumar Murakonda}, {and} \bibinfo{person}{Reza Shokri}.}
  \bibinfo{year}{2021}\natexlab{}.
\newblock \bibinfo{title}{Enhanced Membership Inference Attacks against Machine
  Learning Models}.
\newblock
\newblock
\newblock
\shownote{arXiv preprint arXiv:2111.09679}.


\bibitem[Yeom et~al\mbox{.}(2018)]%
        {yeom2018privacy}
\bibfield{author}{\bibinfo{person}{Samuel Yeom}, \bibinfo{person}{Irene
  Giacomelli}, \bibinfo{person}{Matt Fredrikson}, {and} \bibinfo{person}{Somesh
  Jha}.} \bibinfo{year}{2018}\natexlab{}.
\newblock \showarticletitle{Privacy risk in machine learning: Analyzing the
  connection to overfitting}. In \bibinfo{booktitle}{\emph{IEEE Computer
  Security Foundations Symposium}}. \bibinfo{publisher}{IEEE},
  \bibinfo{address}{New York, NY, USA}, \bibinfo{pages}{268--282}.
\newblock


\end{thebibliography}

\onecolumn
\section*{Appendix} 
This appendix presents additional results of our experiments evaluating LDL. 
We work with four image-based datasets-- CIFAR-10, CIFAR-100, GTSRB, and Face. 
We first examine the performance of LDL against a strong and a weak adversary carrying out a LAB MIA by estimating the minimum amount of noise required for misclassification using a SOTA adversarial model \emph{HopSkipJump}. 
Then, we show that using LDL ensures that misclassification rates accomplished by the addition of adversarial noise is similar for both members and nonmembers, 
thereby rendering these samples indistinguishable to an adversary. 

\begin{figure}[!h]
    \centering
    \begin{tabular}{>{\centering\arraybackslash} m{0.4cm} >{\centering\arraybackslash} m{8cm} >{\centering\arraybackslash} m{8cm}  }
     & {\bf Strong Adversary}&  {\bf Weak Adversary}\\
        \rotatebox{90}{GTSRB} &
        \includegraphics[trim={2cm 0.5cm 2cm 2cm},scale=0.15]{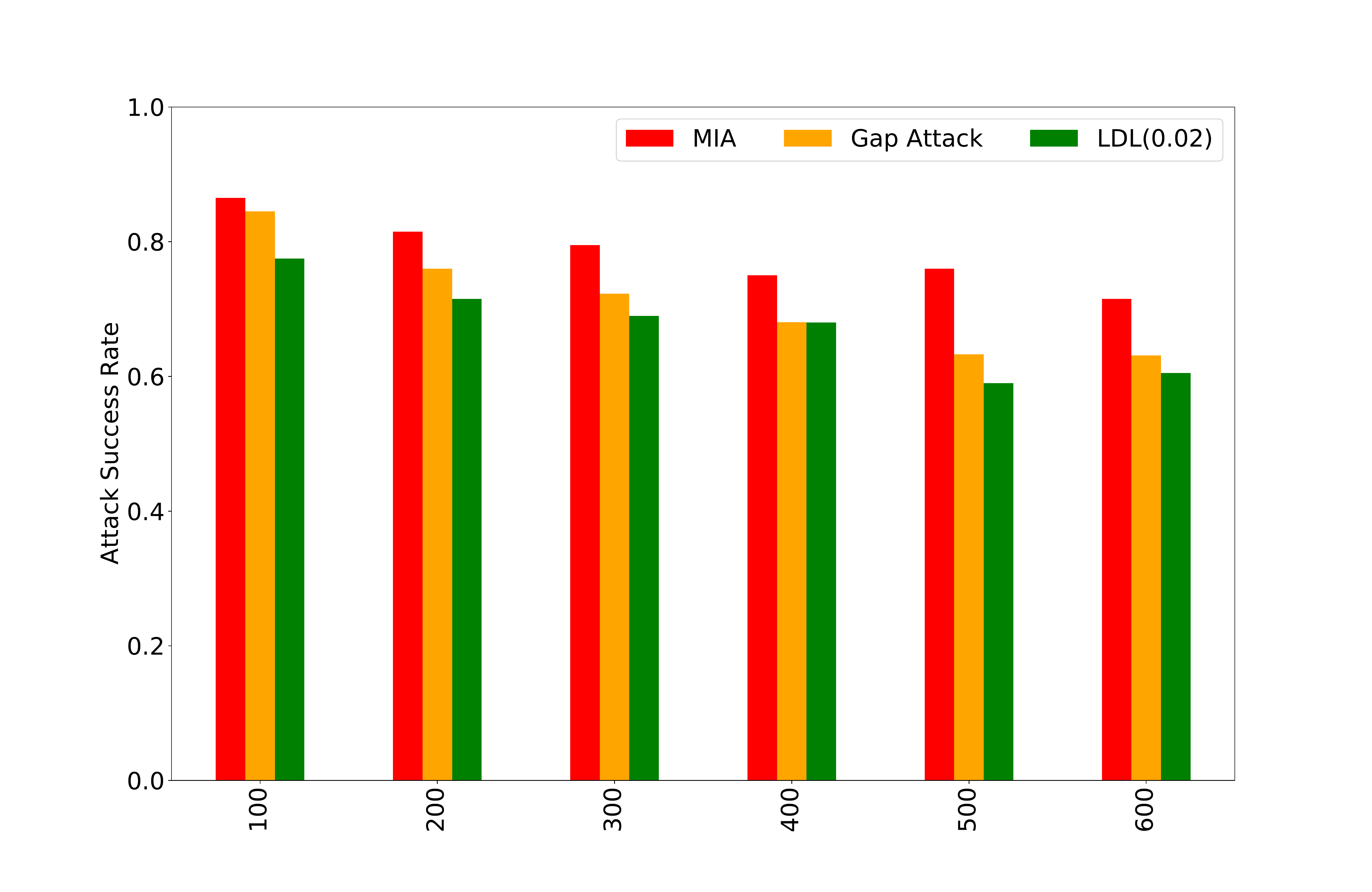}& 
        \includegraphics[trim={2cm 0.5cm 2cm 2cm},scale=0.15]{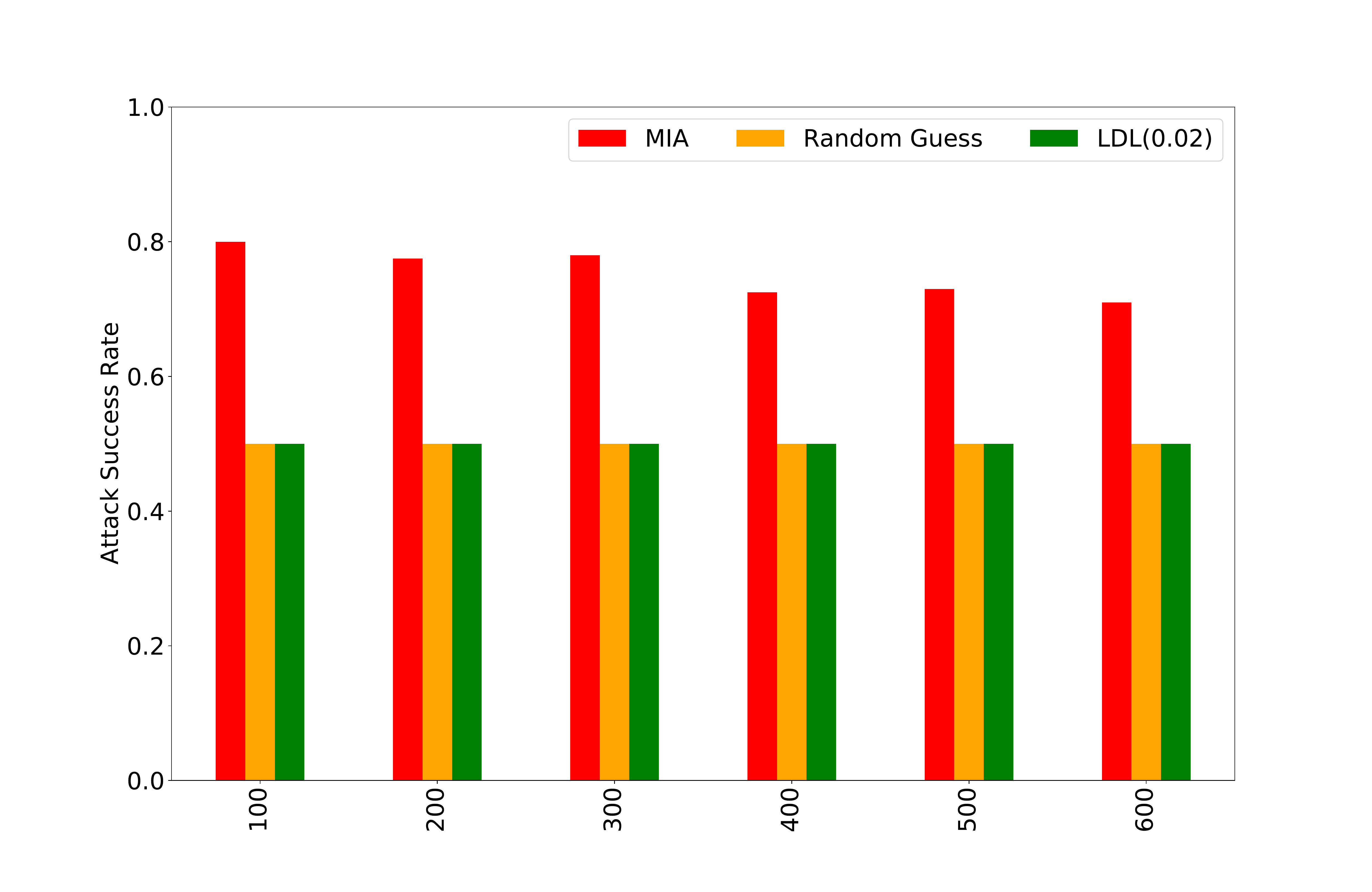}\\
         \rotatebox{90}{Face}&
         \includegraphics[trim={2cm 0.5cm 2cm 2cmm},scale=0.15]{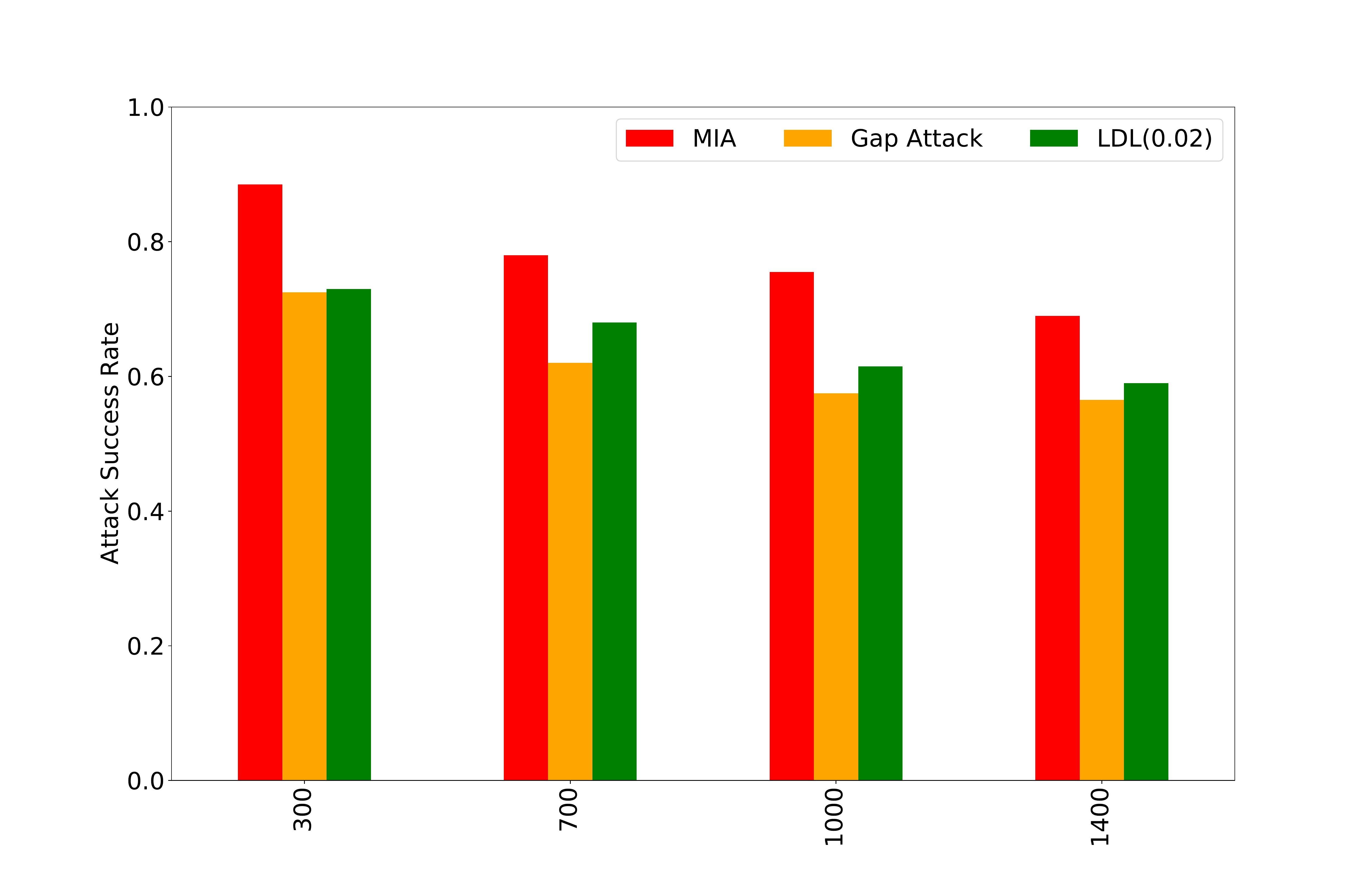}& 
         \includegraphics[trim={2cm 0.5cm 2cm 2cm},scale=0.15]{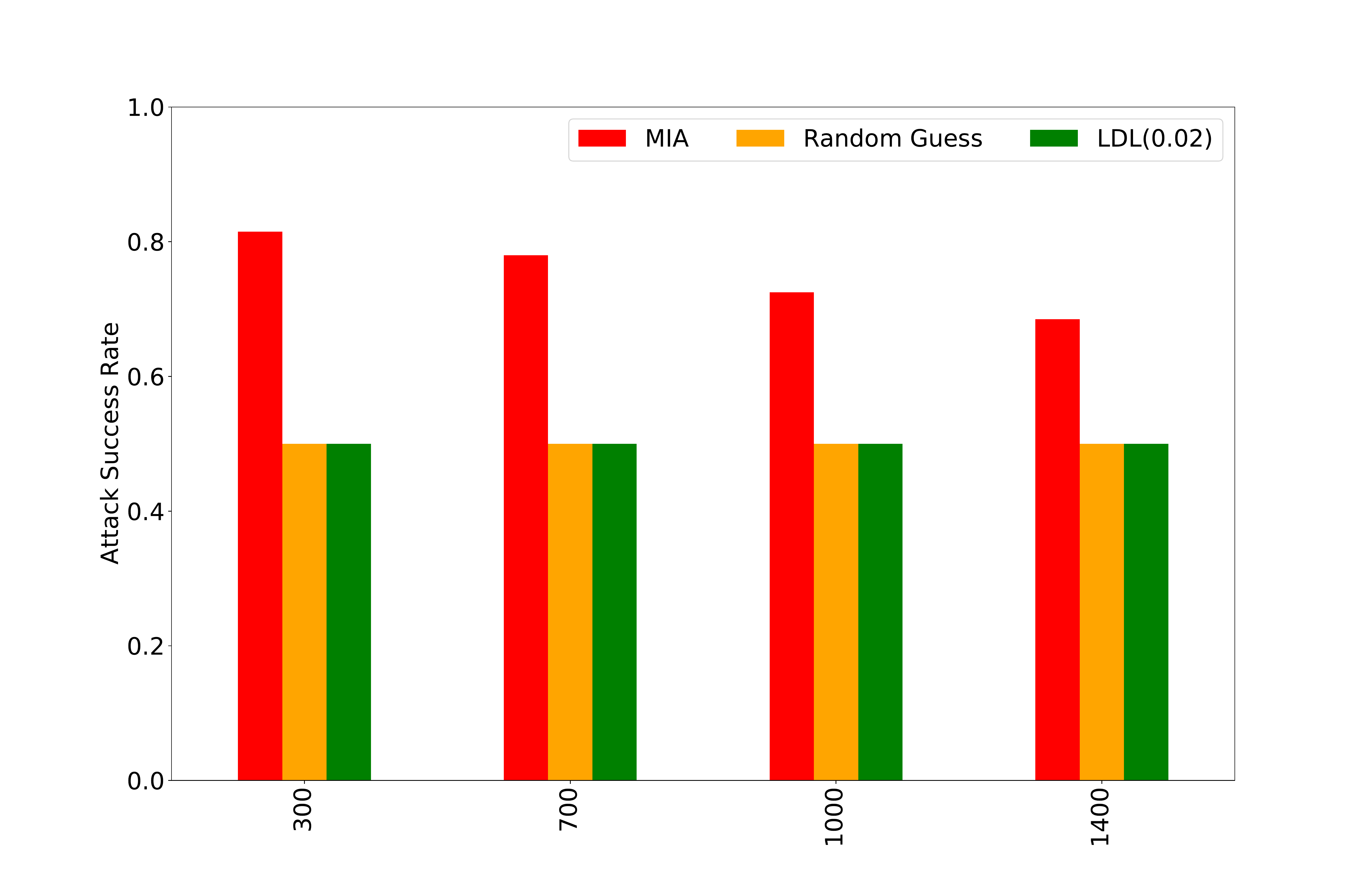}\\ 

    \end{tabular}
    
    \caption{Attack success rate (ASR) of adversary carrying out a LAB MIA using HopSkipJump adversarial noise~\cite{ccs2021} for defense-free models (red bars) and models with LDL deployed (green bars). We evaluate LDL on the GTSRB and Face datasets with different sizes of training sets for a strong adversary (left column) and a weak adversary (right column). The value of $ASR_{gap}$ for a strong adversary carrying out a gap attack and that of a random guess for a weak adversary on a defense-free model are also shown (orange bars). 
    We observe that deploying LDL successfully reduces the $ASR$ to a value close to $ASR_{gap}$ of the defense-free model. 
    }
    \label{fig:strongadv1}
\end{figure}

Figure~\ref{fig:strongadv1} compares the values of $ASR$ in the defense-free scenario (red bars) with the case when LDL is deployed as a defense (green bars) for the GTSRB and Face datasets (also see Fig.~\ref{fig:strongadv} in the main paper). 
We observe that using LDL is effective in reducing the value of $ASR$ to $ASR_{gap}$ for a strong adversary and to $50\%$ for a weak adversary, indicated by the orange bars. 

\begin{figure*}
    \centering
    \begin{tabular}{>{\centering\arraybackslash} m{0.14cm} >{\centering\arraybackslash} m{3.8cm} >{\centering\arraybackslash} m{3.8cm}>{\centering\arraybackslash} m{3.8cm}>{\centering\arraybackslash} m{3.8cm} }
    \rotatebox{90}{Nonmember}&
    \includegraphics[trim={4cm 1cm 8cm 4cm}, scale=0.15]{ 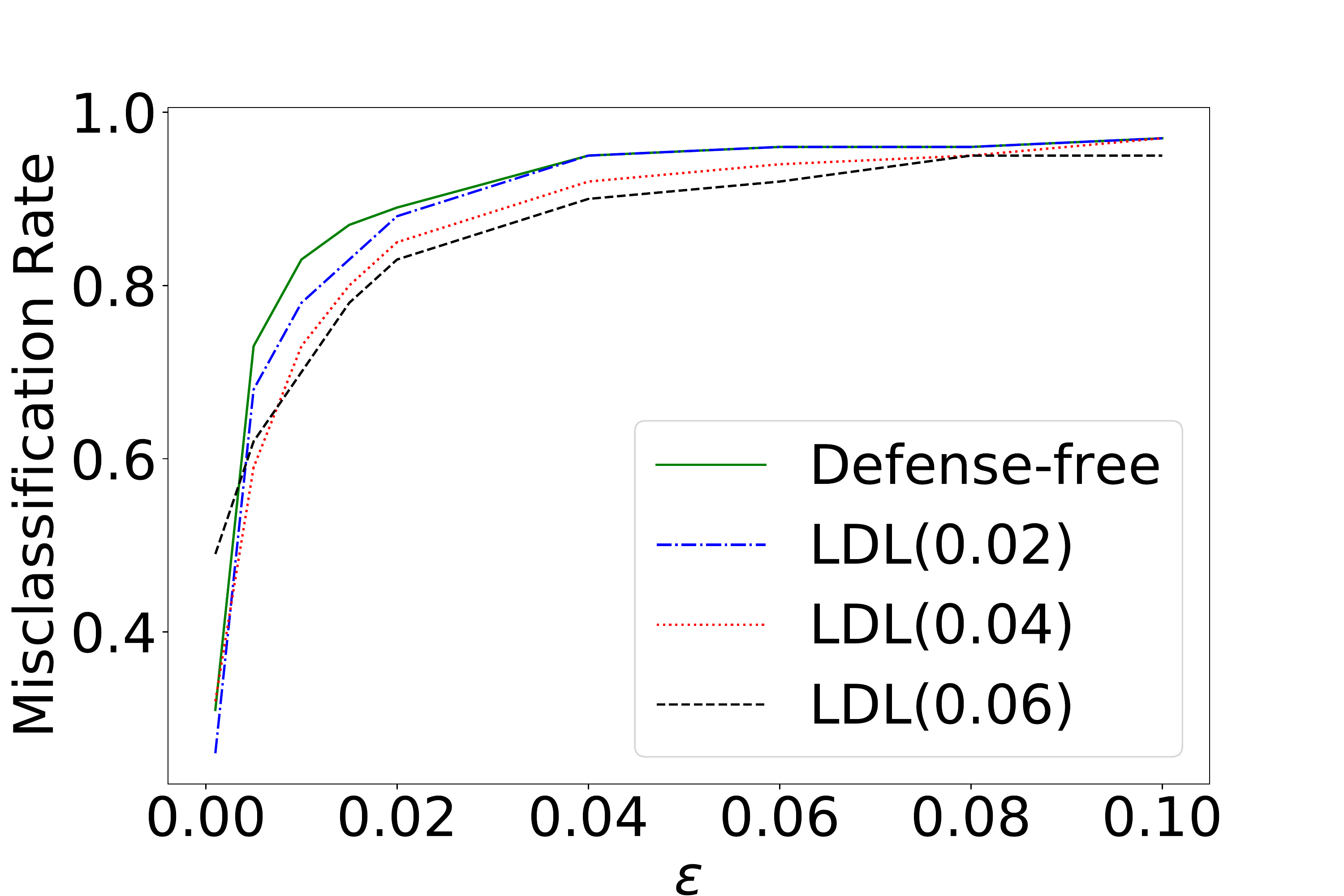}&
    \includegraphics[trim={4cm 1cm 8cm 4cm}, scale=0.15]{ 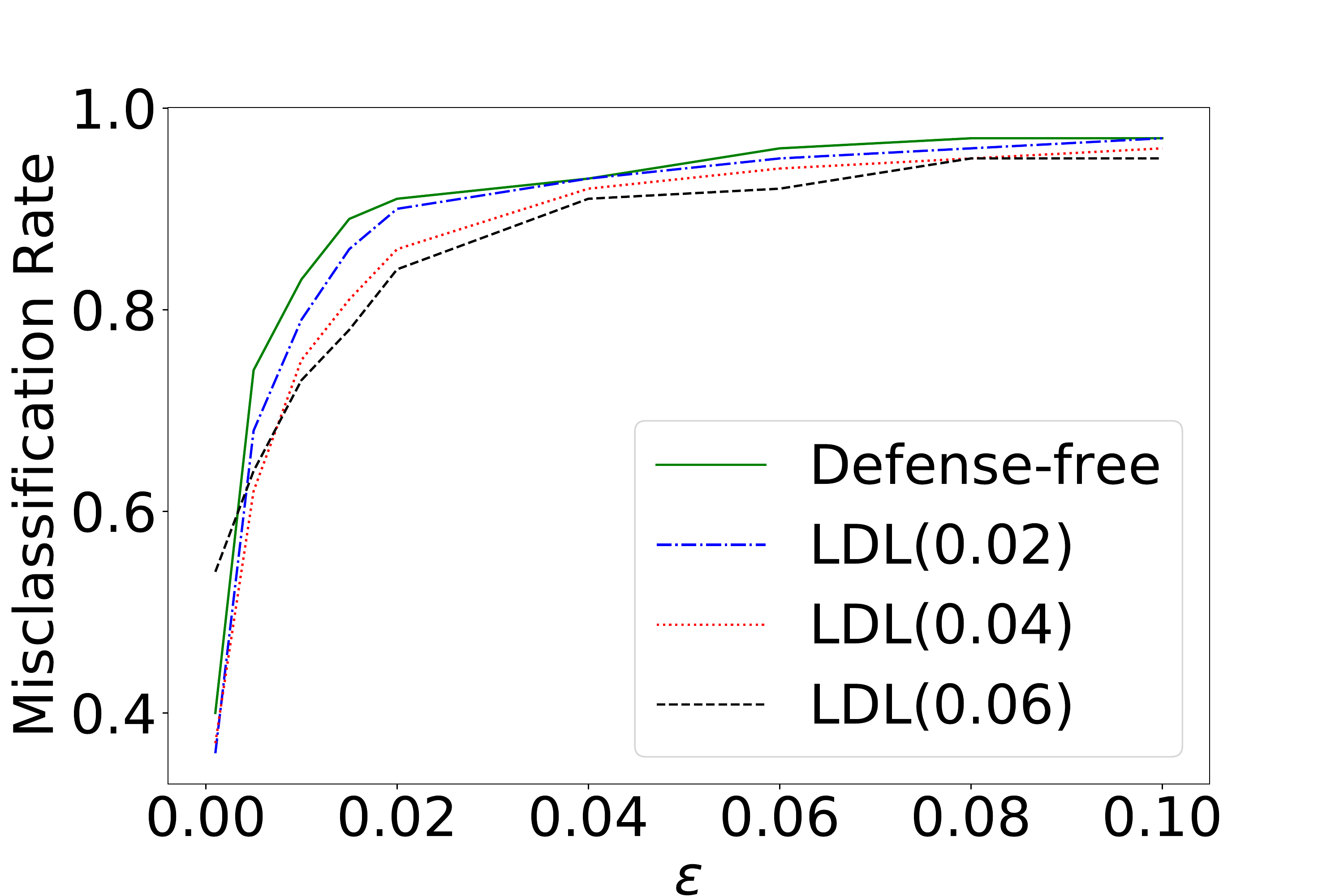}&
    \includegraphics[trim={4cm 1cm 8cm 4cm}, scale=0.15]{ 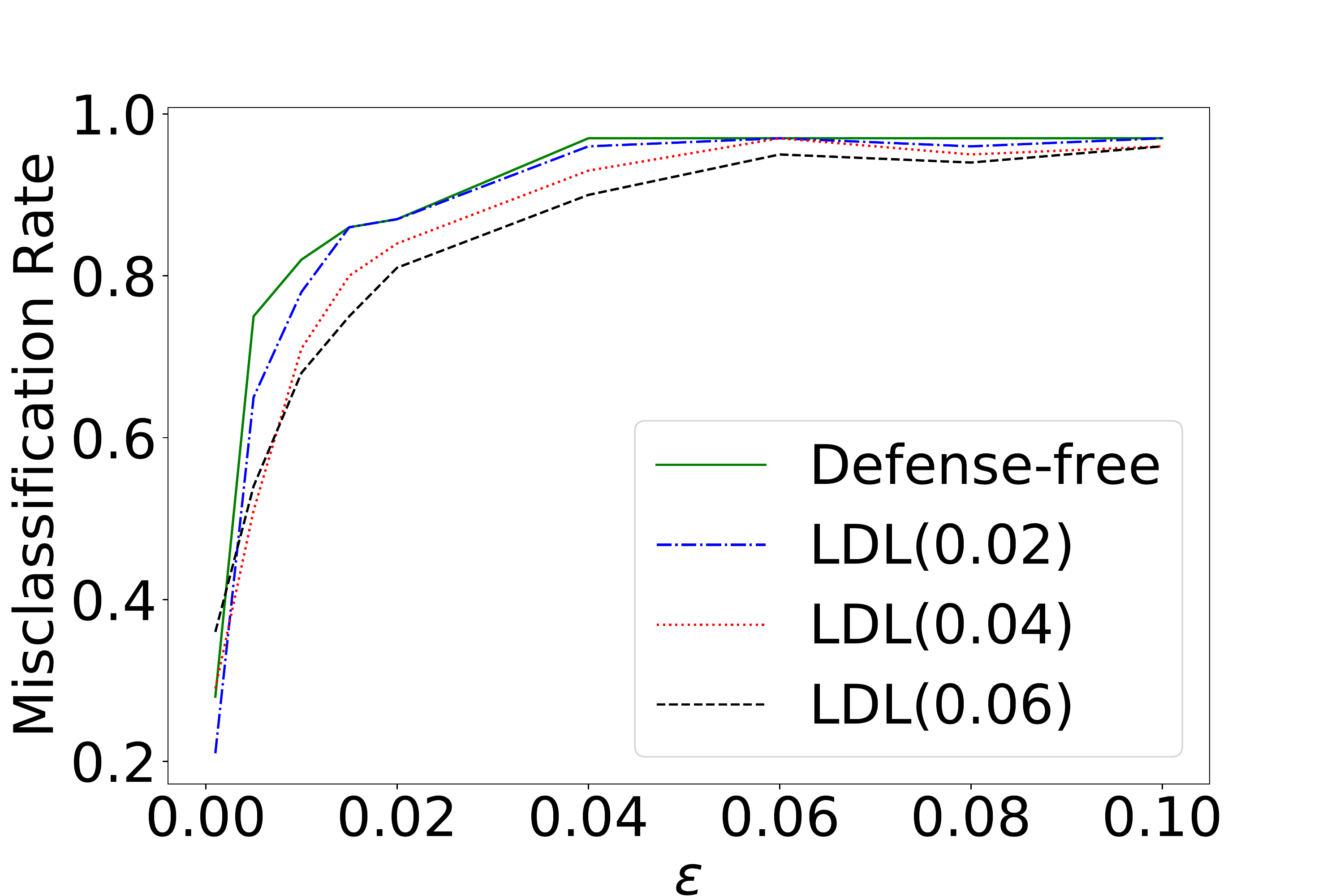}&
    \includegraphics[trim={4cm 1cm 8cm 4cm}, scale=0.15]{ 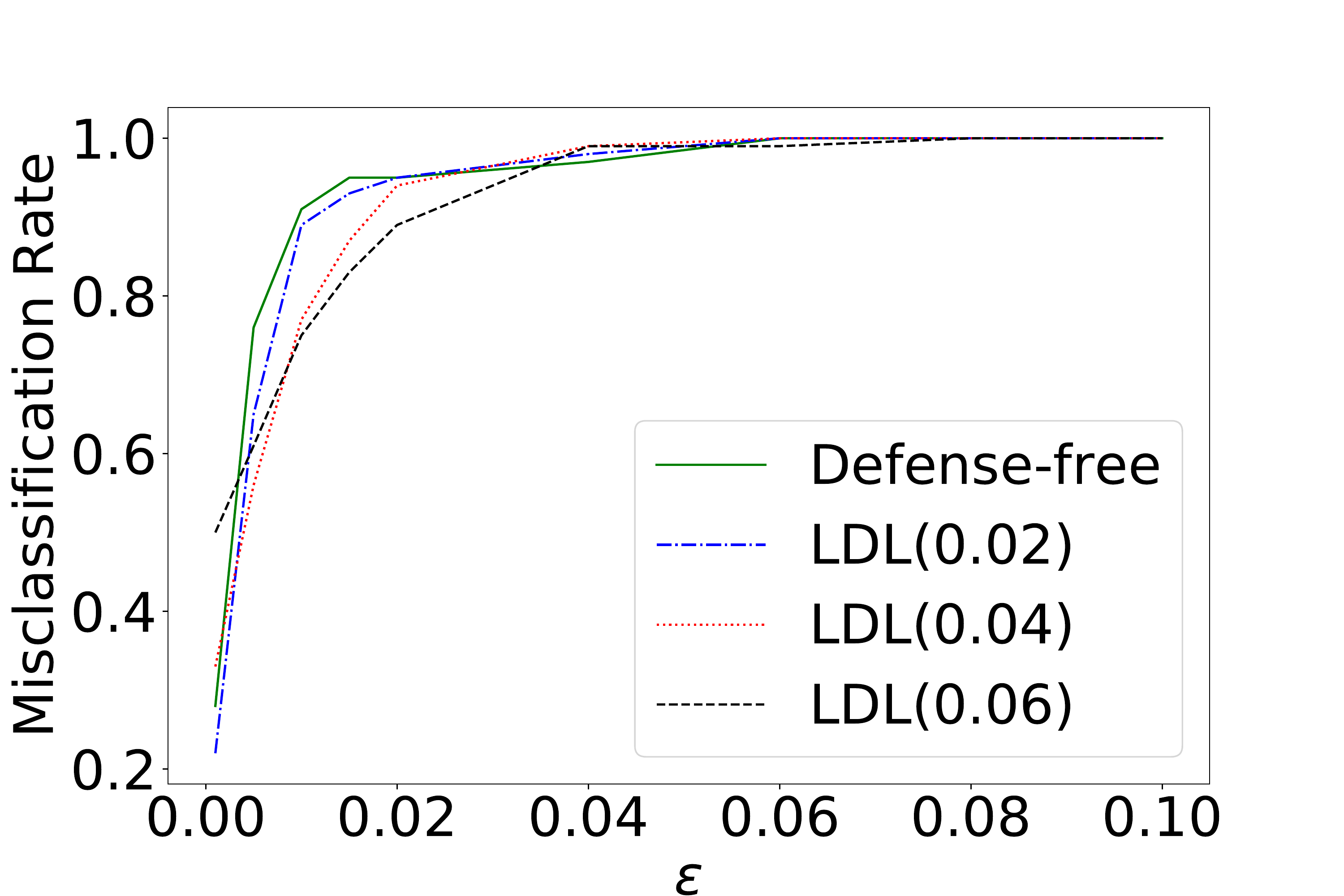}\\
    \\
    \rotatebox{90}{Member}&
    \includegraphics[trim={4cm 0.5cm 8cm 4cm}, scale=0.15]{ 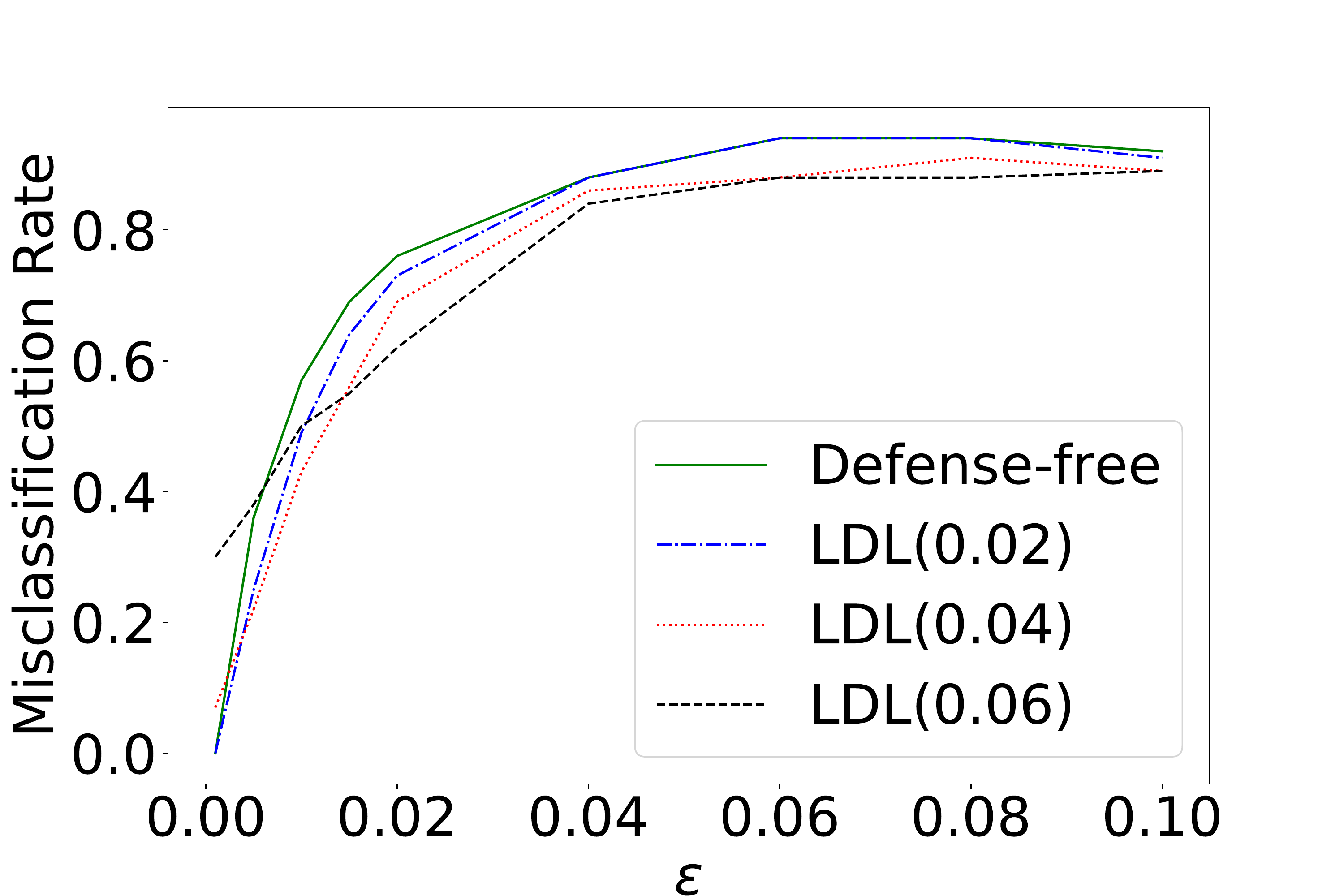}&
    \includegraphics[trim={4cm 0.5cm 8cm 4cm}, scale=0.15]{ 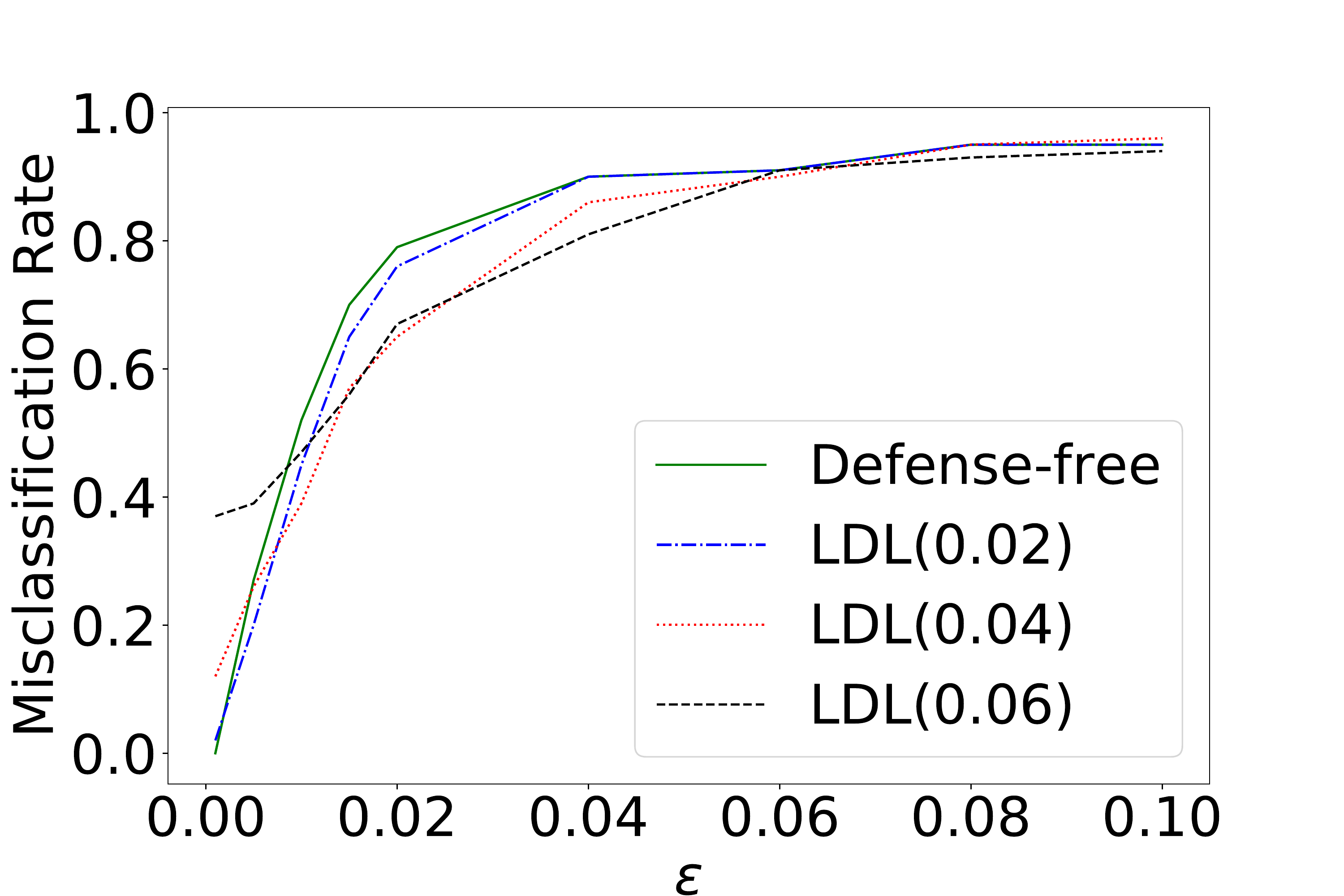}&
    \includegraphics[trim={4cm 0.5cm 8cm 4cm}, scale=0.15]{ 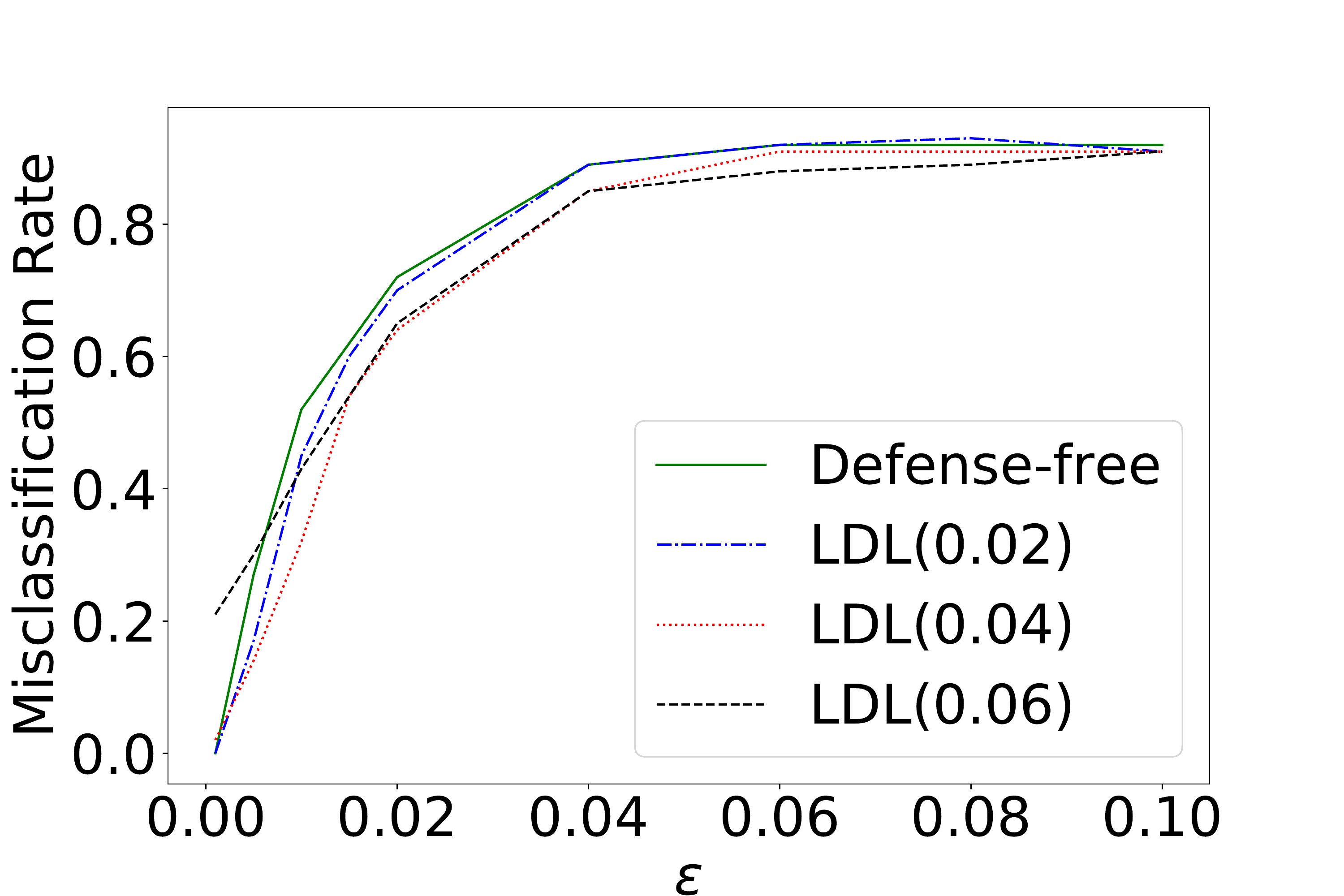}&
    \includegraphics[trim={4cm 0.5cm 8cm 4cm}, scale=0.15]{ 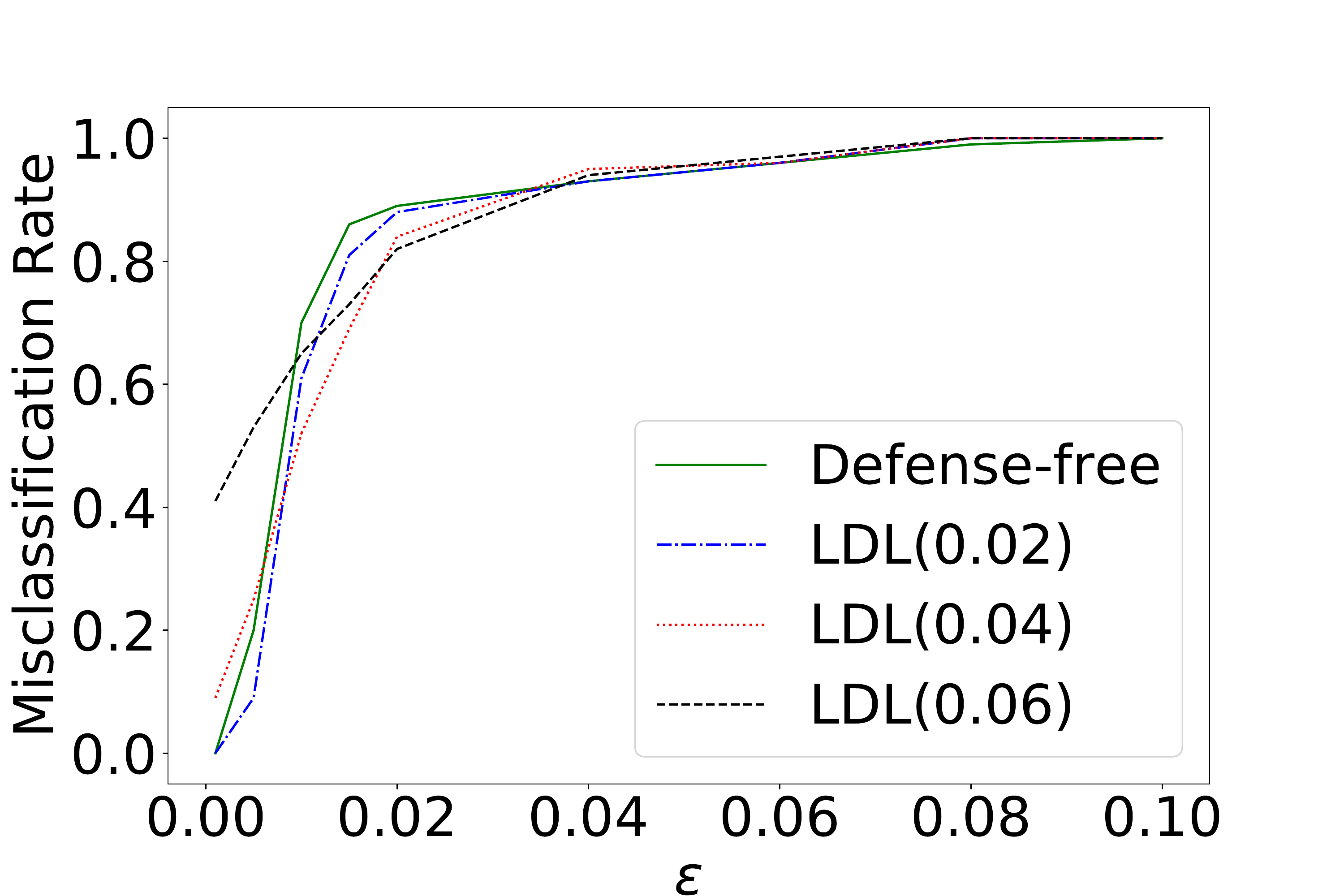}\\
    & CIFAR-100 (40000) & CIFAR-100 (35000) &CIFAR-100 (25000) &CIFAR-100 (15000) \\ \\ \hline 
    \\
    \rotatebox{90}{Nonmember}&
    \includegraphics[trim={4cm 1cm 8cm 4cm}, scale=0.15]{ 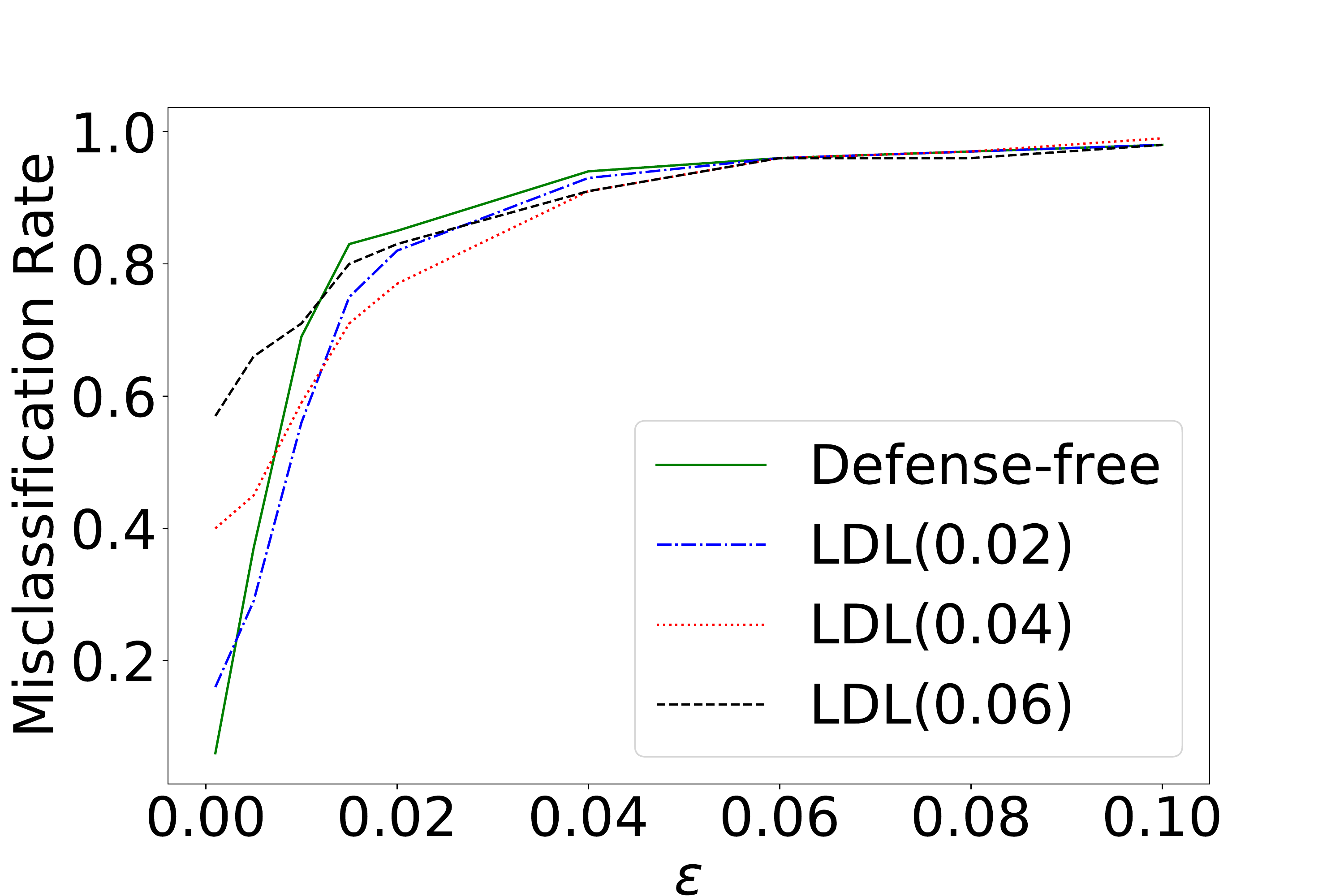}&
    \includegraphics[trim={4cm 1cm 8cm 4cm}, scale=0.15]{ 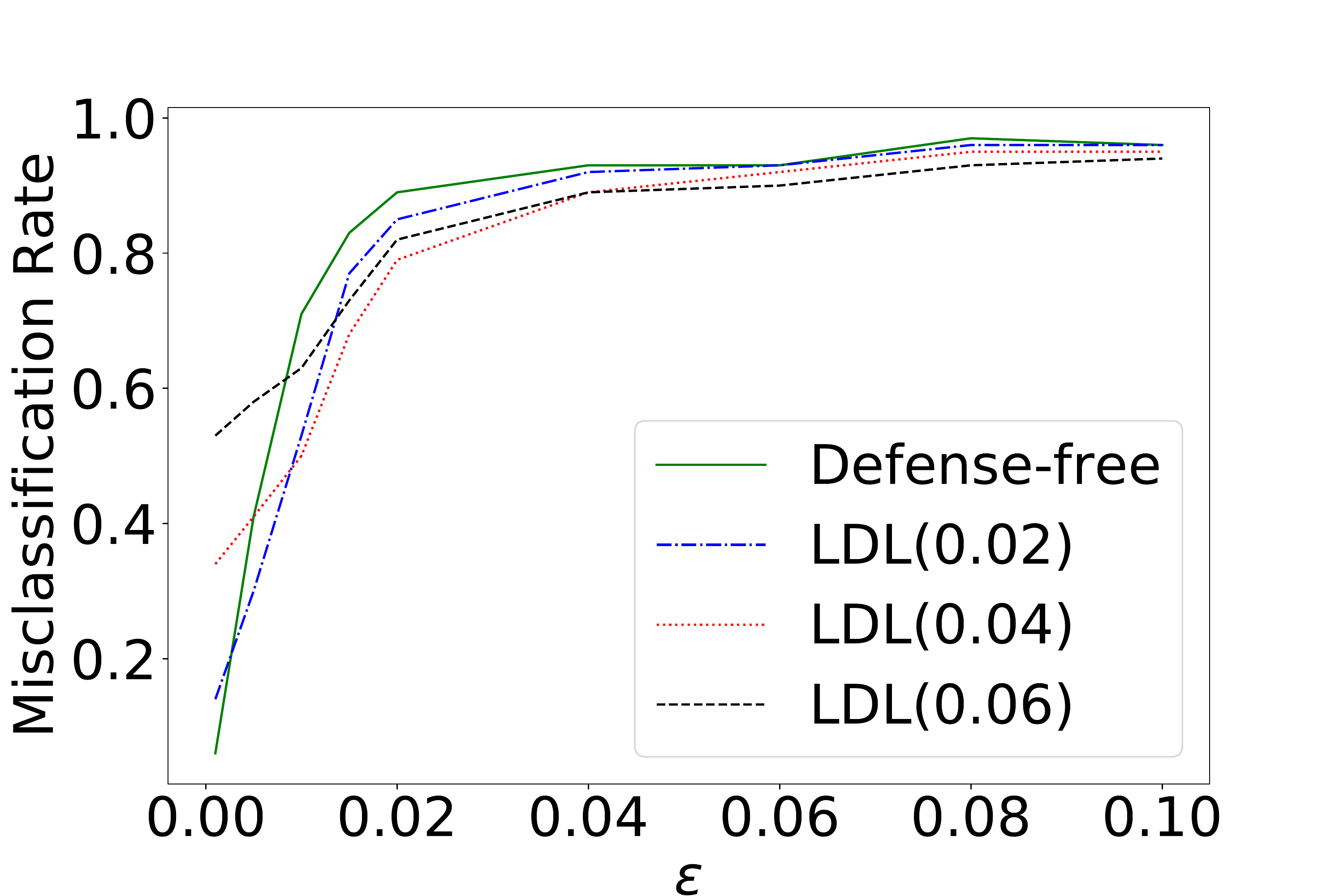}&
    \includegraphics[trim={4cm 1cm 8cm 4cm}, scale=0.15]{ 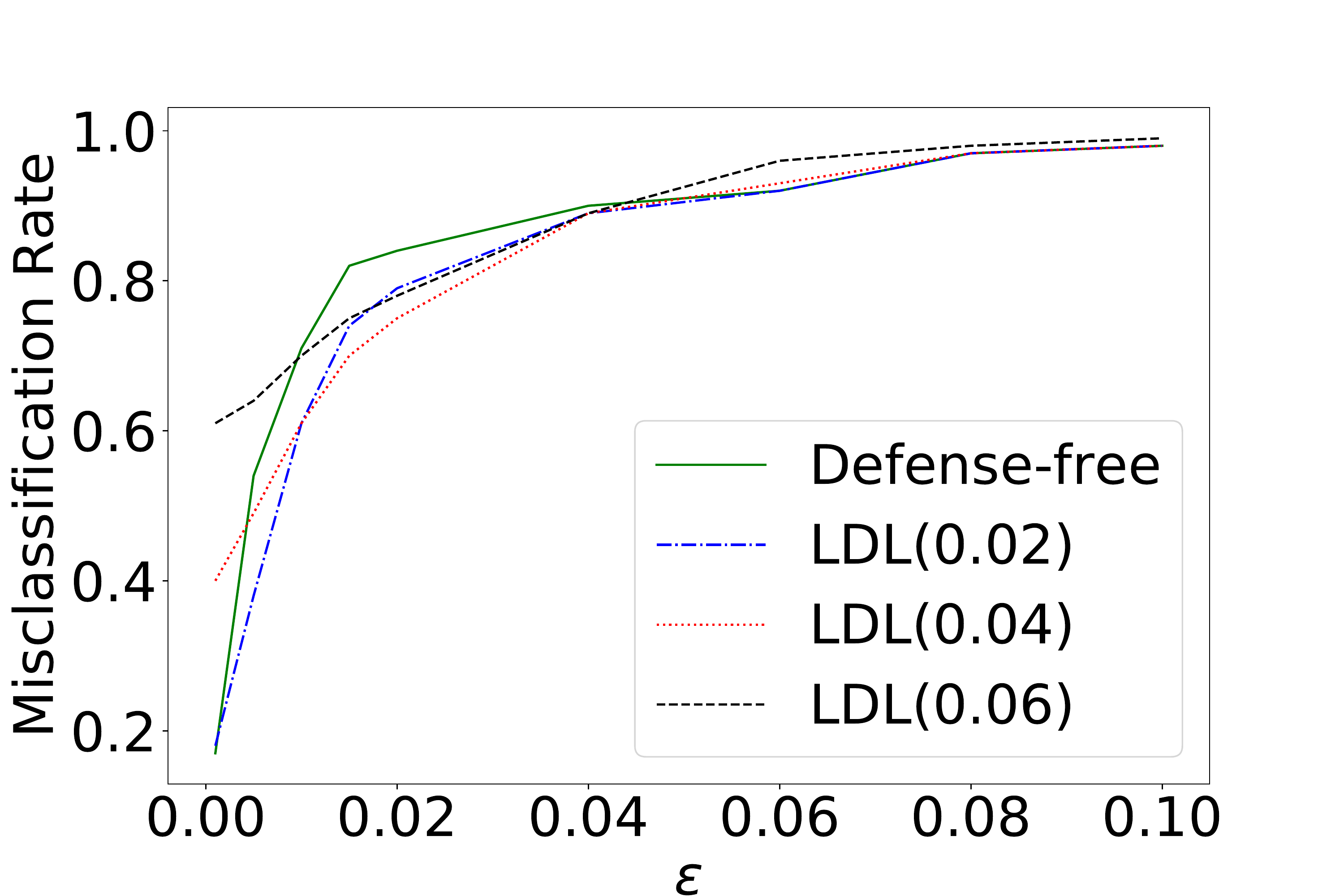}&
    \includegraphics[trim={4cm 1cm 8cm 4cm}, scale=0.15]{ 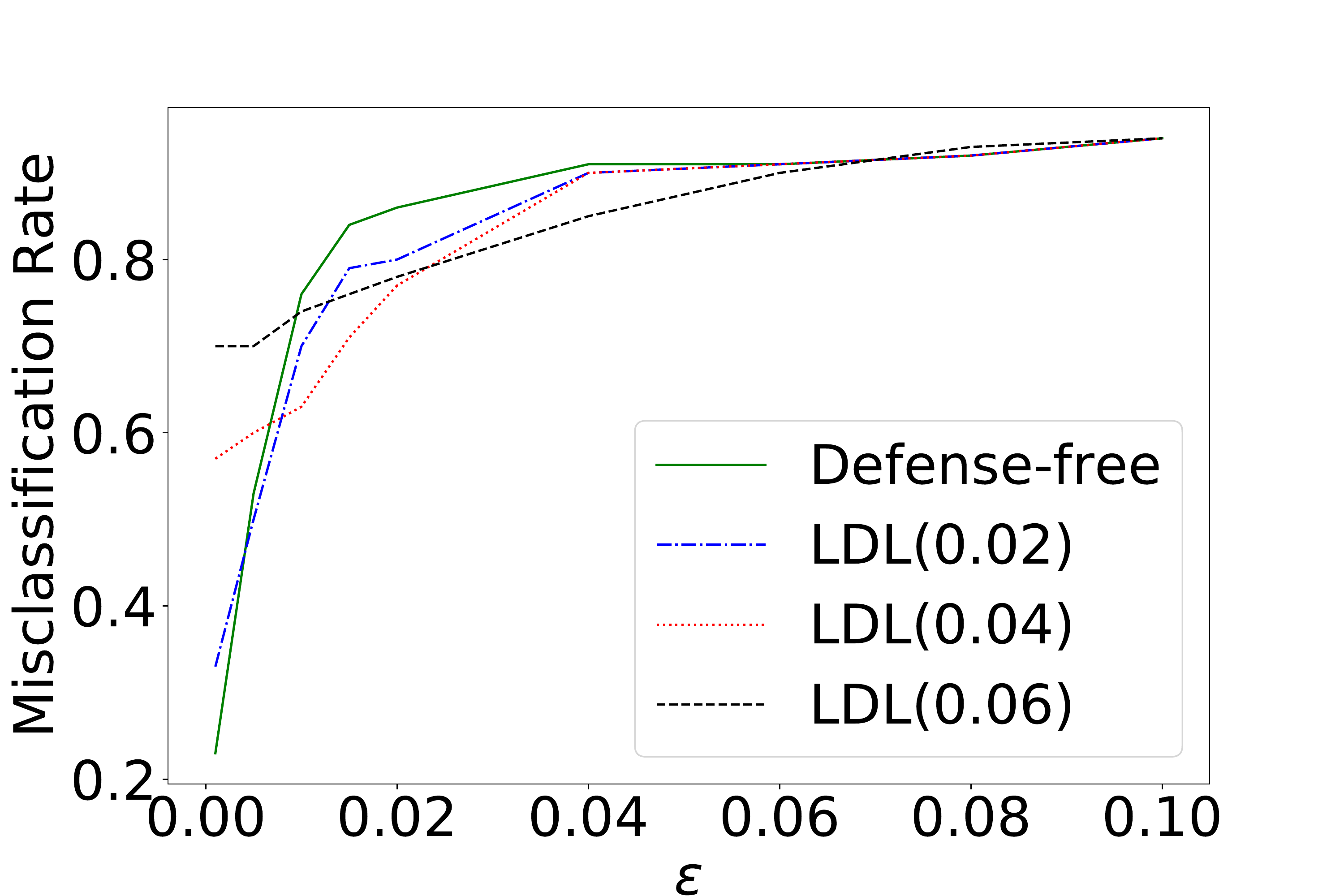}\\
    \\
    \rotatebox{90}{Member}&
    \includegraphics[trim={4cm 0.5cmcm 8cm 4cm}, scale=0.15]{ 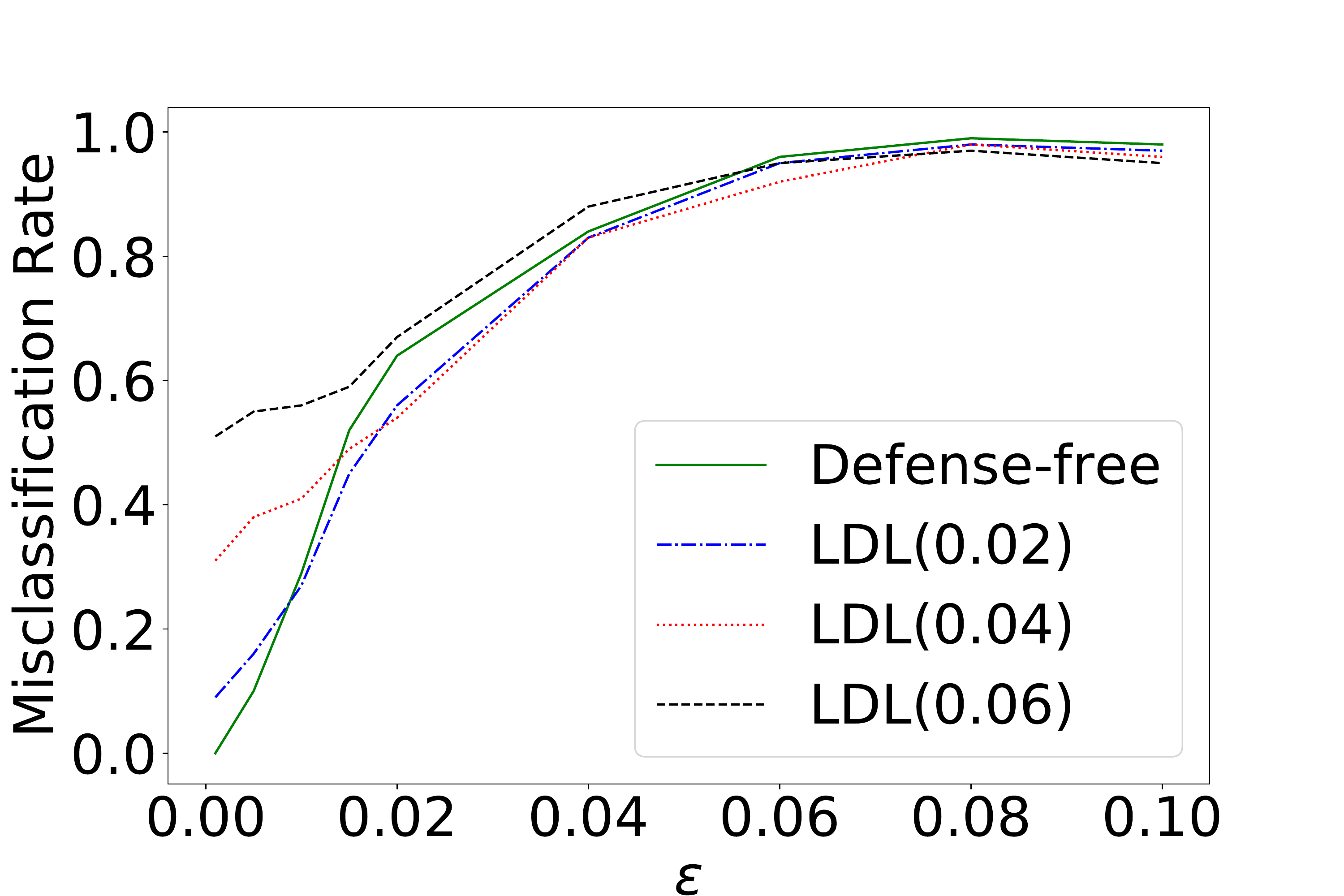}&
    \includegraphics[trim={4cm 0.5cmcm 8cm 4cm}, scale=0.15]{ 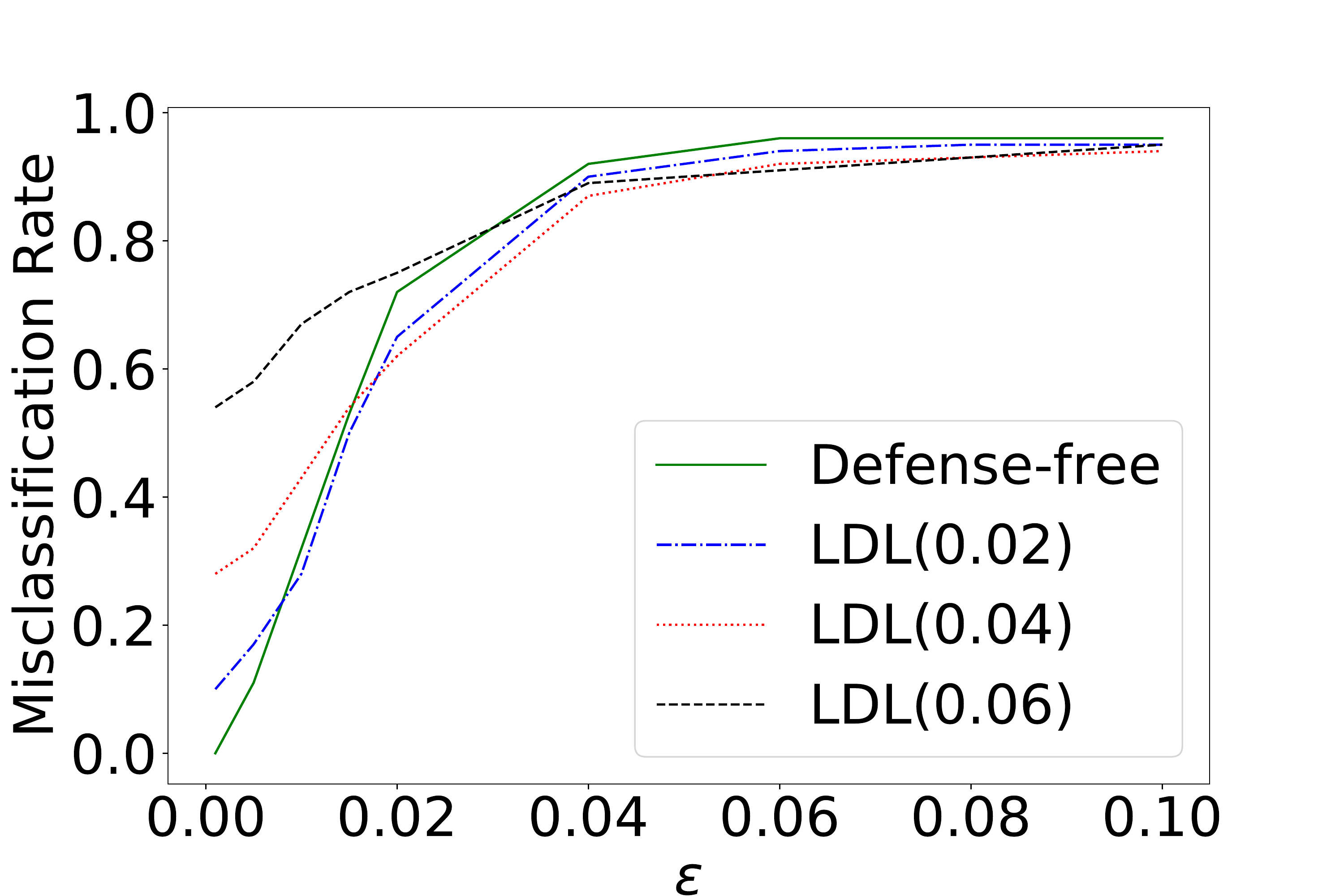}&
    \includegraphics[trim={4cm 0.5cmcm 8cm 4cm}, scale=0.15]{ 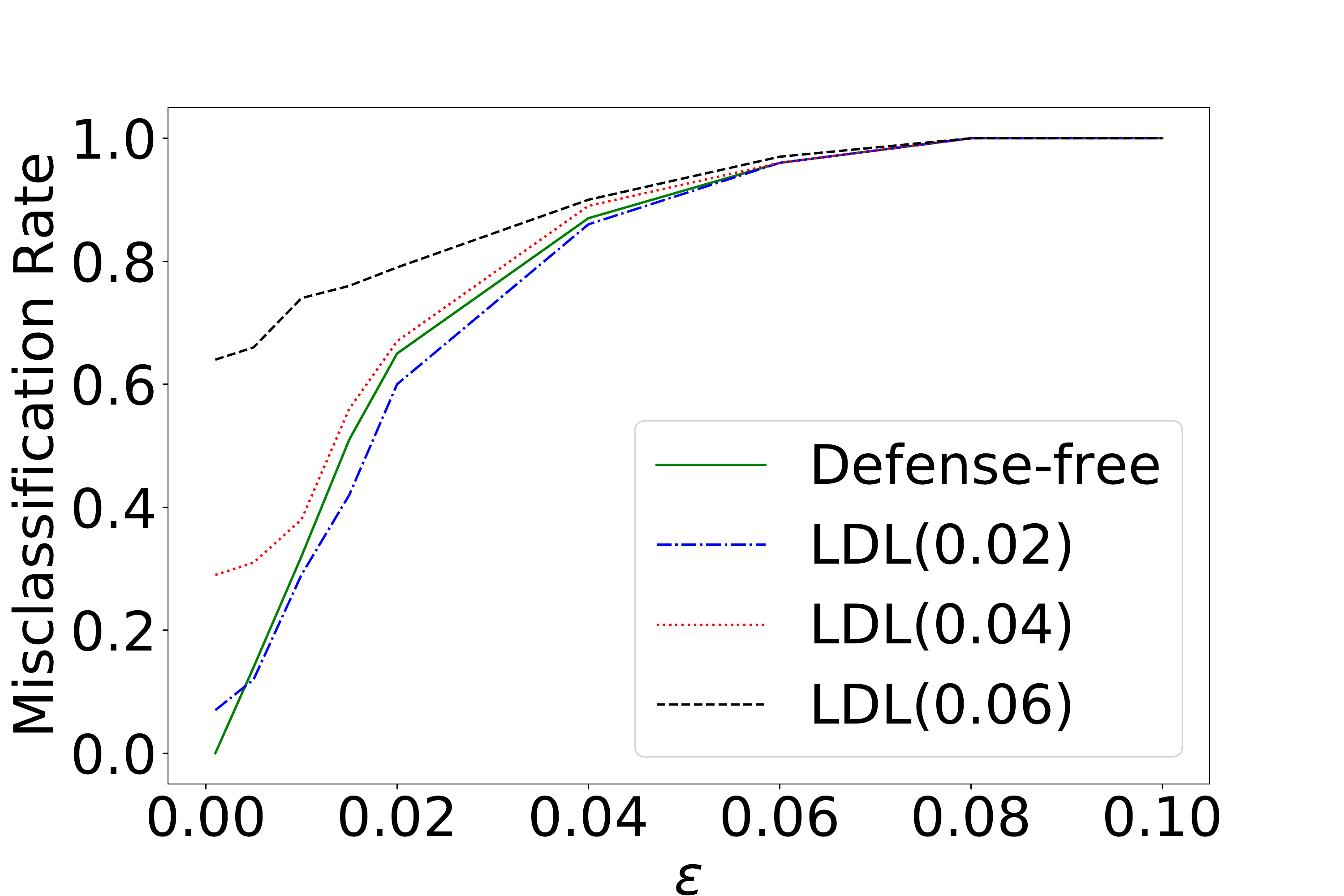}&
    \includegraphics[trim={4cm 0.5cmcm 8cm 4cm}, scale=0.15]{ 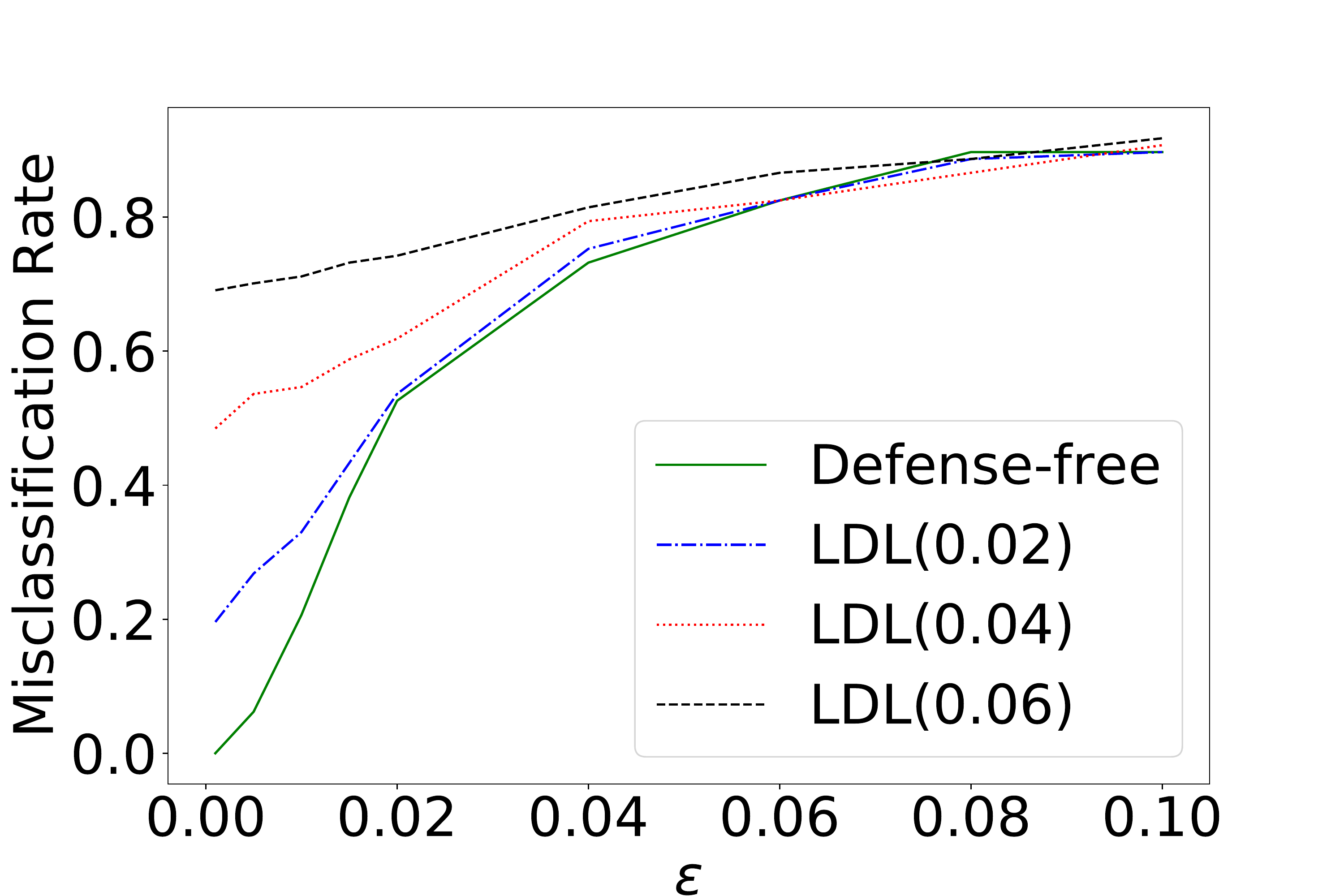}\\
    & GTSRB (600) & GTSRB (500) &GTSRB (300) &GTSRB (100) \\
    \\ \hline 
    \\
    \rotatebox{90}{Nonmember}&
    \includegraphics[trim={4cm 1cm 8cm 4cm}, scale=0.15]{ 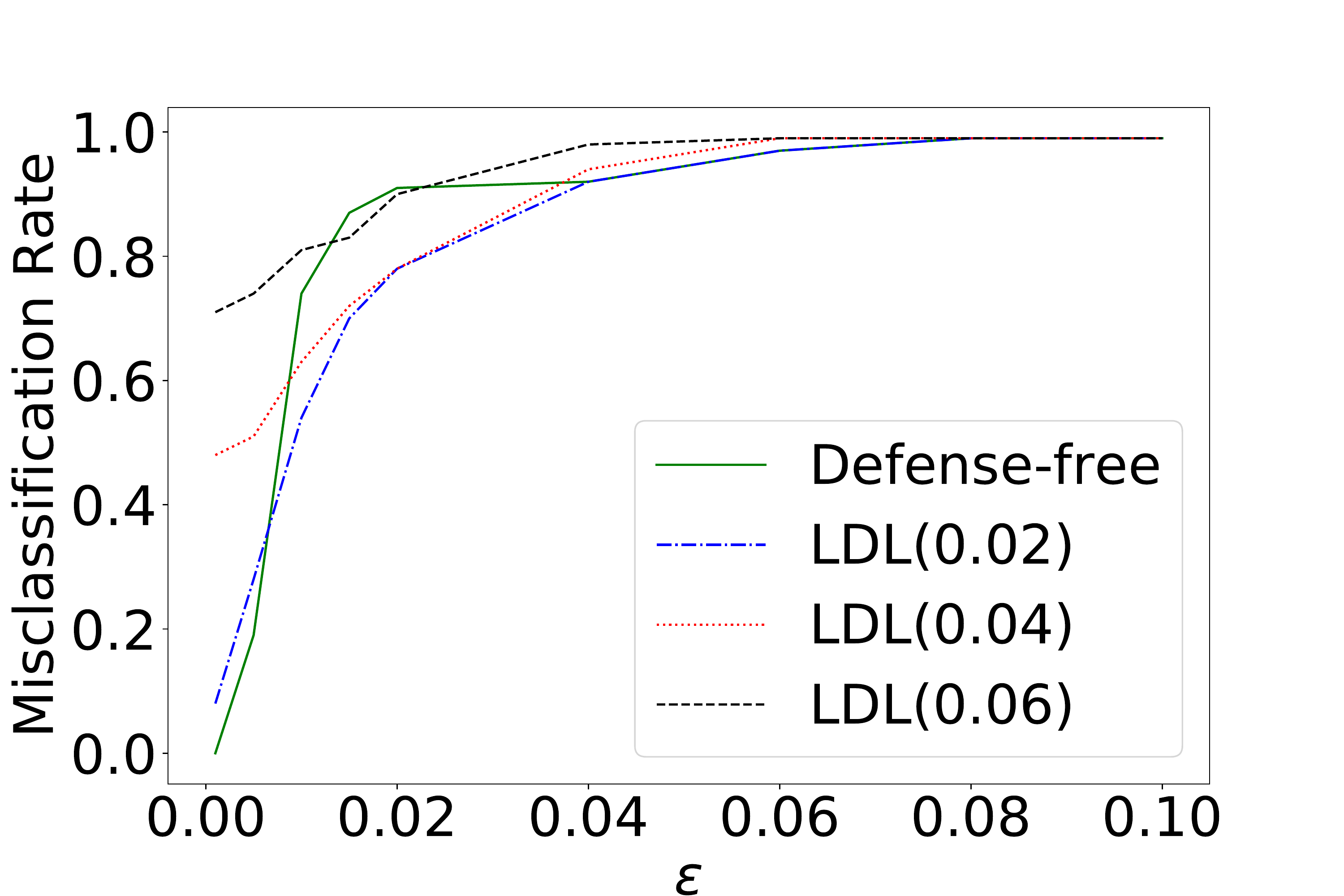}&
    \includegraphics[trim={4cm 1cm 8cm 4cm}, scale=0.15]{ 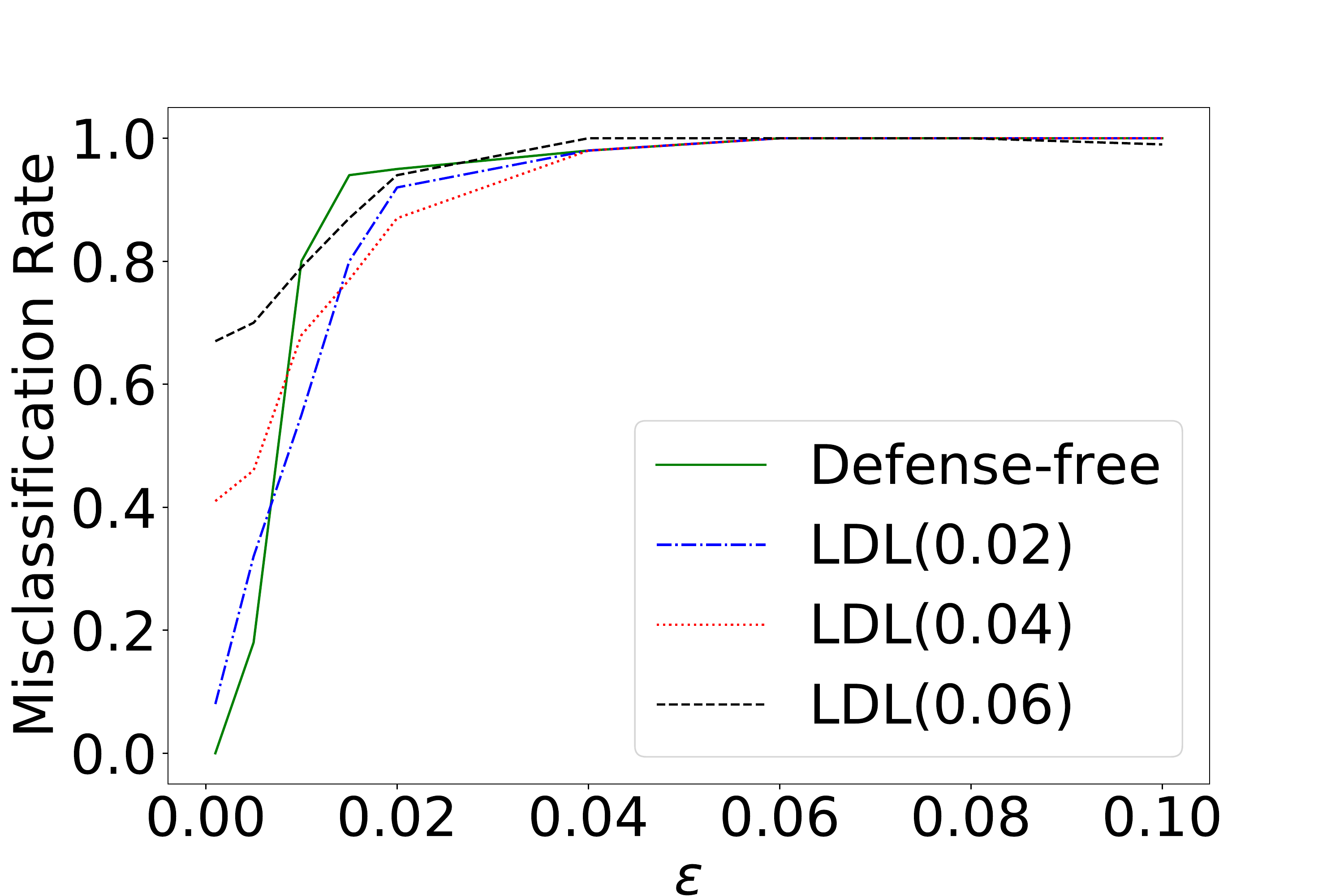}&
    \includegraphics[trim={4cm 1cm 8cm 4cm}, scale=0.15]{ 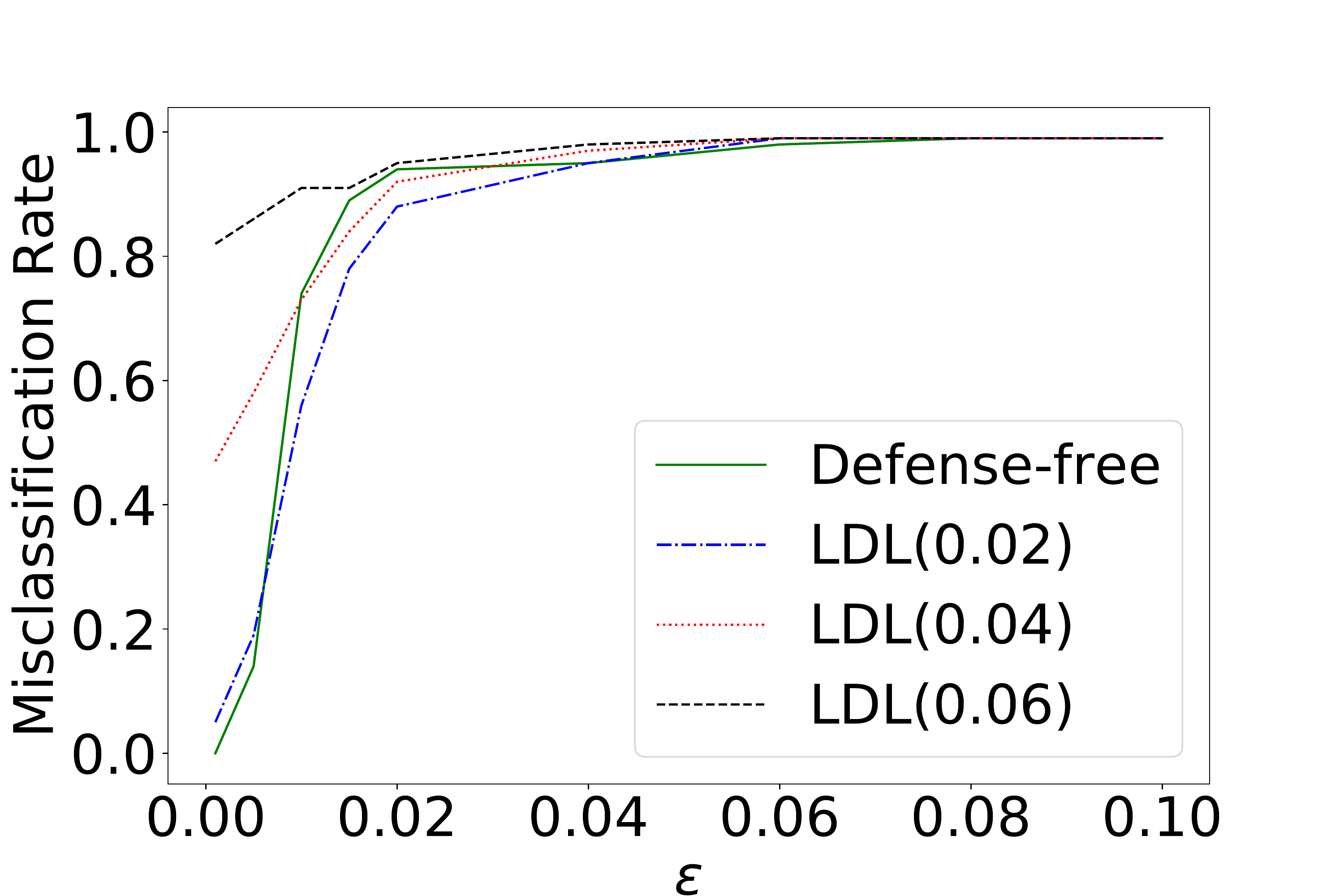}&
    \includegraphics[trim={4cm 1cm 8cm 4cm}, scale=0.15]{ 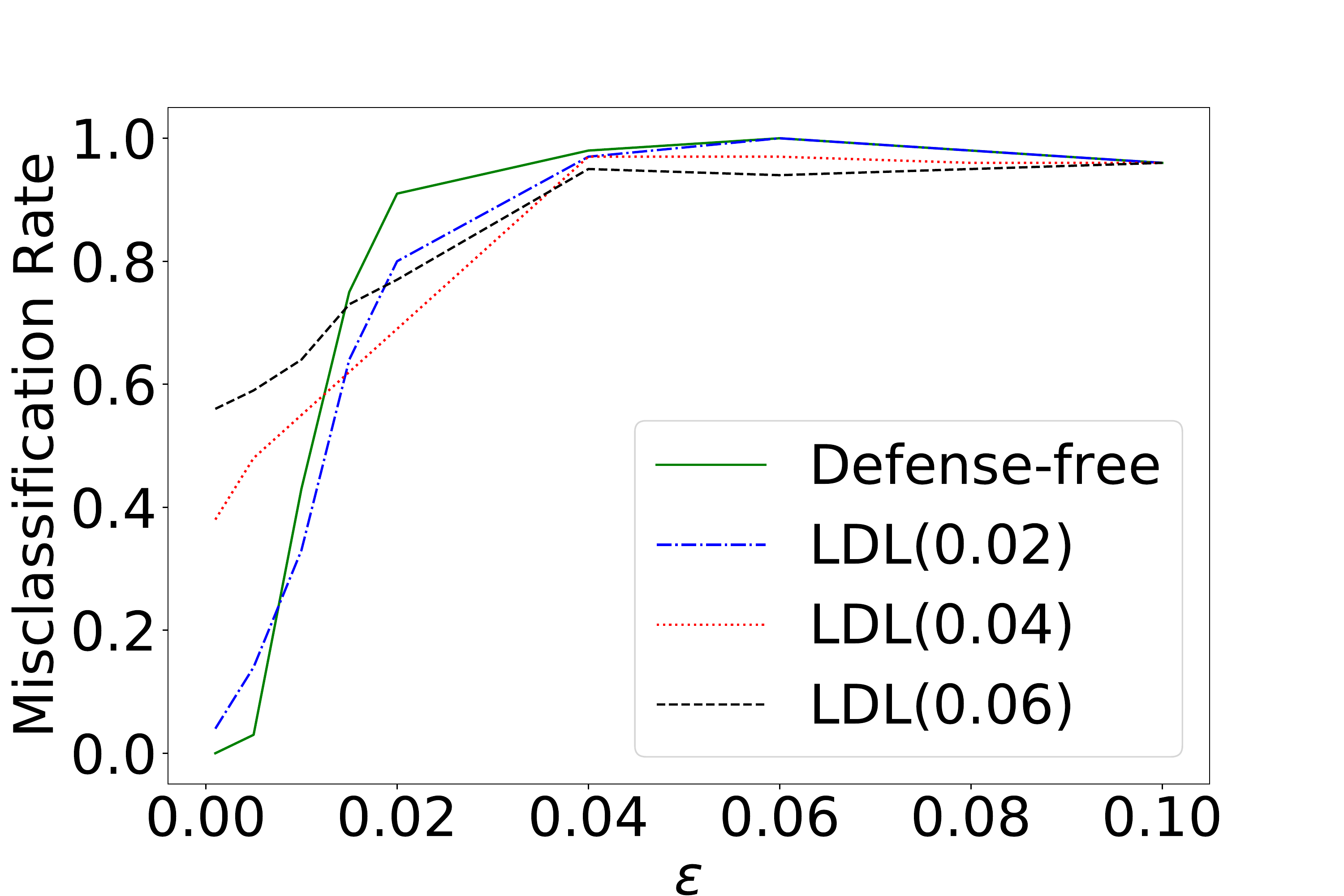}\\
    \\
    \rotatebox{90}{Member}&
    \includegraphics[trim={4cm 0.5cm 8cm 4cm}, scale=0.15]{ 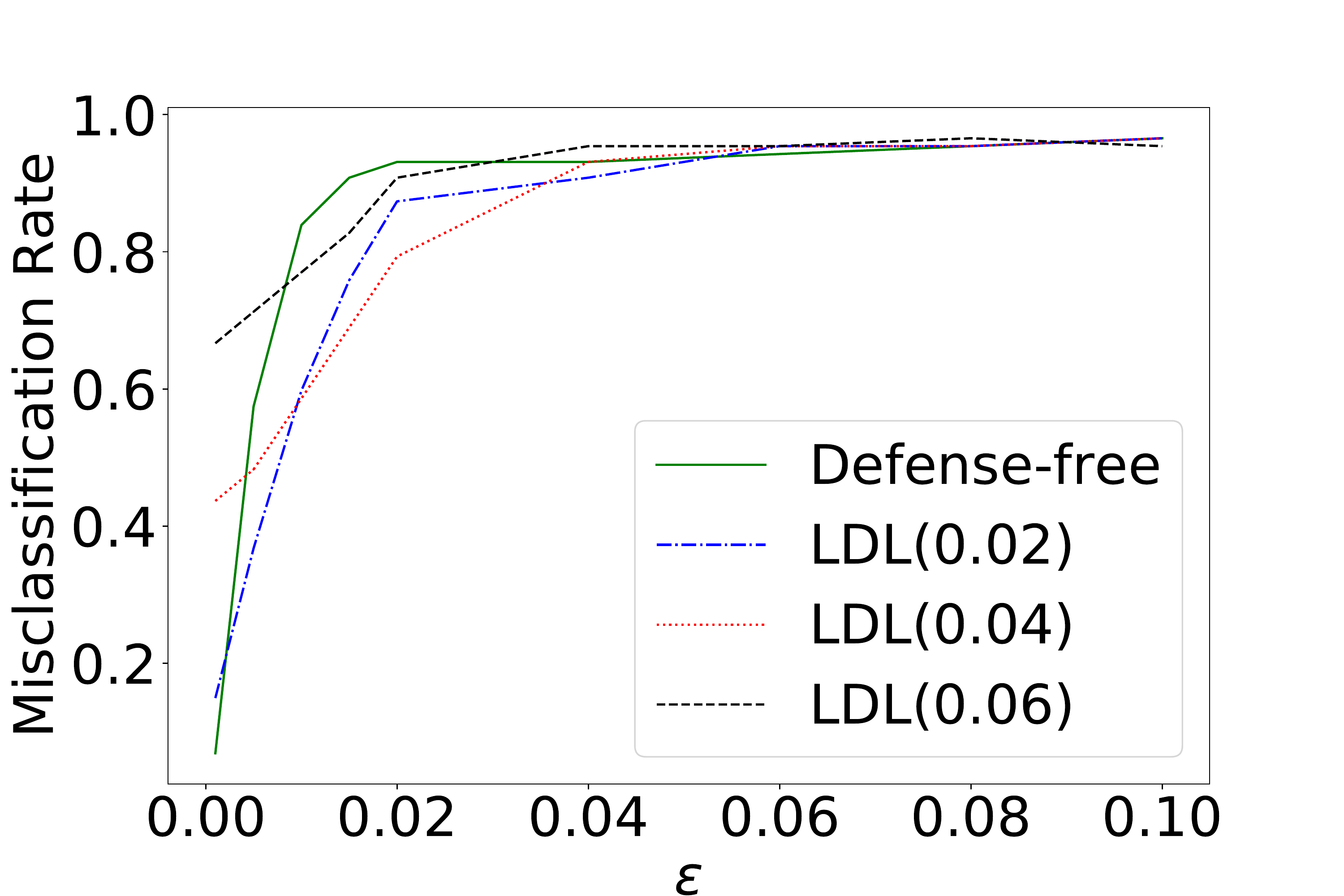}&
    \includegraphics[trim={4cm 0.5cm 8cm 4cm}, scale=0.15]{ 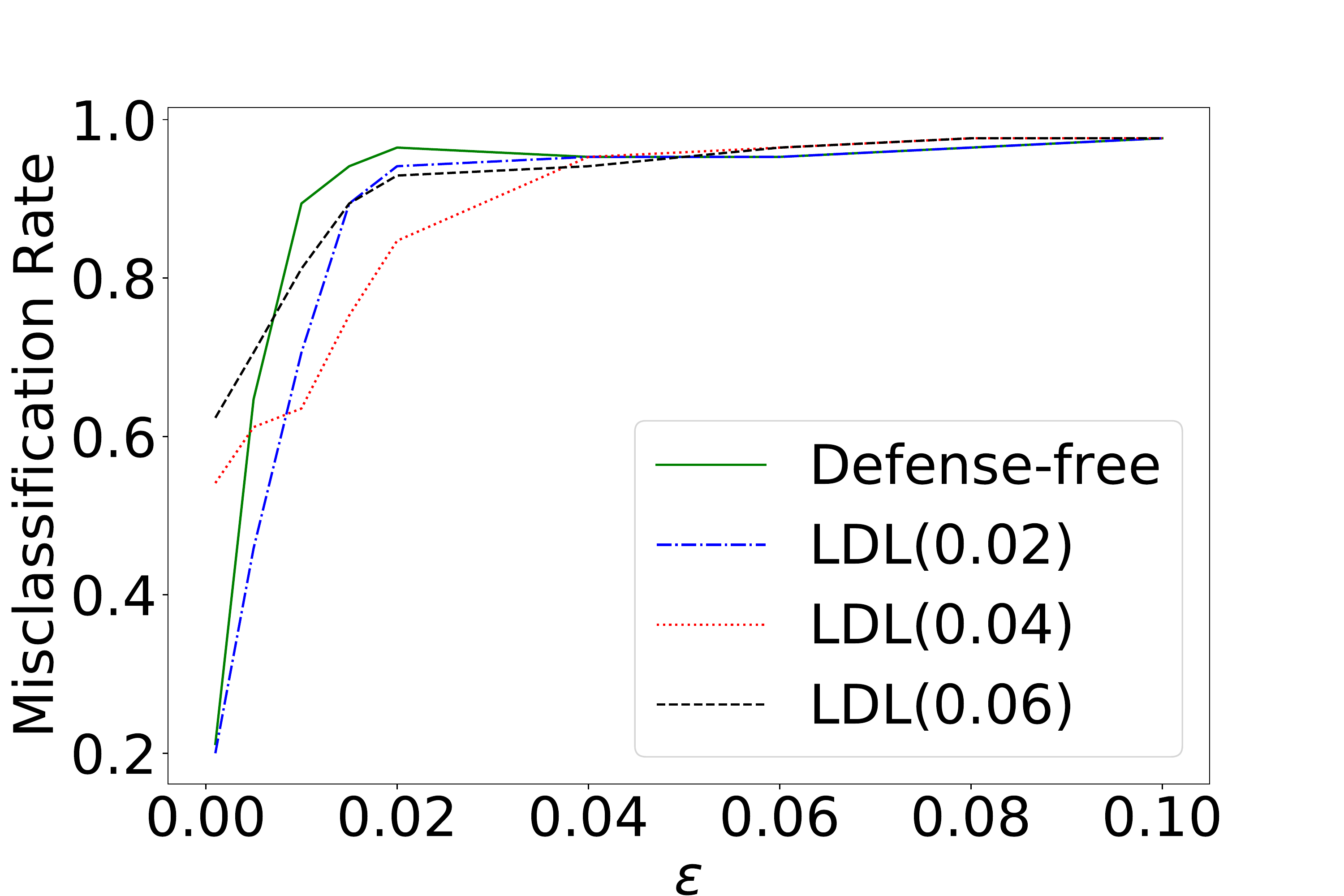}&
    \includegraphics[trim={4cm 0.5cm 8cm 4cm}, scale=0.15]{ 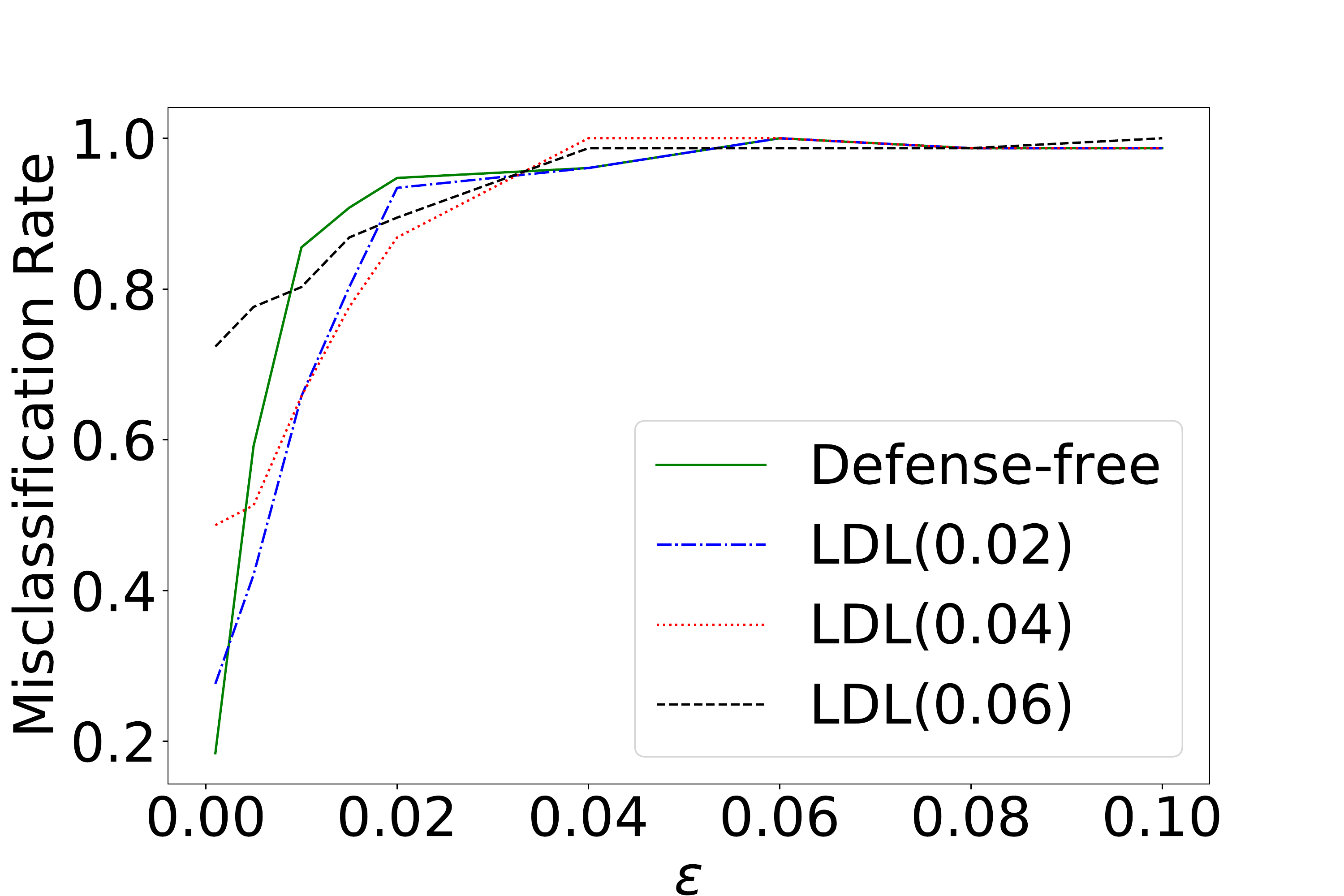}&
    \includegraphics[trim={4cm 0.5cm 8cm 4cm}, scale=0.15]{ 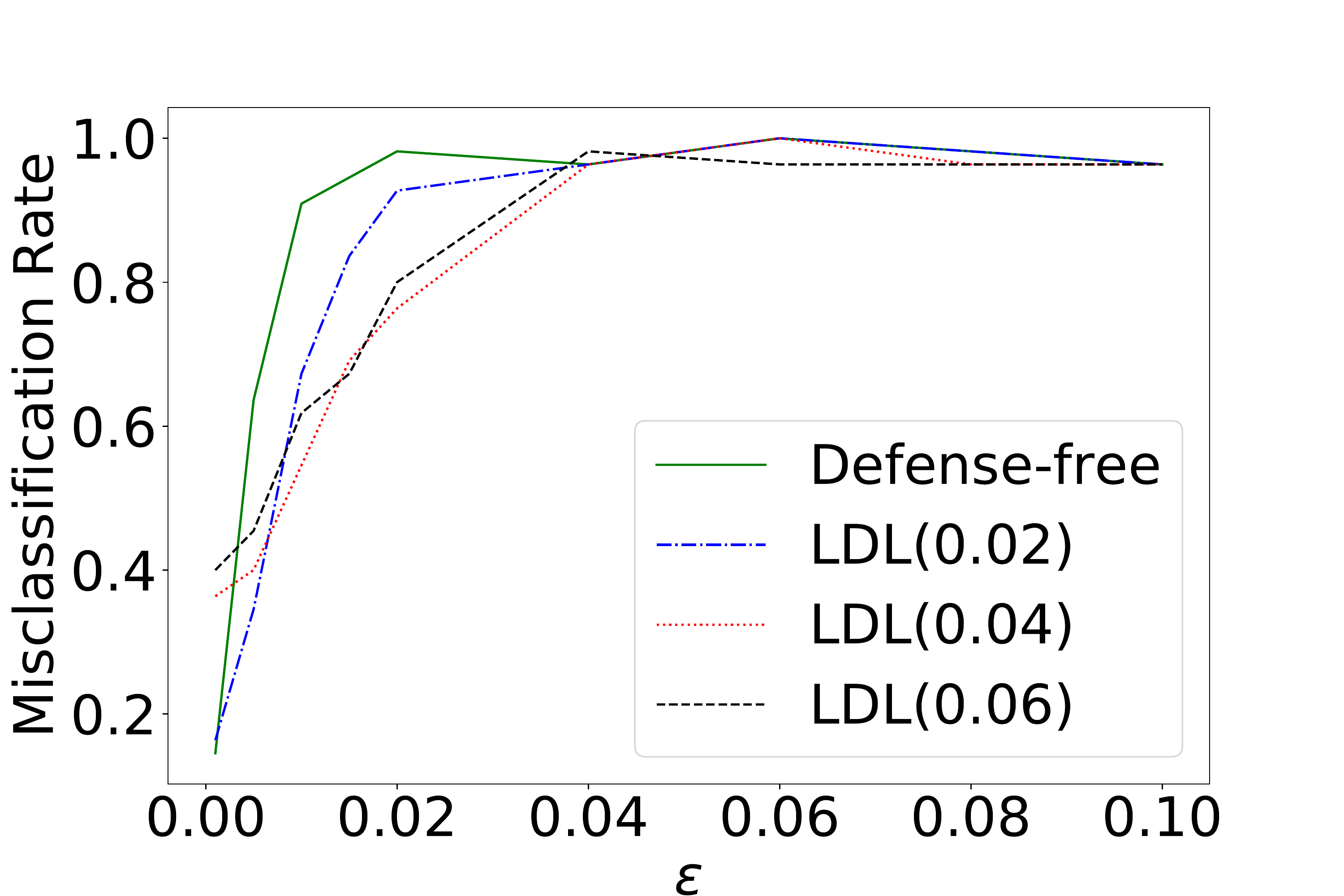}\\
    & Face (1400) & Face (1000) &Face (700) &Face (300) \\
    
    \end{tabular}
    
    \caption{
    Misclassification rates of samples of CIFAR-100, GTSRB and Face datasets with different sizes of training sets with increasing amounts of noise perturbation $\epsilon$. 
%
Deploying LDL ensures that the misclassification rate accomplished by the addition of adversarial noise is similar for both members (training dataset) and nonmembers (testing dataset) in each case. 
}    
    \label{fig:fgsall}
\end{figure*}

Figure~\ref{fig:fgsall} presents misclassification rates of samples of CIFAR-100, GTSRB and Face datasets (also see Fig.~\ref{fig:fgs}) with different sizes of training sets with increasing amounts of noise perturbation $\epsilon$. 
%
We compare the defense-free case and when LDL is deployed using $\sigma^2 = \{0.02, 0.04, 0.06\}$. 
We observe that deploying LDL ensures that the misclassification rate accomplished by the addition of adversarial noise is similar for both members (training dataset) and nonmembers (testing dataset). 

\end{document}